\documentclass{article} %
\usepackage{iclr2027_conference,times}

\usepackage[utf8]{inputenc} %
\usepackage[T1]{fontenc}    %
\usepackage{hyperref}       %
\usepackage{url}            %
\usepackage{booktabs}       %
\usepackage{amsfonts}       %
\usepackage{nicefrac}       %
\usepackage{microtype}      %
\usepackage[dvipsnames]{xcolor}        %
\usepackage{graphicx} %
\usepackage{tabularx} %
\usepackage{multirow}
\usepackage{pifont}
\usepackage{wrapfig}
\usepackage{caption}
\usepackage{subcaption}

\usepackage{amsmath,amsfonts,bm}

\def\eqref#1{equation~\ref{#1}}

\def\1{\bm{1}}

\def\vz{{\bm{z}}}

\DeclareMathAlphabet{\mathsfit}{\encodingdefault}{\sfdefault}{m}{sl}
\SetMathAlphabet{\mathsfit}{bold}{\encodingdefault}{\sfdefault}{bx}{n}

\def\sX{{\mathbb{X}}}
\def\sY{{\mathbb{Y}}}

\newcommand{\E}{\mathbb{E}}

\newcommand{\sigmoid}{\sigma}

\DeclareMathOperator*{\argmin}{arg\,min}

\newcommand{\lvar}{\ensuremath{\theta_{cal}}}
\newcommand{\lset}{\ensuremath{\sY}}
\newcommand{\lopt}{\ensuremath{\mathcal{F}_{cal}}}

\newcommand{\fvar}{\ensuremath{\theta_{train}}}
\newcommand{\fset}{\ensuremath{\sX}}
\newcommand{\fopt}{\ensuremath{\mathcal{F}_{train}}}
\newcommand{\fsolutions}{\ensuremath{S(\lvar)}}

\newcommand{\eitherlevel}{\ensuremath{\mathcal{F}_j}}
\newcommand{\eithervar}{\ensuremath{\theta_j}}

\newcommand{\logits}{\ensuremath{\vz}}
\newcommand{\sample}{\ensuremath{x}}

\newcommand{\numclasses}{\ensuremath{K}}

\newcommand{\temperature}{\ensuremath{T}}
\newcommand{\lrcal}{\ensuremath{\lambda_{c}}}
\newcommand{\lrmodel}{\ensuremath{\lambda_{f}}}

\newcommand{\basemodel}{\ensuremath{f}}

\newcommand{\calibrator}{\ensuremath{c}}
\newcommand{\fullmodel}{\ensuremath{\calibrator \circ \basemodel}}

\newcommand{\losstrain}{\ensuremath{\mathcal{L}_{train}}}
\newcommand{\losscal}{\ensuremath{\mathcal{L}_{cal}}}

\newcommand{\calset}{\ensuremath{\mathcal{D_{\mathrm{cal}}}}}
\newcommand{\trainset}{\ensuremath{\mathcal{D_{\mathrm{train}}}}}

\newcommand{\method}{COG}

\newcommand{\cmark}{\textcolor{green}{\ding{51}}} %

\newcommand{\xmark}{\textcolor{red}{\ding{55}}}

\title{Calibration in One Go: Jointly Learning\\ Accurate and Calibrated Neural Networks}

\author{Nektarios Kalampalikis\thanks{kalampalikis@cs.uni-saarland.de} \\
Dept. of Computer Science\\ 
Saarland University, Germany\\
\And
Pablo M.~Olmos  \\
Dept. of Signal Theory and Communications \\
Universidad Carlos III de Madrid, Spain \\
\AND
Isabel Valera \\
Dept. of Computer Science \\
Saarland University, Germany \\
}

\iclrfinalcopy %
\begin{document}

\maketitle
\vspace{-0.5cm}
\begin{abstract}
Neural networks (NNs), while highly flexible and versatile, are often poorly calibrated. Post-hoc calibration methods adjust the outputs of an already trained model using a separate calibration dataset and additional calibration parameters. This process adds overhead that requires separate optimization of the calibrator and subsequent model selection.
In this paper, we show that standard (post-hoc) calibrators can be trained end-to-end with the backbone via a bilevel Stackelberg formulation.
This formulation eliminates the need for a separate calibration set or a dedicated calibration loss by 
jointly optimizing accuracy and calibration using cross-entropy, introducing only one additional hyperparameter: the calibrator learning rate. %
Although our framework supports calibrators ranging from temperature and matrix scaling to NN layers, our results show that using the final classification layer itself as the calibrator yields accurate, well-calibrated models with no additional parameters.
Our approach outperforms SOTA calibrators with minimal added complexity, while also avoiding the multiplicative overhead of separate calibration components in settings such as multi-task learning.

\end{abstract}

\section{Introduction}
Machine learning, and neural networks in particular, have become an important part of modern society, with their predictions increasingly informing downstream decision making. 
In many classification settings, however, the predicted class alone is insufficient: downstream systems may instead rely on the predicted probability of each outcome as a measure of uncertainty. Such probabilities play an important role in high-stakes applications, including healthcare~\citep{ricci2023towards}, fair decision-making~\citep{guptafacct}, and machine learning-based risk assessment~\citep{brito2023subjective}, among others~\citep{wang2023calibration,he2026survey}. Beyond supporting decision making, predictive probabilities can also provide insight into what a model is unsure about, thus offering an additional layer of model understanding and human-AI collaboration~\citep{TOMSETT2020100049}.

\begin{figure} 
    \centering 
    \begin{subfigure}[t]{0.328\textwidth}
         \centering
          \includegraphics[width=\linewidth]{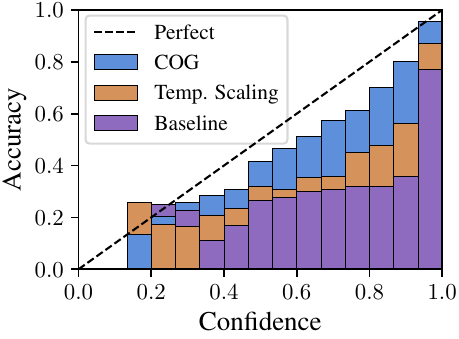} 
         \caption{}
         \label{fig:intro_reliability}
    \end{subfigure}
     \centering
     \begin{subfigure}[t]{0.33\textwidth}
         \centering
         \raisebox{1cm}
         {\includegraphics[trim=0.75cm 0.2cm 0.8cm 0.1cm, clip,width=\linewidth]{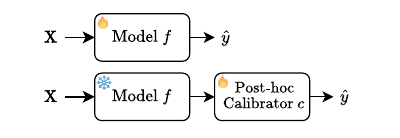} }         
        \caption{}
         \label{fig:post_hoc_arch}
    \end{subfigure}
    \centering
     \begin{subfigure}[t]{0.33\textwidth}
         \centering
         \raisebox{0.7cm}
         {\includegraphics[trim=0.75cm 0.4cm 0.8cm 0.1cm, clip,width=\linewidth]{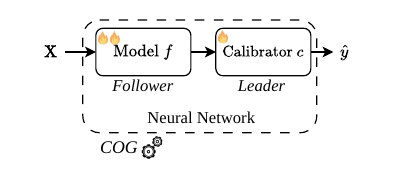} }
         
        \caption{}
         \label{fig:cog_arch}
    \end{subfigure}
     \caption{\textbf{(a)} Reliability diagrams on CIFAR-100 for the baseline, COG, and temperature scaling, with COG achieving the best calibration. \textbf{(b)} Post-hoc calibration: the model is trained and frozen before fitting a calibrator. \textbf{(c)} Our framework, COG, integrates the calibrator into the training procedure and constrains its learning rate to be slower than that of the model.}
     \vspace{-0.425cm}
\end{figure}

However, for the predicted class probabilities to be useful in these settings, they must accurately reflect the model's predictive uncertainty. In practice, machine learning models, and neural networks in particular, are often miscalibrated%
~\citep{guo2017calibration}. 
Classifier calibration aims to address this issue by aligning a classifier's predicted probabilities with the observed likelihood of the corresponding outcomes~\citep{platt1999probabilistic,guo2017calibration,pml2Book}. %
For instance, among predictions assigned 90\% probability to class $C$, class $C$ should be correct 90\% of the time. For example, as illustrated by the reliability diagram in Fig.~\ref{fig:intro_reliability}, a perfectly calibrated classifier lies along the diagonal, whereas deviations from the diagonal indicate calibration error~\citep{guo2017calibration}.
Various approaches have been proposed to calibrate models, including train-time methods and post-hoc methods~\citep{wang2023calibration}. Train-time approaches often incorporate additional calibration-related terms into the training objective, whereas post-hoc approaches train a separate model on the output of a trained model~\citep{guo2017calibration}, as illustrated in Fig.~\ref{fig:post_hoc_arch}.

Although effective, these approaches introduce additional complexity into the model development pipeline.
Train-time methods require the model designer to choose and tune additional terms or use a calibration-specific objective, while post-hoc methods require training an additional model and, typically, a separate held-out calibration set. 
On top of this, both of these families of methods come with their own complexity. For example, post-hoc calibrators can themselves overfit, motivating their regularization~\citep{berta2025rethinking,berta2026calarena,berta2025structured} or using simpler calibrators such as Temperature Scaling~\citep{guo2017calibration}. 
This pipeline complexity is particularly pronounced in settings like Multi-Task Learning (MTL)~\citep{caruana1997multitask,javaloy2022rotograd}, where a shared model serves multiple tasks, potentially requiring a separate calibrator for each. Moreover, the richer annotation requirements of MTL make reserving additional data exclusively for calibration particularly costly. Despite these challenges, calibration in MTL remains comparatively underexplored.

\paragraph{Motivation.} In this paper, we aim to obtain neural networks that are simultaneously accurate and well calibrated \emph{in a single training procedure}, while retaining the efficiency of standard training and avoiding a separate post-hoc calibration stage or additional calibration data. 
To this end, we formulate calibration as a bilevel optimization (BLO) problem~\citep{dempe2002foundations,beck2026linear}, where standard model training is the follower and the calibrator acts as the leader, as shown in Fig.~\ref{fig:cog_arch}. We efficiently solve the resulting Stackelberg problem with a single forward and backward pass per iteration, jointly updating model and calibrator parameters within the same training step using separate learning rates. This introduces one additional hyperparameter beyond standard training: the calibrator learning rate.

The resulting framework integrates calibration directly into end-to-end training while retaining the efficiency of standard training: (i) it requires neither a calibration-specific loss nor a separate calibration dataset; (ii) it imposes no additional architectural constraints and accommodates calibrators ranging from temperature and matrix scaling to neural layers, adding at most the trainable parameters of the corresponding post-hoc calibrator; and (iii) it introduces only one additional hyperparameter, namely the calibrator learning rate. 

Empirically, our approach achieves accuracy and calibration competitive with substantially more complex state-of-the-art methods while limiting the additional calibration overhead to a single hyperparameter. %
\textit{Remarkably, additional calibration parameters are unnecessary: using the final classification layer itself as the calibrator yields accurate, well-calibrated models}, as demonstrated by the example in Fig.~\ref{fig:intro_reliability}. 
We further show that this single-stage, low-overhead formulation extends naturally to more complex settings such as MTL, where conventional calibration is cumbersome.

\vspace{-0.2em}
\section{Related Works}
\vspace{-0.2em}

\paragraph{Post-hoc calibration.} 
Post-hoc calibration modifies the output probabilities of a trained classification model by learning a transformation of its outputs on a held-out calibration set.
Given an input $\sample$, a classifier $\basemodel$ is typically trained with cross-entropy (CE), after which a calibrator $\calibrator$ is fitted to a held-out calibration set, resulting in the calibrated model $\fullmodel(\sample) = \calibrator(\basemodel(\sample))$.
Parametric post-hoc calibrators include Platt Scaling~\citep{platt1999probabilistic}, designed for binary classification, which applies a learned sigmoid transformation to the logits $\logits$, i.e., $\sigmoid(a\logits+b)$. Multi-class extensions include Vector Scaling and Matrix Scaling (MS)~\citep{guo2017calibration}, which learn more flexible transformations of the logits. Temperature Scaling (TS)~\citep{guo2017calibration} uses a single learned scalar $\temperature$ to rescale the logits,~$\sigmoid(\logits/\temperature)$. 
Despite their simplicity, these models can still overfit the calibration set, motivating recent work on regularizing calibrators~\citep{berta2025rethinking,berta2026calarena,berta2025structured}. 
Nonparametric approaches include histogram binning~\citep{zadrozny2001obtaining} and isotonic regression~\citep{zadrozny2002transforming}, which directly estimate calibration mappings from empirical frequencies. However, their flexibility can make them prone to overfitting, particularly with limited calibration data~\citep{menon2012predicting}. Moreover, multiclass extensions typically rely on one-vs-rest schemes~\citep{zadrozny2002transforming,pml2Book}, substantially increasing pipeline complexity. Overviews of post-hoc calibration methods can be found in~\citep{guo2017calibration,pml2Book,berta2026calarena}.

\paragraph{Progressive training approaches.} More recently, multi-stage neural network training procedures have been proposed to improve calibration. In particular, Two-Stage Training (TST)~\citep{pablos,khodaverdian2025a} first trains the network and then retrains only some of the final fully connected layers while freezing earlier representations. 
This can be extended to impose a probabilistic prior on the final hidden layer, turning the model in a last-layer Bayesian Neural Network, that further enhances calibration (V-TST)~\citep{pablos}. 
Other works connect poor calibration to over-compression~\citep{tao2024a,wang2024calibration}, especially in later layers through the information bottleneck principle, with~\citet{wang2024calibration} proposing progressively freezing earlier layers during training.
The aforementioned methods can be cast as bilevel (or multi-level) optimization problems, which we convert into train-time calibration approaches.  
\paragraph{Train-time calibration.} Train-time calibration methods modify the learning objective or training procedure to encourage calibrated outputs. Examples include adding $L_p$ regularization~\citep{joo2020revisiting}, incorporating sliced Wasserstein distance penalties~\citep{joo2020revisiting}, or including differentiable bin membership terms in the loss~\citep{karandikar2021soft}. Other approaches include label smoothing~\citep{szegedy2016rethinking,muller2019does} or replacing CE with focal loss~\citep{mukhoti2020calibrating}, both of which can be interpreted as CE with an additional maximum-entropy regularizer~\citep{meister2020generalized,mukhoti2020calibrating}. 
Finally, several works propose specialized loss functions designed to produce calibrated models~\citep{bohdal2021meta,popordanoska2022consistent}. %
Our approach provides a complementary train-time mechanism: rather than requiring a particular calibration loss, it modifies the optimization structure and can therefore be used with standard training objectives, such as CE, or combined with existing calibration-aware losses, as seen in our experiments.

\paragraph{Bilevel Optimization and Calibration.}
Our work leverages BLO to turn post-hoc calibration into a train-time procedure. Related work has also explored calibration through the lens of BLO.
In~\citep{bohdal2021meta}, meta-learning alternates between CE updates for model training and optimization of the hyperparameters of calibration regularizers. Similarly, BLO has been used to learn validation sample weights to improve calibration~\citep{sanguin2025exploring}. This approach is computationally demanding, requiring multiple optimization steps to approximate the bilevel problem~\citep{blometalearning}, with a follower minimizing CE on $\trainset$ and a leader optimizing validation-sample weights.
These approaches are orthogonal to ours, as they either enhance existing calibration methods via BLO or formulate a different bilevel objective. Our approach instead addresses calibration within standard backpropagation-based training, avoiding expensive optimization and held-out calibration data.
\vspace{-0.55em}
\section{Towards Calibration in One Go}\label{sec:methodology}
\vspace{-0.4em}

\paragraph{(Post-hoc) Calibration as a Bilevel Optimization Problem.}
Bilevel optimization problems are hierarchical optimization problems involving two decision makers, a leader and a follower. The leader makes a decision while anticipating the follower’s best response to that decision, thus inducing a hierarchical dependency between the two levels~\citep{dempe2002foundations,beck2026linear}. Learning a neural network that is both accurate and well calibrated can be formulated as a BLO problem, in which predictive accuracy is optimized at the follower level and calibration at the leader level:
\begin{equation}
\begin{aligned}
& \min_{\lvar} \, \E_{(\sample,y)\sim \mathcal{D}} [\losscal (\calibrator_{\lvar}(\basemodel_{\fvar^*}(\sample)),y) ]  && \text{(Leader)}  \\
& \text{s.t. } \fvar^*=\argmin_{\fvar} \, \E_{(\sample,y) \sim \mathcal{D}}[
\losstrain (\basemodel_{\fvar}(\sample),y)] && \text{(Follower)}
\end{aligned}
\label{eq: bilevel}
\end{equation}
where $\fvar$ and $\lvar$ denote the parameters of the neural network $\basemodel_{\fvar}$ and calibrator $\calibrator_{\lvar}$, respectively, and $\losstrain$ and $\losscal$ their corresponding objectives. Thus, the term $\calibrator_{\lvar}(\basemodel_{\fvar}(\sample))$ denotes the resulting calibrated classifier. $(\sample,y)$ denotes a sample drawn from the data distribution $\mathcal{D}$.

Standard post-hoc calibration first trains the predictive model on a training set and then, keeping the resulting model parameters $\fvar^*$ fixed, learns the calibration parameters on a separate calibration set. That is, calibration is performed only after model training, so the calibration objective does not influence which predictive-model solution is learned, thus skipping their interaction. From a BLO perspective, post-hoc calibration can therefore be interpreted as a decoupled approximation of the BLO problem in Eq.~(\ref{eq: bilevel}), in which the follower is solved first and its solution is subsequently treated as fixed when optimizing the leader.

However, since decoupling is optimal only when the follower problem admits a unique solution~\citep{dempe2002foundations,beck2026linear}, it is generally not justified for neural network training, where multiple equivalent local optima typically exist~\citep{breiman2001statistical,goodfellow2016,rudin2024amazing}.  
This motivates our approach to jointly optimize model training and calibration: rather than fixing an arbitrary follower solution before calibration, \textit{we allow the calibrator  to influence which neural network parameters are learned}. Our goal is therefore to solve the bilevel problem above in a way that jointly learns accurate and well-calibrated neural networks while retaining the efficiency of standard neural network training.

\paragraph{Calibration in One Go (COG).} %
The bilevel formulation of Eq.~(\ref{eq: bilevel}) naturally admits a Stackelberg-game interpretation~\citep{von2010market}: the calibrator acts as the leader, updating the calibration parameters $\lvar$, while anticipating the response of the classifier (the follower), which in turn adapts the neural network parameters $\fvar$ to minimize the training loss.
This joint problem can be addressed efficiently using gradient-based bilevel optimization methods~\citep{liu2021investigating}. Specifically, we employ a first-order, Stackelberg-inspired scheme in which the leader and follower parameters are \textit{simultaneously updated} iteratively using gradients of their respective objectives, with distinct learning rates for the two levels, in the spirit of Gradient Descent-Ascent (GDA)~\citep{stackelbergdynamics,javaloy2022rotograd}. Further details on BLO and GDA are provided in App~\ref{app:theory}.

For computational efficiency, we further make two simplifying choices. First, we use the same loss function for both levels, e.g.,  cross-entropy loss, such that $\losscal := \losstrain$. Second, we approximate both expectations in Eq.~(\ref{eq: bilevel}) using the same minibatch $\text{B}_t$ from the training data, i.e., $\calset=\trainset$. Consequently,  \textit{both objectives can be evaluated on the same forward pass and their gradients computed with a single backpropagation pass}, as in standard neural network training. Thus, the resulting procedure differs from standard training only in that the neural network and calibration parameters are updated with distinct learning rates as: %

\begin{equation}
\begin{aligned}
& \lvar^{t+1}=\lvar^{t} - \lrcal  \sum_{i \in\text{B}_t} \nabla_{\lvar}\losstrain (\calibrator_{\lvar}(\basemodel_{\fvar}(\sample_i)),y_i) \\
& \fvar^{t+1}=\fvar^{t} - \lrmodel \sum_{i \in\text{B}_t} \nabla_{\fvar}\losstrain (\calibrator_{\lvar}(\basemodel_{\fvar}(\sample_i)),y_i), \quad \text{s.t. } \lrcal<\lrmodel\\
\end{aligned}
\vspace{-1em}
\end{equation}

where $\lrcal$ and $\lrmodel$ denote the learning rates of the leader and follower, respectively. The condition $\lrcal<\lrmodel$ preserves the Stackelberg hierarchy by ensuring that the follower adapts on a faster timescale than the leader, which under standard assumptions guarantees asymptotic convergence of the joint dynamics to a local Stackelberg solution~\citep{stackelbergdynamics,javaloy2022rotograd}. 
We remark that, while these joint gradient updates do not guarantee convergence to an optimal Stackelberg solution, they provide a simple and efficient approximation that performs well in practice~\citep{stackelbergdynamics,javaloy2022rotograd}.

\paragraph{From Post-hoc Calibration to COG.}
Implementing standard post-hoc calibration methods within COG is straightforward and independent of the neural network's architecture. For example, Temperature Scaling can be implemented by introducing a single additional parameter (COG-TS), Matrix Scaling by an additional weight matrix and an additional bias vector (COG-MS), and TST by introducing multiple calibration layers (COG-TST).
\textit{COG can also be applied without introducing additional trainable parameters} (COG-FC). Recent work has shown that the internal representations of a neural network can be as effective for prediction as its final classification layer~\citep{wang2024calibration}; thus, the final classification layer can be treated as the calibrator within COG, allowing the method to be applied to arbitrary neural network architectures.
Finally, COG is not only restricted to post-hoc CE-based calibration. For example, by replacing the objectives at both levels with the Label Smoothing objective, we obtain COG+LS. Together, these variants demonstrate the flexibility of the COG framework across different calibration methods.

A summary of the different calibration options and \method\ variants is provided in Tab.~\ref{tab:qualitative_comparison}. The \method\ variants provide more lightweight, end-to-end, alternatives to their corresponding post-hoc calibration approaches in both single-task and multi-task settings (Tab.~\ref{tab:qualitative_comparison_full}). In the single-task setting, \method-FC has a complexity comparable to Label Smoothing, while in the multi-task setting, \method-FC is the simplest calibration approach among all evaluated methods.
\begin{table}
    \small
    \caption{Complexity of different calibration methods in Single-Task settings, where \numclasses\ denotes the number of classes. Tab.~\ref{tab:qualitative_comparison_full} extends this comparison to MTL settings.}
    \label{tab:qualitative_comparison}
    \centering
    \begin{tabular}{cc|cccc}
    \toprule
    &        & \# Extra           & End & Calibration & Extra  \\
    & \textbf{Method} &  Trainable Params. &  To End                     &  Data    & Burden\\
         \midrule
         \multirow{15}{*}{\rotatebox{90}{\textbf{Single-Task}}}
         &Baseline&  0&  \cmark & \xmark & -\\
         &COG-FC&  0&  \cmark & \xmark & 1 Learning Rate\\
         \cmidrule(lr){2-6}
         & Label Smoothing & 0 & \cmark & \xmark & 1 Smoothing Level \\
         & \multirow{2}{*}{COG+LS}
                & \multirow{2}{*}{0}
                & \multirow{2}{*}{\cmark}
                & \multirow{2}{*}{\xmark}
                & \multirow{2}{*}{\begin{tabular}[c]{@{}l@{}}
                (a) 1 Smoothing Level \\
                (b) 1 Learning Rate\\
                \end{tabular}} \\
                & & & & \\
         \cmidrule(lr){2-6}
         &Temp. Scaling& 1 &  \xmark & \cmark & Calibrator Training\\
         &COG-TS& 1 &  \cmark & \xmark &1 Learning Rate\\
         \cmidrule(lr){2-6}
         &Matrix Scaling&  $\numclasses \times \numclasses + \numclasses$ &  \cmark & \cmark & Calibrator Training\\
         &COG-MS&  $\numclasses \times \numclasses + \numclasses$&  \cmark & \xmark & 1 Learning Rate\\
         \cmidrule(lr){2-6}
         &TST    & NN & \xmark & \xmark & NN Training, Architecture \\
         &COG-TST   & NN & \cmark & \xmark &1 Learning Rate\\
         \cmidrule(lr){2-6}
         & \multirow{3}{*}{V-TST}
                & \multirow{3}{*}{2 NNs}
                & \multirow{3}{*}{\xmark}
                & \multirow{3}{*}{\xmark}
                & \multirow{3}{*}{\begin{tabular}[c]{@{}l@{}}
                (a) ELBO training of 2 NNs \\
                (b) NN architecture\\
                (c) MC Sampling
                \end{tabular}} \\
                & & & & \\
                & & & &\\

    \bottomrule
    \end{tabular}

\end{table}

\section{Experiments}\label{sec:experiments}
We structure our empirical evaluation in three stages. First, we compare standard post-hoc calibration methods with their \method\ counterparts to isolate the effect of replacing post-hoc calibration with joint training. Second, we benchmark \method\ against calibrators from different methodological families under increasingly challenging conditions, including in-distribution data, out-of-distribution data, and covariate shift. Finally, we evaluate \method\ in the more complex setting of multi-task learning, where calibration must be achieved jointly across multiple prediction tasks.

\paragraph{Experimental Setup and Metrics} 
We focus on multiclass image classification and train ResNet101 models~\citep{he2016deep} from scratch on SVHN~\citep{netzer2011reading}, CIFAR-10, and CIFAR-100~\citep{krizhevsky2009learning}, following the setup of~\citep{pablos}. Full experimental details are provided in App.~\ref{app:exp_details}, extended results in App.~\ref{app:extended_results}, and experiments with additional CNN backbones in App.~\ref{app:wrn}.
We report accuracy together with two scoring rules, negative log-likelihood (NLL) and the Brier score~\citep{glenn1950verification}, following recent recommendations for calibration evaluation~\citep{minderer2021revisiting,silva2023classifier,berta2025rethinking,khodaverdian2025a,chidambaram2025reassessing}. App.~\ref{app:extended_results} additionally reports ECE~\citep{naeini2015obtaining}, RMSCE~\citep{hendrycks2018deep} and MCE~\citep{naeini2015obtaining} across multiple binning choices.

\subsection{Comparing Post-hoc Methods and Their COG Counterparts}\label{sec:cog_vs_post}
We compare each post-hoc calibrator with its corresponding \method\ variant in a pairwise manner, as shown in Tab.~\ref{tab:qualitative_comparison}, matching the number of calibration parameters within each pair. For all \method\ variants and (V-)TST, calibration is learned jointly on the training set, whereas post-hoc calibrators are fitted on the validation set. %

\begin{figure}
    \centering
    \includegraphics[width=\linewidth]{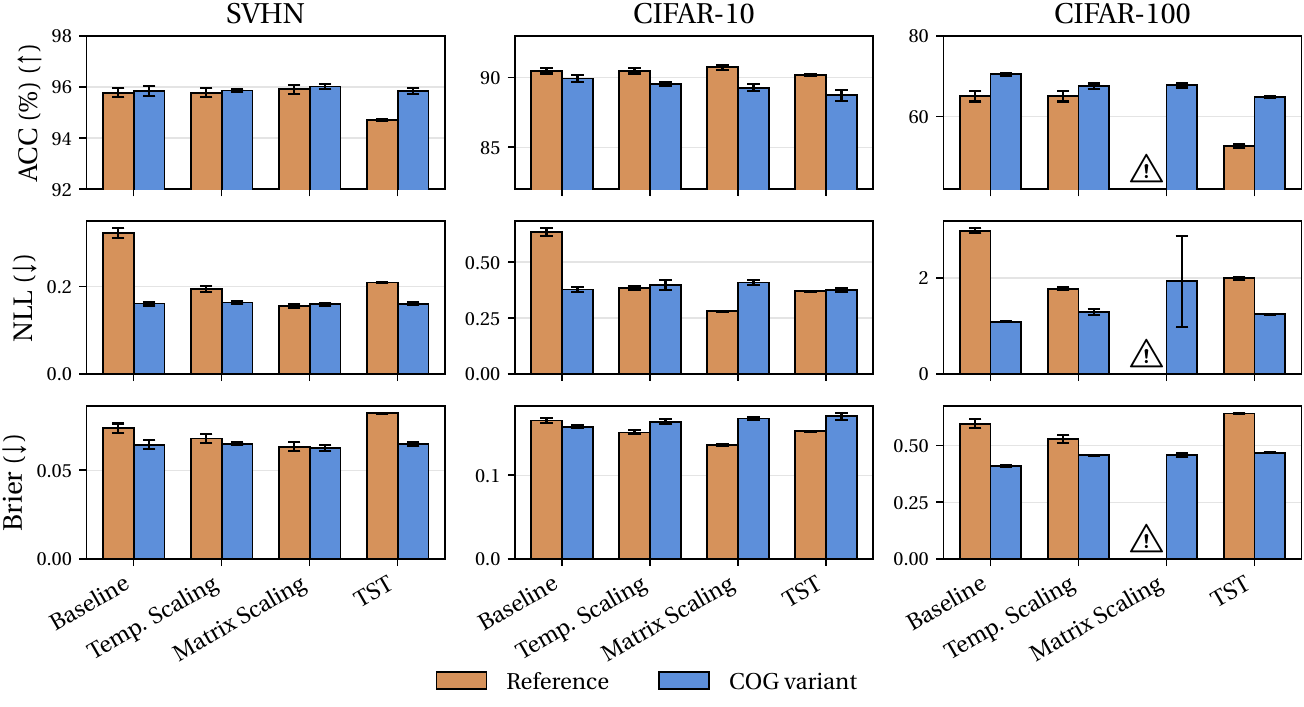}
    \caption{Pairwise comparison of different calibrators and their corresponding \method\ variants. The warning symbol indicates failure to converge.}
    \label{fig:post_vs_cog}
\end{figure}

\paragraph{Results} As shown in Fig.~\ref{fig:post_vs_cog}, the \method\ variants generally match or outperform their post-hoc counterparts across datasets and metrics, with differences otherwise largely within the observed variance. 
One notable exception is Matrix Scaling on CIFAR-10, where the post-hoc variant appears to benefit from direct access to calibration data. On CIFAR-100, however, the post-hoc Matrix Scaling optimization fails to converge, whereas \method-MS remains stable and trains successfully within our joint optimization framework.
We further observe that the different \method\ variants perform very similarly across all three metrics and datasets, suggesting that additional calibrator complexity provides limited benefit. In particular, \method-FC, which is our approach without additional trainable parameters, performs on par with the more expressive calibration approaches. 
Overall, these results demonstrate that \textit{\method-FC provides a simple yet effective way to obtain models that are both accurate and well calibrated, without requiring a separate calibration stage, additional calibration parameters, or calibration data}.
In App.~\ref{app:cog_vs_post}, we additionally report calibration error metrics (Tables \ref{tab:cog_vs_post_ece}-\ref{tab:cog_vs_post_mce}), and reliability diagrams (Fig.~\ref{fig:reliability_diagram_app}).

\subsection{Comparison Across Calibration Families}\label{sec:compare_families}

We compare representative methods from four  families of calibrators: \emph{train-time} calibration via Label Smoothing, \emph{post-hoc} calibration via Temperature Scaling, a Bayesian calibration method via V-TST (with 10 Monte Carlo samples)~\citep{pablos,khodaverdian2025a}, and our  (also train-time) calibration approach \method-FC and its Label Smoothing variant \method+LS. We select \method-FC as the representative \method\ variant because it achieves strong performance while introducing no additional calibration parameters; \method+LS additionally replaces the standard cross-entropy loss with the calibration-regularized objective used by Label Smoothing. %
Moreover, since no single metric fully captures the trade-off between predictive accuracy and calibration quality, several works recommend considering the Pareto front between these objectives, allowing their relative importance to be determined by the application~\citep{minderer2021revisiting,silva2023classifier,chidambaram2025reassessing,sharifdeen2026mvc}. For a compact summary, we compute, for each dataset, the unweighted average rank of each method across Accuracy, NLL, and Brier score~\citep{javaloy2025copa}. We denote the method with the best average rank as \textbf{Top-ranked}.
App.~\ref{app:app_everything_compared} and~\ref{app:shift_ood_extended}, report results for all considered calibration methods and calibration errors.

\begin{figure}
    \centering
    \includegraphics[width=0.93\linewidth]{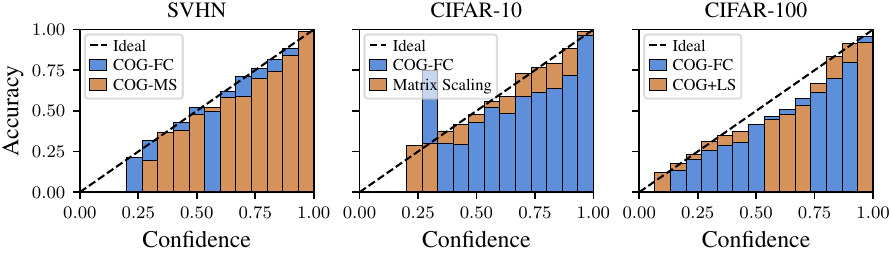}
    \caption{Reliability diagrams for \method-FC and the top-ranked methods using 15 bins. For CIFAR-100, where \method-FC is the top-ranked method, we compare it with \method+LS.}
    \label{fig:reliability_top_ranked}
\end{figure}

\paragraph{In-distribution (ID) results.}
As shown in Tab.~\ref{tab:main_family_comparison}, \method-FC still remains competitive across all methods and is the top ranked method in CIFAR-100. 
We also observe that \method+LS is highly competitive, generally outperforming the Label Smoothing variant especially in accuracy.
More broadly, \method\ variants remain highly competitive, with a \method\ variant being the top-ranked method in two of the three evaluated cases. 
In Fig.~\ref{fig:reliability_top_ranked}, we compare the reliability diagrams of \method-FC and the top-ranked model. For CIFAR-100, we compare \method-FC with \method+LS. Overall, the differences are small, with \method-FC closely matching the top-ranked models, with a deviation for CIFAR-10 around 25\% confidence.
These results further demonstrate that effective \textit{in-distribution calibration of neural networks can be achieved without increasing the model's parameter count or substantially increasing the complexity of the overall pipeline}.

\begin{table}[t]
\caption{Comparison of calibration metrics across datasets, and methods. The best value in each column among the subset of models is bolded. The \textbf{Top $\text{Ranked}^*$} row reports the best-performing model per dataset based on average rank across metrics as determined from the full results in Tab.~\ref{tab:app_all_VS_all_ID}: \method-MS on SVHN, Matrix Scaling on CIFAR-10, and \method-FC on CIFAR-100.}
\label{tab:main_family_comparison}
\centering
\resizebox{\textwidth}{!}{
\begin{tabular}{l|ccc|ccc|ccc}
\toprule
   & \multicolumn{3}{c|}{\textbf{SVHN}} & \multicolumn{3}{c|}{\textbf{CIFAR-10}} & \multicolumn{3}{c}{\textbf{CIFAR-100}} \\
  \textbf{Method} & ACC $(\uparrow)$ & NLL $(\downarrow)$ & Brier $(\downarrow)$ & ACC $(\uparrow)$ & NLL $(\downarrow)$ & Brier $(\downarrow)$ & ACC $(\uparrow)$ & NLL $(\downarrow)$ & Brier $(\downarrow)$ \\
\midrule
 Baseline & $95.77_{0.18}$ & $0.32_{0.01}$ & $0.07_{0.00}$ & $\mathbf{90.50_{0.21}}$ & $0.64_{0.02}$ & $0.17_{0.00}$ & $65.04_{1.33}$ & $2.98_{0.06}$ & $0.60_{0.02}$ \\
 \midrule
  COG-FC & $95.84_{0.20}$ & $\mathbf{0.16_{0.00}}$ & $\mathbf{0.06_{0.00}}$ & $89.94_{0.25}$ & $0.38_{0.01}$ & $0.16_{0.00}$ & $70.46_{0.30}$ & $\mathbf{1.08_{0.01}}$ & $\mathbf{0.41_{0.00}}$ \\
  COG+LS & $\mathbf{96.25_{0.15}}$ & $0.21_{0.00}$ & $0.07_{0.00}$ & $89.92_{0.21}$ & $0.38_{0.00}$ & $0.16_{0.00}$ & $\mathbf{72.67_{0.78}}$ & $1.31_{0.04}$ & $\mathbf{0.41_{0.01}}$ \\
  Label Smoothing & $95.97_{0.12}$ & $0.23_{0.00}$ & $0.07_{0.00}$ & $90.05_{0.31}$ & $0.43_{0.01}$ & $0.16_{0.00}$ & $65.76_{0.88}$ & $1.58_{0.04}$ & $0.48_{0.01}$ \\
  Temp. Scaling & $95.77_{0.18}$ & $0.19_{0.01}$ & $0.07_{0.00}$ & $\mathbf{90.50_{0.21}}$ & $0.38_{0.01}$ & $\mathbf{0.15_{0.00}}$ & $65.04_{1.33}$ & $1.78_{0.03}$ & $0.53_{0.02}$ \\
  V-TST & $94.66_{0.03}$ & $0.21_{0.00}$ & $0.08_{0.00}$ & $90.31_{0.16}$ & $\mathbf{0.34_{0.01}}$ & $\mathbf{0.15_{0.00}}$ & $52.71_{0.25}$ & $1.81_{0.01}$ & $0.61_{0.00}$ \\
\midrule
  Top $\text{Ranked}^*$ & $96.02_{0.11}$ & $0.16_{0.00}$ & $0.06_{0.00}$ & $90.74_{0.17}$ & $0.28_{0.00}$ & $0.14_{0.00}$ & $70.46_{0.30}$ & $1.08_{0.01}$ & $0.41_{0.00}$ \\
\bottomrule
\end{tabular}
}
\end{table}

\paragraph{Out of distribution (OOD) \& covariate shift.} 
Lastly, we assess whether ID calibration transfers to OOD detection and covariate shifts, following~\citep{ovadia2019can,pablos}. For covariate shifts, we use CIFAR-10-C, CIFAR-100-C~\citep{hendrycks2019benchmarking}, and rotated SVHN~\citep{pablos}. OOD samples come from different datasets and lack valid labels in the original label space, whereas covariate shifts preserve the original labels under input corruption.
For OOD detection, we use maximum softmax probability as the confidence score and treat ID samples as positive and OOD samples as negative. We report AUROC~\citep{bradley1997use} and FPR@95TPR, where higher AUROC and lower FPR95 indicate better OOD detection.

As shown in Tab.~\ref{tab:ood_results}, \method-FC performs particularly well for OOD detection, with \method+LS also achieving strong results. Across the three  datasets, a \method\ variant achieves the best performance in each case.
Under covariate shift (Tab.~\ref{tab:shift_results}), we observe trends consistent with prior work~\citep{ovadia2019can}: regularization-based approaches such as Label Smoothing are often more robust, and \method+LS generally improves over \method-FC. At the same time, performance varies across datasets and shift types, highlighting that robustness to distribution shift is distinct from in-distribution calibration~\citep{ovadia2019can}.  We leave further improvements in covariate-shifts to future work.

\begin{table}[t]
\caption{Performance on out-of-distribution detection. The best value in each column among the subset of models is bolded. The \textbf{Top $\text{Ranked}^*$} row represents the best model per dataset based on average rank across metrics as determined from the full results in Tab.~\ref{tab:shift_ood_results_full_ood}: COG-MS on SVHN, and COG-TST on CIFAR-10 and CIFAR-100.}
\label{tab:ood_results}
\centering
\resizebox{0.8\textwidth}{!}{
\begin{tabular}{l|cc|cc|cc}
\toprule
   & \multicolumn{2}{c|}{\textbf{SVHN}} & \multicolumn{2}{c|}{\textbf{CIFAR-10}} & \multicolumn{2}{c}{\textbf{CIFAR-100}} \\
  \textbf{Method} & AUROC $(\uparrow)$ & FPR95 $(\downarrow)$ & AUROC $(\uparrow)$ & FPR95 $(\downarrow)$ & AUROC $(\uparrow)$ & FPR95 $(\downarrow)$ \\
\midrule
 Baseline & $94.78_{0.24}$ & $34.43_{1.27}$ & $83.97_{2.70}$ & $76.39_{3.20}$ & $71.92_{1.33}$ & $87.67_{1.34}$ \\
 \midrule
  COG-FC & $\mathbf{96.60_{0.41}}$ & $\mathbf{21.60_{2.59}}$ & $\mathbf{87.00_{1.32}}$ & $73.98_{4.03}$ & $68.50_{2.22}$ & $85.83_{1.43}$ \\
  COG+LS & $93.80_{0.61}$ & $22.58_{2.45}$ & $81.93_{3.10}$ & $\mathbf{73.43_{3.18}}$ & $65.48_{7.85}$ & $\mathbf{83.71_{3.42}}$ \\
  Label Smoothing & $80.58_{1.66}$ & $42.84_{2.40}$ & $72.17_{6.88}$ & $78.88_{3.09}$ & $\mathbf{74.35_{2.98}}$ & $84.06_{1.86}$ \\
  Temp. Scaling & $95.33_{0.25}$ & $31.14_{1.85}$ & $85.09_{2.43}$ & $75.04_{4.22}$ & $73.23_{1.53}$ & $85.46_{2.00}$ \\
  V-TST & $94.66_{0.47}$ & $35.35_{2.40}$ & $79.23_{2.27}$ & $78.61_{1.96}$ & $70.75_{0.85}$ & $85.78_{0.64}$ \\
\midrule
  Top $\text{Ranked}^*$ & $96.70_{0.28}$ & $20.36_{2.18}$ & $87.41_{0.70}$ & $71.37_{2.34}$ & $73.92_{2.72}$ & $79.96_{2.31}$ \\
\bottomrule
\end{tabular}
}
\end{table}

\begin{table}[t]
\caption{Performance under covariate shifts. The best value in each column among the subset of models is bolded. The \textbf{Top $\text{Ranked}^*$} row reports the best-performing model per dataset based on average rank across metrics as determined from the full results in Tab.~\ref{tab:shift_results_full}: Matrix Scaling on SVHN and CIFAR-10, Label Smoothing on CIFAR-100.} 
\label{tab:shift_results}
\centering
\setlength{\tabcolsep}{3pt}
\resizebox{\textwidth}{!}{
\begin{tabular}{l|ccc|ccc|ccc}
\toprule
   & \multicolumn{3}{c|}{\textbf{SVHN}} & \multicolumn{3}{c|}{\textbf{CIFAR-10}} & \multicolumn{3}{c}{\textbf{CIFAR-100}} \\
  \textbf{Method} & ACC $(\uparrow)$ & NLL $(\downarrow)$ & Brier $(\downarrow)$ & ACC $(\uparrow)$ & NLL $(\downarrow)$ & Brier $(\downarrow)$ & ACC $(\uparrow)$ & NLL $(\downarrow)$ & Brier $(\downarrow)$ \\
\midrule
 Baseline & $\mathbf{47.01_{1.06}}$ & $5.29_{0.23}$ & $0.94_{0.02}$ & $\mathbf{74.28_{0.59}}$ & $2.03_{0.11}$ & $0.45_{0.01}$ & $45.85_{1.36}$ & $5.70_{0.21}$ & $0.92_{0.02}$ \\
 \midrule
  COG-FC & $45.82_{1.32}$ & $2.43_{0.12}$ & $0.80_{0.02}$ & $68.80_{0.68}$ & $1.57_{0.09}$ & $0.50_{0.01}$ & $44.09_{0.42}$ & $2.88_{0.04}$ & $0.77_{0.01}$ \\
  COG+LS & $46.05_{0.57}$ & $\mathbf{1.91_{0.02}}$ & $\mathbf{0.77_{0.00}}$ & $69.31_{0.42}$ & $1.11_{0.02}$ & $0.46_{0.01}$ & $44.92_{0.69}$ & $2.90_{0.11}$ & $0.81_{0.03}$ \\
  Label Smoothing & $46.19_{0.59}$ & $2.19_{0.04}$ & $0.86_{0.01}$ & $74.22_{0.42}$ & $1.03_{0.02}$ & $0.42_{0.01}$ & $\mathbf{47.14_{0.91}}$ & $\mathbf{2.54_{0.10}}$ & $\mathbf{0.71_{0.03}}$ \\
  Temp. Scaling & $\mathbf{47.01_{1.06}}$ & $3.05_{0.12}$ & $0.85_{0.02}$ & $\mathbf{74.28_{0.59}}$ & $1.19_{0.05}$ & $0.41_{0.01}$ & $45.85_{1.36}$ & $3.32_{0.10}$ & $0.81_{0.02}$ \\
  V-TST & $41.66_{0.26}$ & $2.06_{0.07}$ & $0.77_{0.01}$ & $73.91_{0.22}$ & $\mathbf{0.91_{0.03}}$ & $\mathbf{0.38_{0.01}}$ & $38.44_{0.24}$ & $2.66_{0.02}$ & $0.77_{0.00}$ \\
\midrule
  Top $\text{Ranked}^*$ & $47.27_{0.72}$ & $2.14_{0.02}$ & $0.75_{0.01}$ & $74.35_{0.66}$ & $0.84_{0.03}$ & $0.36_{0.01}$ & $47.14_{0.91}$ & $2.54_{0.10}$ & $0.71_{0.03}$ \\
\bottomrule
\end{tabular}
}
\end{table}

\begin{wraptable}[11]{r}{0.42\textwidth} %
\centering
\vspace{-1.15\intextsep}  %
\captionsetup{skip=2.5pt}   %
\caption{Performance with a ViT backbone on CIFAR-10. Bold indicates the best value per metric}
\label{tab:vit_main}
\setlength{\tabcolsep}{3pt}
\resizebox{0.42\textwidth}{!}{
\begin{tabular}{l|ccc}
\toprule
  \textbf{Method} &  ACC $(\uparrow)$ & NLL $(\downarrow)$ & Brier $(\downarrow)$ \\
\midrule
 Baseline & $73.51_{0.57}$ & $0.81_{0.02}$ & $0.38_{0.01}$  \\
  COG-FC & $\mathbf{74.97}_{0.47}$  & $\mathbf{0.76}_{0.02}$  & $\mathbf{0.36}_{0.01}$ \\
  \midrule
 Temp. Scaling & $73.51_{0.57}$  & $\mathbf{0.76_{0.01}}$  & $0.37_{0.01}$ \\
  COG-TS & $73.73_{0.43}$  & $0.81_{0.02}$  & $0.38_{0.00}$  \\
  \midrule
 Matrix Scaling & $74.04_{0.35}$  & $0.74_{0.01}$  & $\mathbf{0.36}_{0.00}$ \\
  COG-MS & $73.14_{0.71}$ & $0.81_{0.02}$  & $0.38_{0.01}$  \\
\bottomrule
\end{tabular}
}
\end{wraptable}
\paragraph{ViT ablation.} Our experiments so far focus on CNN architectures. While App.~\ref{app:wrn} confirms consistent results with a WRN-28-10 backbone~\citep{zagoruyko2016wide}, we further assess \method\ across architectural families using Vision Transformers (ViTs)~\citep{dosovitskiy2021an}.
Applying a subset of post-hoc methods and their corresponding \method\ variants, Tab.~\ref{tab:vit_main}, shows that \mbox{\method-FC} again achieves the best overall performance, suggesting that BLO is not specific to convolutional architectures. Additional calibration metrics are reported in App.~\ref{sec:app_vit}.

\subsection{Use case: Multi-Task Learning}

\paragraph{Prerequisite Questions.}
Calibration in Multi-Task Learning remains relatively under-explored, raising two fundamental questions: how should calibration be applied in MTL, and how should calibration be quantified across heterogeneous tasks?
For the former, should each task be calibrated independently, requiring separate $\numclasses$ post-hoc calibrators or regularizers, or should a single approach be applied across tasks? 
This is particularly challenging when tasks differ in type, e.g., binary classification, multiclass classification, and regression, which may require different calibration methods. Even tasks of the same type may have different calibration requirements, further increasing the complexity of the calibration pipeline. 
\textit{In contrast, \method\ can be applied directly to MTL models using the same training objective as the network, avoiding separate calibration procedures for each task.}
The second question concerns how to quantify calibration across tasks. Given the aforementioned issues, rather than introducing MTL-specific calibration metrics, we evaluate using the previous metrics. 

\paragraph{Experimental Setup.}

\begin{figure}[t]
        \begin{minipage}[t]{0.5\textwidth}
        \captionof{figure}{MTL pipeline with \numclasses\ task-specific heads, each followed by a separate post-hoc calibrator or calibration objective.}
        \centering
        \includegraphics[trim=0.75cm 0.45cm 0.75cm 0.41cm, clip,width=\textwidth,keepaspectratio]{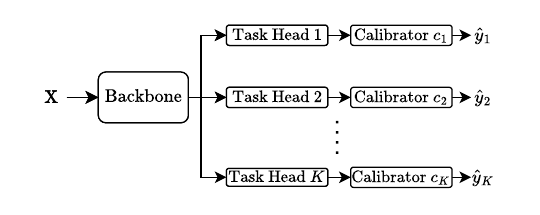}
        \label{fig:mtl_pipeline}
    \end{minipage}
    \hfill
        \begin{minipage}[t]{0.44\textwidth}
        \captionof{table}{MTL calibration performance on CelebA, averaged across tasks. Bold indicates the best value per metric.}
                \label{tab:results_resnet101_celeba_mtl_main}
                \centering
                \setlength{\tabcolsep}{3pt}
                \resizebox{\textwidth}{!}{
        \begin{tabular}{l|ccc}
                \toprule
                \textbf{Method} & ACC $(\uparrow)$ & NLL $(\downarrow)$ & Brier $(\downarrow)$ \\
                \midrule
                 Baseline & $89.88_{0.04}$ & $0.23_{0.00}$ & $0.16_{0.00}$ \\
                 COG-FC & $90.81_{0.04}$ & $\mathbf{0.21}_{0.00}$ & $\mathbf{0.14}_{0.00}$ \\
                \midrule
                 Label Smoothing & $89.91_{0.06}$ & $0.35_{0.00}$ & $0.15_{0.00}$  \\
                 COG+LS & $\mathbf{90.82}_{0.02}$ & $0.34_{0.00}$ & $\mathbf{0.14}_{0.00}$  \\
                \midrule
                 Temp. Scaling & $89.88_{0.04}$ & $0.24_{0.00}$ & $0.15_{0.00}$  \\
                 COG-TS & $90.44_{0.15}$ & $0.22_{0.01}$ & $0.15_{0.00}$ \\
                
                \bottomrule
        \end{tabular}}
    \end{minipage}    

\end{figure}

For this experiment, we use CelebA~\citep{liu2015faceattributes}, where the model predicts 40 facial attributes, i.e., 40 binary tasks. For non-\method\ methods each task is calibrated separately, resulting in 40 separate calibrators or regularization terms, as seen in Fig.~\ref{fig:mtl_pipeline}. We train using linear scalarization~\citep{miettinen1999nonlinear}, with equal task weights, i.e., $L_{total}=\sum_{i=1}^{\numclasses}L_i$ and, for every metric, report unweighted averages across tasks, e.g., $L_{total}/\numclasses$.
\paragraph{Results.} 
The results in Tab.~\ref{tab:results_resnet101_celeba_mtl_main} show broadly comparable performance across methods, particularly in Brier score. All \method\ variants outperform their corresponding baselines with \method-FC achieving the best overall performance. Both Label Smoothing methods, however, exhibit notably worse NLL. Overall, these results demonstrate that effective end-to-end calibration is achievable in MTL without additional trainable parameters or regularization terms. 
In App.~\ref{app:MTL_extended} we provide evaluation based on calibration errors and per-task metrics. According to Fig.~\ref{fig:per_task_metrics_main}, all families exhibit a similar ordering of task difficulty, with \textit{Big Lips}, for instance, consistently ranking among the most challenging tasks, suggesting that difficulty is primarily task-dependent.
In pairwise comparisons, for example, \method-FC achieves both higher Accuracy and lower Brier score than Temperature Scaling on 32 of 40 tasks (80\%), and than Label Smoothing on all 40 tasks (100\%).

\section{Conclusion}

In this work, we simplify neural network calibration by reframing post-hoc calibration as a bilevel optimization problem and solving it through Stackelberg-inspired joint optimization. 
Our framework, \method, converts post-hoc calibrators into train-time, end-to-end alternatives that require no held-out calibration data and introduce only minimal additional complexity: a single additional learning rate. 
Across a broad range of experiments, architectures, and calibration approaches, \method\ achieves strong and consistent performance, matching or outperforming existing methods in most settings. In particular, \method-FC \textit{achieves state-of-the-art calibration performance while introducing no additional trainable parameters}. We further find that \method\ is stable to optimize and performs well for OOD detection, demonstrating its flexibility across architectures, post-hoc methods, and regularization-based approaches.
We further study calibration in MTL settings, where task heterogeneity requires conventional calibration methods to be selected and applied separately for each task. In contrast, \method\ works out of the box for MTL, providing a unified calibration approach with strong overall and task-specific performance without additional trainable parameters or separate calibration procedures.
Overall, we recommend \method-FC as a simple, stable, and lightweight default for calibration, given its strong performance across diverse settings and lack of additional trainable parameters.

\paragraph{Future Work and Limitations.} 
Several directions remain for future work. First, our evaluation could be extended to alternative definitions of calibration and emerging calibration metrics, particularly those addressing limitations of standard metrics such as ECE.
Second, we do not explicitly explore optimistic and pessimistic bilevel optimization~\citep{dempe2002foundations,beck2026linear}. In optimistic BLO, the leader assumes that, among multiple optimal solutions of the follower, the follower selects the one that is most favorable to the leader. In contrast, pessimistic BLO considers the worst-case selection among the follower's optimal solutions, accounting for the least favorable outcome for the leader's objective~\citep{dempe2002foundations,beck2026linear}. Pessimistic bilevel optimization, which is closely related to robust optimization~\citep{beck2022brief,beck2023survey}, could be particularly relevant for calibration under adversarial settings or covariate shifts, while optimistic bilevel optimization may yield even better solutions for the calibrator. Further theoretical analysis of calibration from a bilevel optimization perspective could also provide additional insight.
For MTL, future work could consider more challenging and heterogeneous settings, including tasks with different output spaces or objectives, alongside state-of-the-art MTL training methods and designing calibration metrics specifically for multi-task problems. 
Finally, although our experiments focus on computer vision, \method\ is not inherently tied to a particular architecture or modality and could be extended to other differentiable models and domains.

\subsection*{AI use statement}

In this work, we used generative AI tools for research execution, specifically for code assistance, including but not limited to writing comments in code and preparing documentation. Additionally, we used generative AI tools to polish and improve the readability of the text. We did not use generative AI tools for the other required-disclosure tasks. We have reviewed all AI-assisted work, verified the correctness of the code and documentation, and reviewed the polished text to ensure that it accurately reflects the authors’ intended content. We take responsibility for the final content of this work, including text, claims, code, and other artifacts produced with the aid of generative AI.

\subsection*{Reproducibility statement}

Implementation and training details, including optimization procedures, learning-rate schedules, checkpoint selection, calibration procedures, and experimental settings, are provided in App.~\ref{app:exp_details}. The hyperparameter configurations used for the different datasets, backbones, and COG variants are reported in Tab.~\ref{tab:bestconfigs}. Dataset splits, preprocessing, augmentation, and validation procedures are described in the dataset details in App.~\ref{app:exp_details}. Additional information on the experimental hardware and computational requirements is also provided in App.~\ref{app:exp_details}. All experiments use fixed random seeds as described in the appendix, and the code and configuration files required to reproduce the reported results are provided in the supplementary materials.

\subsubsection*{Acknowledgments}
We would like to thank every member of the Probabilistic Machine Learning lab and especially Elina Rojin Celik for their
invaluable feedback.
This work was funded in part by the Deutsche Forschungsgemeinschaft (DFG, German Research Foundation)
– GRK 2853/1 “Neuroexplicit Models of Language, Vision, and Action” - project number 471607914. 
Pablo M. Olmos has been supported by the Ministerio de Ciencia, Innovación y Universidades, Spain, under Grant PID2024-157856NB-I00 (CARTESIAN), funded by MCIN/AEI/10.13039/501100011033 and, as appropriate, by “ERDF A way of making Europe”; by the ELLIS Unit Madrid (European Laboratory for Learning and Intelligent Systems); and by the Comunidad de Madrid under grant IND2024/TIC-34728 and the IDEA-CM project (TEC-2024/COM-89).
Views and opinions
expressed are, however, those of the author(s) only and do not necessarily reflect those of the aforementioned
funding agencies. None of the aforementioned parties can be held responsible for them.

\bibliography{mybibliography}
\bibliographystyle{iclr2027_conference}

\appendix
\section{Stackelberg and Gradient-Descent Ascent}\label{app:theory}

In this section, we provide more details about topics mentioned Section~\ref{sec:methodology}. We strongly encourage readers to refer to the original works~\citep{stackelbergdynamics,lin2020gradient, beck2026linear}.

\paragraph{Bilevel Optimization Applications.}
BLO has been applied to a wide range of tasks in machine learning~\citep{zhang2024introduction}, 
including hyperparameter optimization~\citep{blometalearning}, multi-task learning~\citep{javaloy2022rotograd}, 
neural architecture search~\citep{liu2018darts,tao2024a}, 
robustness~\citep{gupta2023shrink}, 
generalization~\citep{jenni2018deep}, 
and more recently calibration~\citep{bohdal2021meta,sanguin2025exploring}.

Gradient-based approaches to solve the BLO problem usually require calculating the Jacobian of the best-response. Thus, most methods fall back to approximating it via: $T$ gradient updates of the dynamical system, using implicit-differentiation and calculating the Hessian. 
For more details and methods on gradient-based BLO we direct the reader to~\citep{liu2021investigating}. 

\subsection{Definition of Equilibria}
Following the main text, the formulation of a Stackelberg game (or bilevel optimization problem) is:
\begin{equation}
\begin{aligned}
& \min_{\lvar \in \lset, \fvar \in \fsolutions}
	\lopt(\fvar,\lvar) && \text{(Leader)}  \\
& \text{where } \fsolutions= \argmin_{\fvar \in \fset}\fopt(\fvar,\lvar) && \text{(Follower)}
\end{aligned}
\end{equation}
$\fsolutions$ is the solution set of the follower's problem, accounting for the possibility of multiple optimal solutions. It depends on the leader's variables $\lvar$, since the leader's choice determines the follower's optimization problem and, consequently, its optimal solution set. 

An equilibrium can be viewed as a joint solution satisfying the relevant optimality conditions of all players. A Nash equilibrium is a joint strategy profile in which each player's strategy is an optimal response to the others, while a Stackelberg equilibrium is the solution of the corresponding bilevel problem, with the leader accounting for the follower's optimal response.

Following~\citep{stackelbergdynamics}, a solution $\theta^*=(\lvar^*,\fvar^*)$ is considered a differential Nash equilibrium (NSE) when for both players $j$: 
\begin{equation}
    \frac{\partial \eitherlevel (\theta^*)}{\partial {\eithervar}} = 0 \quad \text{and} \quad
    \frac{\partial^2 \eitherlevel (\theta^*)}{\partial {\eithervar}^2} > 0    
\end{equation}
Note that for DNE we require only the diagonal blocks of each player's Hessian to be positive definite, i.e., each player has a local optimum along their own coordinate axis, with no consideration of cross-player effects. 

Meanwhile a solution is considered a differential Stackelberg equilibrium (DSE) if for the leader:
\begin{equation}
    d\lopt(\theta^*) = 0 \quad \text{and} \quad 
    d^2 \lopt(\theta^*) > 0    
\end{equation}  
and for the follower:
\begin{equation}
    \frac{\partial\fopt(\theta^*)}{\partial{\fvar^*}} = 0 \quad \text{and} \quad 
    \frac{\partial^2 \fopt(\theta^*)}{\partial {\fvar^*}^2} > 0    
\end{equation}
Note that in DSE the leader's total derivative is used, capturing the fact that the leader operates under the assumption that the follower plays a (local) best response to $\lvar$, due to the hierarchical nature of the game. \textit{This is the key difference between the two equilibrium concepts}. Specifically, the total derivative accounts for the implicit dependence of the follower's response:

\begin{equation}
    d\lopt(\theta) = \frac{\partial\lopt(\theta)}{\partial{\lvar}} + \frac{\partial\lopt(\theta)}{\partial{\fvar}}\frac{\partial r(\lvar)}{\partial{\lvar}}
\end{equation}
where $r(\lvar)$ is the implicit best-response function of the follower, via the implicit function theorem. 

\paragraph{Subset relationship of DSE and DNE.} In zero-sum games, every Nash equilibrium is also a Stackelberg equilibrium, but the converse does not hold~\citep{bacsar1998dynamic,stackelbergdynamics}. In the continuous setting,~\citet{stackelbergdynamics} show that this inclusion follows from a second-order relationship revealed by the Schur complement of the Stackelberg Jacobian. In particular, the second-order conditions for a DNE $(\frac{\partial^2 \eitherlevel (\theta^*)}{\partial {\eithervar}^2} > 0  )$ imply those for a DSE $(\frac{d^2 \lopt(\theta^*)}{d {\eithervar}^2} > 0 )$ via the Schur complement. 

\paragraph{Implications for \method}
Gradient Descent–Ascent iteratively and jointly updates the strategies of both players via gradient descent.  GDA is designed to target Nash equilibria and can therefore yield valid but restrictive solutions for Stackelberg games. Moreover, GDA does not reliably converge to Nash equilibria due to its characteristic rotational dynamics and may instead converge to spurious stable points, making it theoretically less principled than bilevel optimization (BLO) methods~\citep{stackelbergdynamics}. Nevertheless, GDA remains a practical approach for approximating Stackelberg problems because of its simplicity and computational efficiency, as demonstrated in prior work~\citep{javaloy2022rotograd} and in our experiments.

\begin{table}
    \small
    \caption{Complexity of different calibration methods in Single-Task and Multi-Task learning settings, where \numclasses\ denotes the number of classes.}
    \label{tab:qualitative_comparison_full}
    \centering
    \begin{tabular}{cc|cccc}
    \toprule
    &        & \# Extra           & End & Calibration & Extra  \\
    & \textbf{Method} &  Trainable Params. &  To End                     &  Data    & Burden\\
         \midrule
         \multirow{15}{*}{\rotatebox{90}{\textbf{Single-Task}}}
         &Baseline&  0&  \cmark & \xmark & -\\
         &COG-FC&  0&  \cmark & \xmark & 1 Learning Rate\\
         \cmidrule(lr){2-6}
         & Label Smoothing & 0 & \cmark & \xmark & 1 Smoothing Level \\
         & \multirow{2}{*}{COG+LS}
                & \multirow{2}{*}{0}
                & \multirow{2}{*}{\cmark}
                & \multirow{2}{*}{\xmark}
                & \multirow{2}{*}{\begin{tabular}[c]{@{}l@{}}
                (a) 1 Smoothing Level \\
                (b) 1 Learning Rate\\
                \end{tabular}} \\
                & & & & \\
         \cmidrule(lr){2-6}
         &Temp. Scaling& 1 &  \xmark & \cmark & Calibrator Training\\
         &COG-TS& 1 &  \cmark & \xmark &1 Learning Rate\\
         \cmidrule(lr){2-6}
         &Matrix Scaling&  $\numclasses \times \numclasses + \numclasses$ &  \cmark & \cmark & Calibrator Training\\
         &COG-MS&  $\numclasses \times \numclasses + \numclasses$&  \cmark & \xmark & 1 Learning Rate\\
         \cmidrule(lr){2-6}
         &TST    & NN & \xmark & \xmark & NN Training, Architecture \\
         &COG-TST   & NN & \cmark & \xmark &1 Learning Rate\\
         \cmidrule(lr){2-6}
         & \multirow{3}{*}{V-TST}
                & \multirow{3}{*}{2 NNs}
                & \multirow{3}{*}{\xmark}
                & \multirow{3}{*}{\xmark}
                & \multirow{3}{*}{\begin{tabular}[c]{@{}l@{}}
                (a) ELBO training of 2 NNs \\
                (b) NN architecture\\
                (c) MC Sampling
                \end{tabular}} \\
                & & & & \\
                & & & &\\
                         
         \midrule
         \midrule
         \multirow{8}{*}{\rotatebox{90}{\textbf{Multi-Task}}}
         &Baseline&  0&  \cmark & \xmark & -\\
         &COG-FC    & 0& \cmark & \xmark &1 Learning Rate\\
         \cmidrule(lr){2-6}
         & Label Smoothing & 0 & \cmark & \xmark & \numclasses\ Smoothing Levels \\
         & \multirow{2}{*}{COG+LS}
                & \multirow{2}{*}{0}
                & \multirow{2}{*}{\cmark}
                & \multirow{2}{*}{\xmark}
                & \multirow{2}{*}{\begin{tabular}[c]{@{}l@{}}
                (a) K Smoothing Levels \\
                (b) 1 Learning Rate\\
                \end{tabular}} \\
                & & & & \\
         \cmidrule(lr){2-6}
         &Temp. Scaling    & \numclasses& \xmark & \cmark &\numclasses\ Models Trained\\
         &COG-TS    & \numclasses& \cmark & \xmark &1 Learning Rate\\
    \bottomrule
    \end{tabular}

\end{table}
\section{Implementation Details}\label{app:exp_details}

\subsection{Training Details}
For the backbone considered in our experiments, ResNet-101~\citep{he2016deep}, we trained the models for 300 epochs in the single-task setting and 100 epochs in the multitask setting. All experiments used the Adam optimizer~\citep{kingma2014adam} with a batch size of 256. Unless otherwise stated, we used five random seeds for every experiment reported in the main paper. All experiments used PyTorch~\citep{pytorchlib} version 2.11. 

For the COG experiments, we used a step learning-rate scheduler (as recommended in~
\citep{stackelbergdynamics,kwon2023fully}) that reduced the learning rate by a factor of 10 every 10 epochs for both tasks. For CelebA, the learning rate was reduced by a factor of 10 every 3 epochs. For both single and multi task settings we trained with cross-entropy loss. Specifically, for the MTL setting, we used equal weights across all tasks. 

 All post-hoc calibrators and evaluation metrics are implemented using the Probmetrics library (version 1.2.0)~\citep{berta2025rethinking}, except for Temperature Scaling, for which we use the original implementation~\citep{guo2017calibration}. All post-hoc calibrators were fitted on the validation set. During training, for all models considered in the experiments, we selected the best model checkpoint based on the validation Brier score, as this criterion leads to better-calibrated models~\citep{berta2025rethinking}. For label smoothing, in all cases, we used a smoothing value of $0.1$.

\paragraph{ViT Experiments.} For the ViT ablation we use the ViT architecture seen in~\citep{pablos}. We train for a maximum of 400 epochs, with a early stopping patience of 50 validation checks. The learning rate for the calibrator ($\lrcal$) is set to $1e-5$ and for the backbone's learning rate ($\lrmodel$) is set to $1e-4$.

\paragraph{MTL Experiments.} For the MTL experiment, we trained with a maximum of 100 epochs, and again we picked the best model based on Brier score. For the baselines we used a learning rate of $1e-4$. 

\subsection{Model Selection}

We performed hyperparameter tuning primarily over the learning rates for the two problems in COG. Hyperparameters were selected based on validation CE loss. Unless otherwise stated, all subsequent experiments used the best configurations identified during this tuning procedure (seen in Table~\ref{tab:bestconfigs}). For example, the COG-Label Smoothing experiment on CIFAR-100 with ResNet-101 used the corresponding configuration selected during this procedure.

For the baseline models, we used a learning rate of $10^{-4}$. For TST, we used the default hyperparameter settings reported in~\citep{pablos}.

\begin{table}[t]
\centering
\caption{Hyperparameter configurations selected based on validation performance for the different COG variants. $\lrmodel$ and $\lrcal$ denote the learning rates used for model training and calibration, respectively. A dash indicates that the corresponding component was not used for that method.}
\label{tab:bestconfigs}
\setlength{\tabcolsep}{3pt}
\resizebox{0.7\textwidth}{!}{
\begin{tabular}{cl|cc|cc|cc|cc}
\toprule
 & & \multicolumn{2}{c|}{\textbf{CIFAR-10}}& \multicolumn{2}{c|}{\textbf{CIFAR-100}}& \multicolumn{2}{c|}{\textbf{SVHN}} &\multicolumn{2}{c}{\textbf{CelebA}}\\
 & \textbf{Method} & $\lrmodel$ & $\lrcal$ & $\lrmodel$ & $\lrcal$ & $\lrmodel$ & $\lrcal$  & $\lrmodel$ &$\lrcal$  \\
\midrule
\multirow{4}{*}{\rotatebox{90}{\textbf{ResNet}}}
  &COG-FC & 1e-3& 5e-5 & 1e-3& 1e-5  & 1e-3 & 1e-5  & 1e-3&1e-5  \\
  &COG-TS & 1e-3& 1e-4 & 1e-3 & 1e-5 & 1e-3& 1e-4  & 1e-2&5e-5  \\
  &COG-MS & 1e-3& 5e-5 & 5e-4& 1e-5 & 5e-4& 1e-5  & -&-\\
  &COG-TST & 1e-3& 5e-5 & 5e-4 & 1e-5 & 5e-4& 1e-4  & -&-\\
\midrule
\midrule
  \multirow{4}{*}{\rotatebox{90}{\textbf{WRN}}}
 & COG-FC & 1e-3 & 1e-5 & 1e-2& 1e-5  & 1e-3 & 5e-5  & -&-\\
 & COG-TS & 1e-3 & 1e-5 & 1e-2& 1e-5 & 1e-3 & 1e-4  & -&-\\
 & COG-MS & 1e-2& 5e-5 & 1e-2 & 1e-3 & 5e-4 & 1e-4  & -&-\\
 & COG-TST & 1e-2& 1e-3 & 1e-3& 1e-5 & 1e-3& 5e-4  & -&-\\
\end{tabular}
}
\end{table}

\subsection{Dataset Details}

For CIFAR-10~\citep{krizhevsky2009learning}, CIFAR-100~\citep{krizhevsky2009learning}, and SVHN~\citep{netzer2011reading}, the original training sets were randomly partitioned into training and validation subsets using a fixed random seed. The validation fractions were 20\%, 5\%, and 20\% for CIFAR-10, CIFAR-100, and SVHN, respectively. Training images were augmented using a random $32\times32$ crop with 4-pixel padding, followed by random horizontal flipping. Validation and test images were evaluated without stochastic data augmentation.

For CelebA~\citep{liu2015faceattributes}, we used the predefined training, validation, and test splits. Images were resized and center-cropped to $32\times32$, with random horizontal flipping applied only during training. All datasets were normalized using dataset-specific channel-wise means and standard deviations.

The held-out validation set was used both for model checkpoint selection and for fitting the post-hoc calibration methods. In the multitask setting, all 40 tasks were used.

\subsection{Hardware Requirements} All experiments were conducted on a high-performance computing
cluster with automatic job scheduling, ensuring efficient resource allocation. We list the hardware requirements we used. For all datasets and backbones we utilized 6 CPU cores, 16GB of RAM, and a GPU with at least 8GB of VRAM.

\section{Extended Results}\label{app:extended_results}
In this section, we provide extended versions of the experiments presented in the main text. For all experiments, we additionally report Expected Calibration Error (ECE)~\citep{naeini2015obtaining}, Root Mean Square Calibration Error (RMSCE)~\citep{hendrycks2018deep} and maximum calibration error (MCE)~\citep{naeini2015obtaining}. These calibration metrics are reported as percentages. In cases where NLL is not reported in the main text, we additionally include it here to provide a more comprehensive evaluation.

Standard calibration metrics such as ECE have several well-documented limitations, particularly in the multiclass setting~\citep{minderer2021revisiting,chidambaram2025reassessing}.
The limitations discussed below represent only a subset of those identified in the literature. First, ECE is a biased estimator, whose magnitude of the positive bias increases with model accuracy~\citep{minderer2021revisiting}. Furthermore, its reported value can vary substantially with the choice of the number of bins, making comparisons between models sensitive on the binning scheme~\citep{minderer2021revisiting,silva2023classifier,tao2024a}.
 Moreover, the relationship between ECE and Accuracy appears to hold primarily for highly performant models~\citep{tao2024a} and may depend on the architecture considered~\citep{sharifdeen2026mvc}. Comparisons between in-hoc and post-hoc calibration methods can also be problematic when relying solely on standard calibration metrics~\citep{ashukha2020pitfalls,tao2024a}. Lastly, ECE evaluates calibration only with respect to the top-1 predicted class, and recently it has been shown that even a trivial predictor that always outputs the mean confidence can outperform post-hoc calibration methods according to standard calibration metrics~\citep{chidambaram2025reassessing}.

\subsection{Comparison of COG and Post-hoc approaches}\label{app:cog_vs_post}
Here we provide a more detailed version of the results found in Section~\ref{sec:cog_vs_post}.

\begin{figure}
    \centering
    \includegraphics[width=1\linewidth]{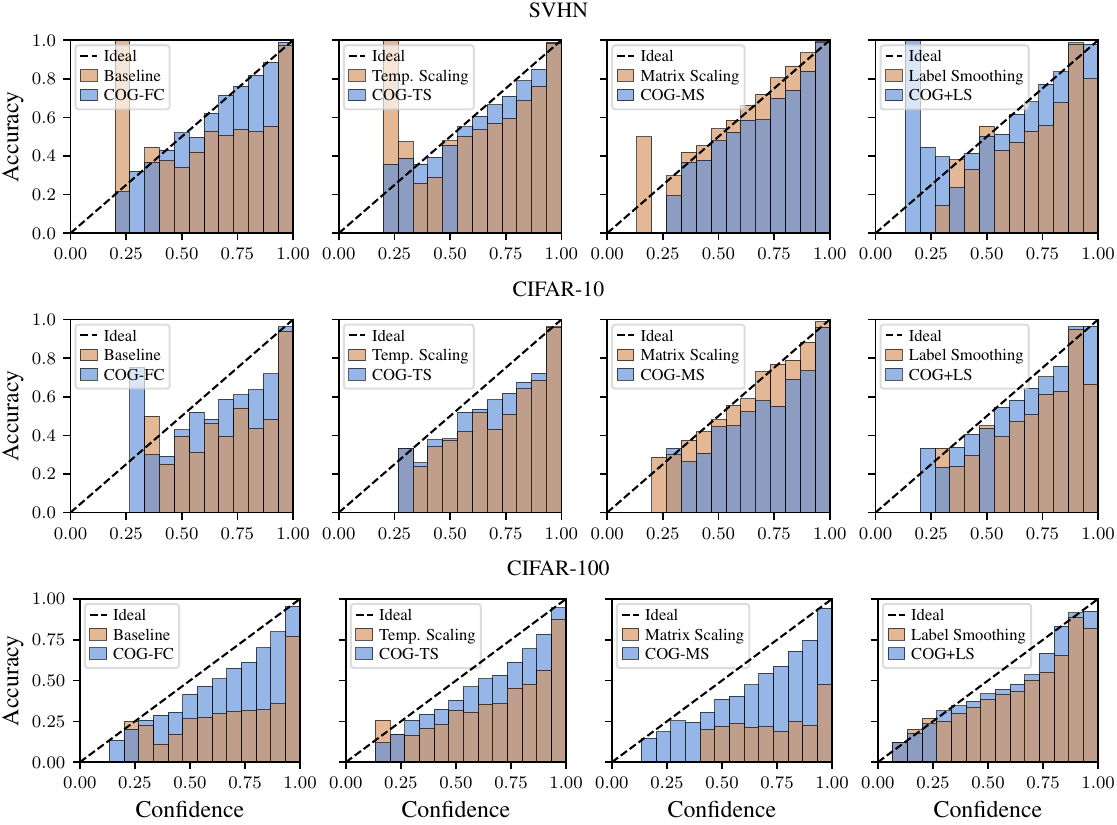}
    \caption{Reliability Diagrams for all \method\ variants and their corresponding calibration methods using 15 bins.}
    \label{fig:reliability_diagram_app}
\end{figure}
Following standard practice~\citep{minderer2021revisiting,silva2023classifier,tao2024a}, we report calibration errors under different norms and across multiple choices of the number of bins. This provides a more comprehensive evaluation and reduces the dependence of our conclusions on any particular choice of calibration metric or binning scheme. 
In Tab.~\ref{tab:cog_vs_post_ece}, we notice that, in almost all pairwise comparisons, the \method\ variants outperform the corresponding post-hoc methods in terms of ECE. A notable exception is Matrix Scaling, which, as also observed in Fig.~\ref{fig:post_vs_cog}, substantially worsens the performance of the Baseline on CIFAR-100. Among all \method\ variants, \method-TST performs the best in ECE overall.
We observe the same qualitative trends for RMSCE, in Tab.~\ref{tab:cog_vs_post_rmsce}, with \method\ generally outperforming the post-hoc calibration methods across these standard calibration metrics.
Finally, we report MCE results in Tab.~\ref{tab:cog_vs_post_mce}. For all calibration metrics, we notice that the relative ranking of methods can change with the number of bins, with the effect being more dramatic for MCE. Furthermore, the reported MCE values can increase dramatically as the number of bins increases. This provides a further illustration of the sensitivity of MCE to the binning scheme and highlights the importance of evaluating calibration across multiple metric configurations rather than relying on a single choice of bins.

\begin{table}[h]
\caption{Comparison of ECE across different \method\ variants and their corresponding post-hoc counterparts for varying numbers of bins. We report $\text{mean}_{\text{std}} \text{ [Rank]}$.}
\label{tab:cog_vs_post_ece}
\centering
\setlength{\tabcolsep}{3pt}
\resizebox{\textwidth}{!}{
\begin{tabular}{l|cccc|cccc|cccc}
\toprule
   & \multicolumn{4}{c|}{\textbf{SVHN}} & \multicolumn{4}{c|}{\textbf{CIFAR-10}} & \multicolumn{4}{c}{\textbf{CIFAR-100}} \\
  \textbf{Method} & ECE-10 & ECE-15 & ECE-100 & ECE-500 & ECE-10 & ECE-15 & ECE-100 & ECE-500 & ECE-10 & ECE-15 & ECE-100 & ECE-500 \\
\midrule
 Baseline & $3.26_{0.11}$ [8] & $3.25_{0.11}$ [8] & $3.29_{0.10}$ [8] & $3.45_{0.10}$ [8] & $7.38_{0.25}$ [8] & $7.39_{0.23}$ [8] & $7.47_{0.23}$ [8] & $7.89_{0.22}$ [8] & $26.56_{0.79}$ [7] & $26.56_{0.78}$ [7] & $26.60_{0.79}$ [7] & $26.76_{0.83}$ [7] \\
  COG-FC & $1.14_{0.30}$ [2] & $1.16_{0.29}$ [2] & $1.32_{0.17}$ [2] & $1.86_{0.07}$ [2] & $5.32_{0.16}$ [4] & $5.32_{0.14}$ [4] & $5.49_{0.11}$ [4] & $6.22_{0.12}$ [5] & $8.00_{0.35}$ [2] & $8.00_{0.36}$ [2] & $8.13_{0.37}$ [2] & $9.20_{0.28}$ [2] \\
  \midrule
 Temp Scaling & $2.19_{0.09}$ [7] & $2.20_{0.09}$ [7] & $2.26_{0.08}$ [7] & $2.57_{0.05}$ [7] & $5.40_{0.29}$ [5] & $5.41_{0.28}$ [5] & $5.57_{0.26}$ [5] & $6.21_{0.14}$ [4] & $18.88_{0.41}$ [6] & $18.88_{0.41}$ [6] & $18.91_{0.42}$ [6] & $19.26_{0.43}$ [6] \\
  COG-TS & $1.23_{0.14}$ [4] & $1.24_{0.14}$ [4] & $1.41_{0.11}$ [4] & $1.95_{0.11}$ [4] & $5.52_{0.40}$ [6] & $5.52_{0.40}$ [6] & $5.66_{0.40}$ [6] & $6.40_{0.40}$ [6] & $11.14_{2.05}$ [3] & $11.14_{2.05}$ [3] & $11.21_{2.03}$ [3] & $11.96_{1.77}$ [3] \\
  \midrule
 Matrix Scaling & $\mathbf{0.71_{0.16}}$ [1] & $\mathbf{0.73_{0.15}}$ [1] & $\mathbf{1.05_{0.07}}$ [1] & $\mathbf{1.84_{0.09}}$ [1] & $\mathbf{0.73_{0.12}}$ [1] & $\mathbf{0.89_{0.10}}$ [1] & $\mathbf{1.76_{0.07}}$ [1] & $\mathbf{3.74_{0.22}}$ [1] & $50.18_{1.79}$ [8] & $50.18_{1.79}$ [8] & $50.19_{1.79}$ [8] & $50.25_{1.79}$ [8] \\
  COG-MS & $1.37_{0.08}$ [5] & $1.38_{0.08}$ [5] & $1.51_{0.06}$ [5] & $1.97_{0.10}$ [5] & $5.67_{0.24}$ [7] & $5.68_{0.24}$ [7] & $5.84_{0.25}$ [7] & $6.61_{0.15}$ [7] & $11.46_{2.24}$ [4] & $11.46_{2.24}$ [4] & $11.55_{2.19}$ [4] & $12.43_{1.82}$ [4] \\
  \midrule
 TST & $1.47_{0.03}$ [6] & $1.48_{0.02}$ [6] & $1.69_{0.05}$ [6] & $2.32_{0.07}$ [6] & $4.90_{0.07}$ [2] & $4.91_{0.05}$ [2] & $5.10_{0.05}$ [2] & $6.01_{0.07}$ [2] & $15.78_{1.46}$ [5] & $15.80_{1.45}$ [5] & $15.82_{1.41}$ [5] & $16.22_{1.40}$ [5] \\
  COG-TST & $1.21_{0.11}$ [3] & $1.21_{0.11}$ [3] & $1.39_{0.09}$ [3] & $1.91_{0.07}$ [3] & $4.96_{0.32}$ [3] & $4.96_{0.32}$ [3] & $5.14_{0.30}$ [3] & $6.13_{0.17}$ [3] & $\mathbf{4.04_{0.55}}$ [1] & $\mathbf{4.07_{0.57}}$ [1] & $\mathbf{4.70_{0.50}}$ [1] & $\mathbf{7.18_{0.35}}$ [1] \\
\bottomrule
\end{tabular}
}
\end{table}

\begin{table}[h]
\caption{Comparison of RMSCE across different \method\ variants and their corresponding post-hoc counterparts for varying numbers of bins. We report $\text{mean}_{\text{std}} \text{ [Rank]}$.}
\label{tab:cog_vs_post_rmsce}
\centering
\setlength{\tabcolsep}{3pt}
\resizebox{\textwidth}{!}{
\begin{tabular}{l|cccc|cccc|cccc}
\toprule
   & \multicolumn{4}{c|}{\textbf{SVHN}} & \multicolumn{4}{c|}{\textbf{CIFAR-10}} & \multicolumn{4}{c}{\textbf{CIFAR-100}} \\
  \textbf{Method} & RMSCE-10 & RMSCE-15 & RMSCE-100 & RMSCE-500 & RMSCE-10 & RMSCE-15 & RMSCE-100 & RMSCE-500 & RMSCE-10 & RMSCE-15 & RMSCE-100 & RMSCE-500 \\
\midrule
 Baseline & $5.03_{0.44}$ [8] & $5.57_{0.28}$ [8] & $7.22_{0.22}$ [8] & $8.85_{0.22}$ [8] & $8.98_{0.43}$ [8] & $9.65_{0.41}$ [8] & $11.89_{0.54}$ [8] & $14.58_{0.46}$ [8] & $27.69_{0.86}$ [7] & $28.27_{0.84}$ [7] & $30.69_{0.83}$ [7] & $32.47_{0.84}$ [7] \\
  COG-FC & $1.96_{0.66}$ [2] & $2.16_{0.67}$ [2] & $3.17_{0.33}$ [2] & $5.59_{0.17}$ [2] & $6.98_{0.36}$ [4] & $7.27_{0.29}$ [4] & $9.08_{0.36}$ [4] & $11.87_{0.34}$ [4] & $9.18_{0.46}$ [2] & $9.39_{0.47}$ [2] & $10.27_{0.51}$ [2] & $12.98_{0.46}$ [2] \\
\midrule
 Temp Scaling & $3.62_{0.27}$ [7] & $3.89_{0.30}$ [7] & $4.93_{0.30}$ [7] & $6.72_{0.29}$ [7] & $7.03_{0.40}$ [6] & $7.58_{0.41}$ [6] & $9.46_{0.46}$ [7] & $12.18_{0.33}$ [6] & $20.57_{0.39}$ [6] & $21.07_{0.44}$ [6] & $22.51_{0.55}$ [6] & $24.14_{0.50}$ [6] \\
  COG-TS & $2.03_{0.35}$ [3] & $2.23_{0.39}$ [4] & $3.33_{0.25}$ [5] & $5.83_{0.35}$ [6] & $6.99_{0.51}$ [5] & $7.49_{0.56}$ [5] & $9.20_{0.65}$ [5] & $11.97_{0.78}$ [5] & $12.34_{2.19}$ [3] & $12.56_{2.28}$ [3] & $13.51_{2.42}$ [3] & $15.85_{2.03}$ [3] \\
\midrule
 Matrix Scaling & $\mathbf{1.26_{0.25}}$ [1] & $\mathbf{1.40_{0.23}}$ [1] & $\mathbf{2.63_{0.16}}$ [1] & $\mathbf{5.34_{0.14}}$ [1] & $\mathbf{1.46_{0.25}}$ [1] & $\mathbf{1.90_{0.34}}$ [1] & $\mathbf{4.15_{0.28}}$ [1] & $\mathbf{8.52_{0.37}}$ [1] & $50.30_{1.79}$ [8] & $50.38_{1.79}$ [8] & $50.91_{1.73}$ [8] & $51.63_{1.64}$ [8] \\
  COG-MS & $2.31_{0.14}$ [6] & $2.53_{0.21}$ [6] & $3.56_{0.13}$ [6] & $5.80_{0.19}$ [3] & $7.17_{0.49}$ [7] & $7.61_{0.60}$ [7] & $9.33_{0.53}$ [6] & $12.19_{0.41}$ [7] & $12.75_{2.35}$ [4] & $13.06_{2.48}$ [4] & $14.00_{2.52}$ [4] & $16.35_{2.28}$ [4] \\
\midrule
 TST & $2.11_{0.07}$ [5] & $2.18_{0.07}$ [3] & $3.29_{0.16}$ [3] & $5.83_{0.19}$ [5] & $6.52_{0.12}$ [2] & $6.79_{0.10}$ [2] & $8.54_{0.14}$ [3] & $11.65_{0.21}$ [3] & $17.07_{1.57}$ [5] & $17.25_{1.63}$ [5] & $17.90_{1.60}$ [5] & $19.61_{1.44}$ [5] \\
  COG-TST & $2.05_{0.28}$ [4] & $2.25_{0.29}$ [5] & $3.33_{0.23}$ [4] & $5.80_{0.18}$ [4] & $6.54_{0.49}$ [3] & $6.81_{0.63}$ [3] & $8.34_{0.54}$ [2] & $11.50_{0.34}$ [2] & $\mathbf{4.59_{0.68}}$ [1] & $\mathbf{4.70_{0.63}}$ [1] & $\mathbf{6.22_{0.74}}$ [1] & $\mathbf{10.20_{0.50}}$ [1] \\
\bottomrule
\end{tabular}
}
\end{table}

\begin{table}[h]
\caption{Comparison of MCE across different \method\ variants and their corresponding post-hoc counterparts for varying numbers of bins. We report $\text{mean}_{\text{std}} \text{ [Rank]}$. The relative ranking of methods changes across binning choices, while the MCE values increase substantially as the number of bins increases, highlighting the sensitivity of MCE to the binning scheme. }
\label{tab:cog_vs_post_mce}
\centering
\setlength{\tabcolsep}{3pt}
\resizebox{\textwidth}{!}{
\begin{tabular}{l|cccc|cccc|cccc}
\toprule
   & \multicolumn{4}{c|}{\textbf{SVHN}} & \multicolumn{4}{c|}{\textbf{CIFAR-10}} & \multicolumn{4}{c}{\textbf{CIFAR-100}} \\
  \textbf{Method} & MCE-10 & MCE-15 & MCE-100 & MCE-500 & MCE-10 & MCE-15 & MCE-100 & MCE-500 & MCE-10 & MCE-15 & MCE-100 & MCE-500 \\
\midrule
 Baseline & $41.31_{18.97}$ [8] & $44.57_{16.42}$ [8] & $67.18_{6.90}$ [6] & $86.85_{2.54}$ [8] & $36.20_{3.99}$ [8] & $38.15_{2.71}$ [7] & $62.04_{7.32}$ [4] & $89.10_{1.47}$ [8] & $48.99_{2.05}$ [7] & $52.19_{1.21}$ [7] & $68.53_{10.04}$ [7] & $83.80_{4.79}$ [7] \\
  COG-FC & $17.27_{6.00}$ [3] & $19.03_{6.81}$ [3] & $48.96_{19.74}$ [2] & $74.04_{1.04}$ [2] & $24.65_{3.82}$ [7] & $28.05_{9.34}$ [5] & $64.03_{12.07}$ [6] & $74.84_{2.75}$ [3] & $15.22_{1.28}$ [2] & $15.74_{1.50}$ [2] & $40.79_{27.35}$ [3] & $72.55_{19.63}$ [2] \\
  \midrule
 Temp Scaling & $20.24_{5.18}$ [4] & $33.59_{25.20}$ [6] & $66.79_{16.02}$ [5] & $78.23_{3.06}$ [5] & $24.28_{3.67}$ [6] & $25.01_{1.70}$ [4] & $59.11_{13.25}$ [3] & $78.52_{4.12}$ [7] & $36.93_{1.15}$ [6] & $37.08_{1.10}$ [6] & $53.48_{17.14}$ [6] & $82.83_{1.14}$ [4] \\
  COG-TS & $33.36_{28.15}$ [7] & $35.96_{26.44}$ [7] & $80.52_{2.18}$ [8] & $80.67_{1.99}$ [7] & $22.90_{4.49}$ [2] & $\mathbf{23.01_{2.66}}$ [1] & $\mathbf{50.63_{17.00}}$ [1] & $\mathbf{74.07_{4.23}}$ [1] & $19.63_{3.45}$ [3] & $20.76_{3.28}$ [3] & $42.74_{22.05}$ [4] & $83.17_{1.57}$ [6] \\
  \midrule
 Matrix Scaling & $30.13_{29.41}$ [6] & $30.16_{29.38}$ [5] & $68.95_{21.59}$ [7] & $78.93_{2.55}$ [6] & $\mathbf{13.89_{8.78}}$ [1] & $23.63_{30.26}$ [2] & $70.17_{14.95}$ [8] & $77.00_{1.27}$ [5] & $59.68_{2.69}$ [8] & $67.51_{3.68}$ [8] & $77.04_{2.03}$ [8] & $92.62_{2.22}$ [8] \\
  COG-MS & $20.63_{8.74}$ [5] & $26.11_{4.82}$ [4] & $\mathbf{43.57_{16.80}}$ [1] & $\mathbf{73.42_{6.72}}$ [1] & $23.96_{3.41}$ [4] & $24.99_{3.38}$ [3] & $52.36_{16.96}$ [2] & $74.15_{4.45}$ [2] & $21.13_{4.24}$ [4] & $22.67_{5.13}$ [4] & $43.17_{22.18}$ [5] & $83.00_{1.05}$ [5] \\
  \midrule
 TST & $\mathbf{10.61_{4.73}}$ [1] & $\mathbf{12.00_{4.10}}$ [1] & $58.87_{27.22}$ [3] & $77.56_{2.47}$ [4] & $23.07_{4.78}$ [3] & $38.30_{22.05}$ [8] & $66.90_{12.15}$ [7] & $77.93_{2.33}$ [6] & $28.67_{6.88}$ [5] & $26.84_{2.65}$ [5] & $38.05_{2.66}$ [2] & $82.39_{14.30}$ [3] \\
  COG-TST & $11.24_{4.32}$ [2] & $15.63_{3.58}$ [2] & $61.39_{23.73}$ [4] & $77.33_{2.09}$ [3] & $24.21_{8.73}$ [5] & $33.00_{24.36}$ [6] & $62.43_{17.53}$ [5] & $75.76_{4.75}$ [4] & $\mathbf{10.79_{3.30}}$ [1] & $\mathbf{9.10_{1.74}}$ [1] & $\mathbf{32.23_{11.60}}$ [1] & $\mathbf{72.45_{24.42}}$ [1] \\
\bottomrule
\end{tabular}
}
\end{table}

\clearpage
\subsection{All methods compared} \label{app:app_everything_compared}
Here we provide a more detailed version of the results found in Section~\ref{sec:compare_families}. 

\paragraph{Metrics used in the main text.} As shown in Tab.~\ref{tab:app_all_VS_all_ID}, \method-FC achieves the highest overall rank across both backbones. Notably, \method-FC achieves this performance without introducing any additional trainable parameters, while being end to end and without a held out calibration set.

\begin{table}[h]
\caption{Comparison of all models and all variants seen in this work. We report $\text{mean}_{\text{std}}$. The best value in each column is shown in bold. \textbf{Rank} denotes the mean rank across ACC, NLL, and Brier for a given dataset (1 = best; ties share a rank).}
\label{tab:app_all_VS_all_ID}
\centering
\setlength{\tabcolsep}{3pt}
\resizebox{\textwidth}{!}{
\begin{tabular}{l|cccc|cccc|cccc}
\toprule
   & \multicolumn{4}{c|}{\textbf{SVHN}} & \multicolumn{4}{c|}{\textbf{CIFAR-10}} & \multicolumn{4}{c}{\textbf{CIFAR-100}} \\
  \textbf{Method} & ACC $(\uparrow)$ & NLL $(\downarrow)$ & Brier $(\downarrow)$ & \textbf{Rank} $(\downarrow)$ & ACC $(\uparrow)$ & NLL $(\downarrow)$ & Brier $(\downarrow)$ & \textbf{Rank} $(\downarrow)$ & ACC $(\uparrow)$ & NLL $(\downarrow)$ & Brier $(\downarrow)$ & \textbf{Rank} $(\downarrow)$ \\
\midrule
 Baseline & $95.77_{0.18}$ & $0.32_{0.01}$ & $0.07_{0.00}$ & 8.67 & $90.50_{0.21}$ & $0.64_{0.02}$ & $0.17_{0.00}$ & 8.00 & $65.04_{1.33}$ & $2.98_{0.06}$ & $0.60_{0.02}$ & 8.33 \\
 \midrule
  COG-FC & $95.84_{0.20}$ & $\mathbf{0.16_{0.00}}$ & $\mathbf{0.06_{0.00}}$ & 2.67 & $89.94_{0.25}$ & $0.38_{0.01}$ & $0.16_{0.00}$ & 5.67 & $70.46_{0.30}$ & $\mathbf{1.08_{0.01}}$ & $\mathbf{0.41_{0.00}}$ & $\color{PineGreen}{\mathbf{1.33}}$ \\
  COG+LS & $\mathbf{96.25_{0.15}}$ & $0.21_{0.00}$ & $0.07_{0.00}$ & 4.67 & $89.92_{0.21}$ & $0.38_{0.00}$ & $0.16_{0.00}$ & 6.00 & $\mathbf{72.67_{0.78}}$ & $1.31_{0.04}$ & $\mathbf{0.41_{0.01}}$ & 2.00 \\
  COG-TS & $95.86_{0.05}$ & $\mathbf{0.16_{0.00}}$ & $\mathbf{0.06_{0.00}}$ & 2.33 & $89.54_{0.15}$ & $0.40_{0.02}$ & $0.16_{0.00}$ & 8.00 & $67.55_{0.62}$ & $1.30_{0.06}$ & $0.46_{0.00}$ & 3.33 \\
  COG-MS & $96.02_{0.11}$ & $\mathbf{0.16_{0.00}}$ & $\mathbf{0.06_{0.00}}$ & $\color{PineGreen}{\mathbf{1.33}}$ & $89.27_{0.26}$ & $0.41_{0.01}$ & $0.17_{0.00}$ & 10.33 & $67.74_{0.56}$ & $1.93_{0.95}$ & $0.46_{0.01}$ & 4.67 \\
  COG-TST & $95.84_{0.10}$ & $\mathbf{0.16_{0.00}}$ & $\mathbf{0.06_{0.00}}$ & 2.67 & $88.73_{0.41}$ & $0.37_{0.01}$ & $0.17_{0.00}$ & 8.33 & $64.77_{0.25}$ & $1.24_{0.01}$ & $0.47_{0.00}$ & 5.00 \\
  \midrule
 LabelSmoothing & $95.97_{0.12}$ & $0.23_{0.00}$ & $0.07_{0.00}$ & 6.33 & $90.05_{0.31}$ & $0.43_{0.01}$ & $0.16_{0.00}$ & 7.33 & $65.76_{0.88}$ & $1.58_{0.04}$ & $0.48_{0.01}$ & 5.33 \\
 \midrule
 Temp Scaling & $95.77_{0.18}$ & $0.19_{0.01}$ & $0.07_{0.00}$ & 6.67 & $90.50_{0.21}$ & $0.38_{0.01}$ & $0.15_{0.00}$ & 3.00 & $65.04_{1.33}$ & $1.78_{0.03}$ & $0.53_{0.02}$ & 6.33 \\
  Matrix Scaling & $95.91_{0.18}$ & $\mathbf{0.16_{0.00}}$ & $\mathbf{0.06_{0.00}}$ & 2.00 & $\mathbf{90.74_{0.17}}$ & $\mathbf{0.28_{0.00}}$ & $\mathbf{0.14_{0.00}}$ & $\color{PineGreen}{\mathbf{1.00}}$ & $46.39_{1.68}$ & $19.10_{2.15}$ & $1.02_{0.03}$ & 12.00 \\
  TST & $94.69_{0.04}$ & $0.21_{0.00}$ & $0.08_{0.00}$ & 9.00 & $90.17_{0.06}$ & $0.37_{0.00}$ & $0.15_{0.00}$ & 3.33 & $52.69_{0.45}$ & $1.99_{0.03}$ & $0.64_{0.00}$ & 10.33 \\
  \midrule
 V-TST (MC=1) & $93.74_{0.15}$ & $0.25_{0.00}$ & $0.10_{0.00}$ & 11.67 & $89.62_{0.08}$ & $0.39_{0.02}$ & $0.16_{0.00}$ & 7.33 & $50.50_{0.39}$ & $1.94_{0.01}$ & $0.63_{0.00}$ & 10.00 \\
  V-TST (MC=10) & $94.66_{0.03}$ & $0.21_{0.00}$ & $0.08_{0.00}$ & 9.33 & $90.31_{0.16}$ & $0.34_{0.01}$ & $0.15_{0.00}$ & 2.67 & $52.71_{0.25}$ & $1.81_{0.01}$ & $0.61_{0.00}$ & 8.33 \\

\bottomrule
\end{tabular}
}
\end{table}

\paragraph{Calibration Errors.}  Here we report calibration errors under different norms and across multiple choices of the number of bins for all models and variants seen in this work (ECE in Tab.~\ref{tab:app_all_VS_all_ECE}, RMSCE in Tab.~\ref{tab:app_all_VS_all_RMSCE}, MCE in Tab.~\ref{tab:app_all_VS_all_MCE}). Here, we again notice the switch in rank between different number of bins used to calculate the calibration metrics.

\begin{table}[h]
\caption{Comparison of ECE ($\downarrow$ lower is better) for all models and all variants seen in this work for multiple choices of the number of bins. We report $\text{mean}_{\text{std}}$. The best value in each column is shown in bold.}
\label{tab:app_all_VS_all_ECE}
\centering
\setlength{\tabcolsep}{3pt}
\resizebox{\textwidth}{!}{
\begin{tabular}{l|cccc|cccc|cccc}
\toprule
   & \multicolumn{4}{c|}{\textbf{SVHN}} & \multicolumn{4}{c|}{\textbf{CIFAR-10}} & \multicolumn{4}{c}{\textbf{CIFAR-100}} \\
  \textbf{Method} & ECE-10 & ECE-15 & ECE-100 & ECE-500 & ECE-10 & ECE-15 & ECE-100 & ECE-500 & ECE-10 & ECE-15 & ECE-100 & ECE-500 \\
\midrule
 Baseline & $3.26_{0.11}$ & $3.25_{0.11}$ & $3.29_{0.10}$ & $3.45_{0.10}$ & $7.38_{0.25}$ & $7.39_{0.23}$ & $7.47_{0.23}$ & $7.89_{0.22}$ & $26.56_{0.79}$ & $26.56_{0.78}$ & $26.60_{0.79}$ & $26.76_{0.83}$ \\
 \midrule
  COG-FC & $1.14_{0.30}$ & $1.16_{0.29}$ & $1.32_{0.17}$ & $1.86_{0.07}$ & $5.32_{0.16}$ & $5.32_{0.14}$ & $5.49_{0.11}$ & $6.22_{0.12}$ & $8.00_{0.35}$ & $8.00_{0.36}$ & $8.13_{0.37}$ & $9.20_{0.28}$ \\
  COG+LS & $7.44_{0.20}$ & $7.51_{0.19}$ & $7.82_{0.17}$ & $8.32_{0.17}$ & $4.90_{0.18}$ & $5.10_{0.31}$ & $5.65_{0.31}$ & $7.17_{0.26}$ & $6.81_{2.10}$ & $6.94_{1.85}$ & $7.95_{1.59}$ & $9.33_{1.07}$ \\
  COG-TS & $1.23_{0.14}$ & $1.24_{0.14}$ & $1.41_{0.11}$ & $1.95_{0.11}$ & $5.52_{0.40}$ & $5.52_{0.40}$ & $5.66_{0.40}$ & $6.40_{0.40}$ & $11.14_{2.05}$ & $11.14_{2.05}$ & $11.21_{2.03}$ & $11.96_{1.77}$ \\
  COG-MS & $1.37_{0.08}$ & $1.38_{0.08}$ & $1.51_{0.06}$ & $1.97_{0.10}$ & $5.67_{0.24}$ & $5.68_{0.24}$ & $5.84_{0.25}$ & $6.61_{0.15}$ & $11.46_{2.24}$ & $11.46_{2.24}$ & $11.55_{2.19}$ & $12.43_{1.82}$ \\
  COG-TST & $1.21_{0.11}$ & $1.21_{0.11}$ & $1.39_{0.09}$ & $1.91_{0.07}$ & $4.96_{0.32}$ & $4.96_{0.32}$ & $5.14_{0.30}$ & $6.13_{0.17}$ & $4.04_{0.55}$ & $4.07_{0.57}$ & $4.70_{0.50}$ & $\mathbf{7.18_{0.35}}$ \\
  \midrule
 Label Smoothing & $6.81_{0.29}$ & $7.13_{0.22}$ & $7.33_{0.22}$ & $7.66_{0.22}$ & $6.37_{0.45}$ & $5.32_{0.30}$ & $6.72_{0.27}$ & $7.59_{0.23}$ & $4.06_{2.05}$ & $4.51_{1.90}$ & $5.31_{1.67}$ & $7.90_{1.22}$ \\
 \midrule
 Temp. Scaling & $2.19_{0.09}$ & $2.20_{0.09}$ & $2.26_{0.08}$ & $2.57_{0.05}$ & $5.40_{0.29}$ & $5.41_{0.28}$ & $5.57_{0.26}$ & $6.21_{0.14}$ & $18.88_{0.41}$ & $18.88_{0.41}$ & $18.91_{0.42}$ & $19.26_{0.43}$ \\
  Matrix Scaling & $\mathbf{0.71_{0.16}}$ & $\mathbf{0.73_{0.15}}$ & $\mathbf{1.05_{0.07}}$ & $\mathbf{1.84_{0.09}}$ & $\mathbf{0.73_{0.12}}$ & $\mathbf{0.89_{0.10}}$ & $\mathbf{1.76_{0.07}}$ & $\mathbf{3.74_{0.22}}$ & $50.18_{1.79}$ & $50.18_{1.79}$ & $50.19_{1.79}$ & $50.25_{1.79}$ \\
  TST & $1.47_{0.03}$ & $1.48_{0.02}$ & $1.69_{0.05}$ & $2.32_{0.07}$ & $4.90_{0.07}$ & $4.91_{0.05}$ & $5.10_{0.05}$ & $6.01_{0.07}$ & $15.78_{1.46}$ & $15.80_{1.45}$ & $15.82_{1.41}$ & $16.22_{1.40}$ \\
  \midrule
 V-TST (MC=1) & $1.53_{0.86}$ & $1.56_{0.82}$ & $1.85_{0.69}$ & $2.60_{0.55}$ & $2.07_{0.83}$ & $2.14_{0.80}$ & $2.82_{0.67}$ & $4.44_{0.34}$ & $6.60_{1.70}$ & $6.61_{1.70}$ & $6.93_{1.64}$ & $9.00_{1.29}$ \\
  V-TST (MC=10) & $3.02_{0.99}$ & $3.02_{0.99}$ & $3.07_{0.97}$ & $3.52_{0.80}$ & $1.09_{0.40}$ & $1.14_{0.36}$ & $1.99_{0.23}$ & $3.92_{0.22}$ & $\mathbf{2.28_{1.28}}$ & $\mathbf{2.33_{1.24}}$ & $\mathbf{3.73_{0.70}}$ & $7.19_{0.63}$ \\
\bottomrule
\end{tabular}
}
\end{table}

\begin{table}[t]
\caption{Comparison of RMSCE ($\downarrow$ lower is better) for all models and all variants seen in this work for multiple choices of the number of bins. We report $\text{mean}_{\text{std}}$. The best value in each column is shown in bold.}
\label{tab:app_all_VS_all_RMSCE}
\centering
\setlength{\tabcolsep}{3pt}
\resizebox{\textwidth}{!}{
\begin{tabular}{l|cccc|cccc|cccc}
\toprule
   & \multicolumn{4}{c|}{\textbf{SVHN}} & \multicolumn{4}{c|}{\textbf{CIFAR-10}} & \multicolumn{4}{c}{\textbf{CIFAR-100}} \\
  \textbf{Method} & RMSCE-10 & RMSCE-15 & RMSCE-100 & RMSCE-500 & RMSCE-10 & RMSCE-15 & RMSCE-100 & RMSCE-500 & RMSCE-10 & RMSCE-15 & RMSCE-100 & RMSCE-500 \\
\midrule
 Baseline & $5.03_{0.44}$ & $5.57_{0.28}$ & $7.22_{0.22}$ & $8.85_{0.22}$ & $8.98_{0.43}$ & $9.65_{0.41}$ & $11.89_{0.54}$ & $14.58_{0.46}$ & $27.69_{0.86}$ & $28.27_{0.84}$ & $30.69_{0.83}$ & $32.47_{0.84}$ \\
 \midrule
  COG-FC & $1.96_{0.66}$ & $2.16_{0.67}$ & $3.17_{0.33}$ & $5.59_{0.17}$ & $6.98_{0.36}$ & $7.27_{0.29}$ & $9.08_{0.36}$ & $11.87_{0.34}$ & $9.18_{0.46}$ & $9.39_{0.47}$ & $10.27_{0.51}$ & $12.98_{0.46}$ \\
  COG+LS & $7.59_{0.17}$ & $7.70_{0.14}$ & $8.10_{0.13}$ & $9.19_{0.14}$ & $5.28_{0.28}$ & $5.42_{0.42}$ & $6.53_{0.37}$ & $9.73_{0.25}$ & $9.65_{2.03}$ & $10.91_{2.71}$ & $11.87_{2.54}$ & $14.68_{1.98}$ \\
  COG-TS & $2.03_{0.35}$ & $2.23_{0.39}$ & $3.33_{0.25}$ & $5.83_{0.35}$ & $6.99_{0.51}$ & $7.49_{0.56}$ & $9.20_{0.65}$ & $11.97_{0.78}$ & $12.34_{2.19}$ & $12.56_{2.28}$ & $13.51_{2.42}$ & $15.85_{2.03}$ \\
  COG-MS & $2.31_{0.14}$ & $2.53_{0.21}$ & $3.56_{0.13}$ & $5.80_{0.19}$ & $7.17_{0.49}$ & $7.61_{0.60}$ & $9.33_{0.53}$ & $12.19_{0.41}$ & $12.75_{2.35}$ & $13.06_{2.48}$ & $14.00_{2.52}$ & $16.35_{2.28}$ \\
  COG-TST & $2.05_{0.28}$ & $2.25_{0.29}$ & $3.33_{0.23}$ & $5.80_{0.18}$ & $6.54_{0.49}$ & $6.81_{0.63}$ & $8.34_{0.54}$ & $11.50_{0.34}$ & $4.59_{0.68}$ & $4.70_{0.63}$ & $6.22_{0.74}$ & $10.20_{0.50}$ \\
  \midrule
 Label Smoothing & $7.05_{0.24}$ & $7.33_{0.28}$ & $7.79_{0.35}$ & $8.91_{0.31}$ & $7.30_{0.70}$ & $7.08_{0.42}$ & $8.84_{0.23}$ & $11.47_{0.30}$ & $5.24_{2.72}$ & $5.71_{3.02}$ & $7.10_{2.67}$ & $11.12_{2.17}$ \\
 \midrule
 Temp. Scaling & $3.62_{0.27}$ & $3.89_{0.30}$ & $4.93_{0.30}$ & $6.72_{0.29}$ & $7.03_{0.40}$ & $7.58_{0.41}$ & $9.46_{0.46}$ & $12.18_{0.33}$ & $20.57_{0.39}$ & $21.07_{0.44}$ & $22.51_{0.55}$ & $24.14_{0.50}$ \\
  Matrix Scaling & $\mathbf{1.26_{0.25}}$ & $\mathbf{1.40_{0.23}}$ & $\mathbf{2.63_{0.16}}$ & $\mathbf{5.34_{0.14}}$ & $\mathbf{1.46_{0.25}}$ & $\mathbf{1.90_{0.34}}$ & $\mathbf{4.15_{0.28}}$ & $\mathbf{8.52_{0.37}}$ & $50.30_{1.79}$ & $50.38_{1.79}$ & $50.91_{1.73}$ & $51.63_{1.64}$ \\
  TST & $2.11_{0.07}$ & $2.18_{0.07}$ & $3.29_{0.16}$ & $5.83_{0.19}$ & $6.52_{0.12}$ & $6.79_{0.10}$ & $8.54_{0.14}$ & $11.65_{0.21}$ & $17.07_{1.57}$ & $17.25_{1.63}$ & $17.90_{1.60}$ & $19.61_{1.44}$ \\
  \midrule
 V-TST (MC=1) & $2.42_{1.15}$ & $2.55_{1.13}$ & $3.50_{0.88}$ & $5.85_{0.60}$ & $2.72_{1.09}$ & $2.99_{0.93}$ & $4.83_{0.79}$ & $8.99_{0.42}$ & $7.43_{1.79}$ & $7.55_{1.86}$ & $8.46_{1.78}$ & $11.94_{1.43}$ \\
  V-TST (MC=10) & $4.37_{1.23}$ & $4.50_{1.27}$ & $5.03_{1.10}$ & $6.75_{0.74}$ & $2.09_{0.32}$ & $2.24_{0.29}$ & $4.27_{0.26}$ & $8.69_{0.26}$ & $\mathbf{2.73_{1.41}}$ & $\mathbf{2.82_{1.42}}$ & $\mathbf{4.93_{0.95}}$ & $\mathbf{9.57_{0.83}}$ \\
\bottomrule
\end{tabular}
}
\end{table}

\begin{table}[t]
\caption{Comparison of MCE ($\downarrow$ lower is better) for all models and all variants seen in this work for multiple choices of the number of bins. We report $\text{mean}_{\text{std}}$. The best value in each column is shown in bold.}
\label{tab:app_all_VS_all_MCE}
\centering
\setlength{\tabcolsep}{3pt}
\resizebox{\textwidth}{!}{
\begin{tabular}{l|cccc|cccc|cccc}
\toprule
   & \multicolumn{4}{c|}{\textbf{SVHN}} & \multicolumn{4}{c|}{\textbf{CIFAR-10}} & \multicolumn{4}{c}{\textbf{CIFAR-100}} \\
  \textbf{Method} & MCE-10 & MCE-15 & MCE-100 & MCE-500 & MCE-10 & MCE-15 & MCE-100 & MCE-500 & MCE-10 & MCE-15 & MCE-100 & MCE-500 \\
\midrule
 Baseline & $41.31_{18.97}$ & $44.57_{16.42}$ & $67.18_{6.90}$ & $86.85_{2.54}$ & $36.20_{3.99}$ & $38.15_{2.71}$ & $62.04_{7.32}$ & $89.10_{1.47}$ & $48.99_{2.05}$ & $52.19_{1.21}$ & $68.53_{10.04}$ & $83.80_{4.79}$ \\
 \midrule
  COG-FC & $17.27_{6.00}$ & $19.03_{6.81}$ & $48.96_{19.74}$ & $74.04_{1.04}$ & $24.65_{3.82}$ & $28.05_{9.34}$ & $64.03_{12.07}$ & $74.84_{2.75}$ & $15.22_{1.28}$ & $15.74_{1.50}$ & $40.79_{27.35}$ & $72.55_{19.63}$ \\
  COG+LS & $26.96_{36.96}$ & $31.07_{34.26}$ & $64.62_{19.09}$ & $82.10_{10.28}$ & $\mathbf{13.53_{4.93}}$ & $\mathbf{17.21_{8.08}}$ & $62.42_{17.53}$ & $74.22_{1.98}$ & $25.15_{4.64}$ & $25.52_{5.26}$ & $42.97_{10.78}$ & $90.75_{7.47}$ \\
  COG-TS & $33.36_{28.15}$ & $35.96_{26.44}$ & $80.52_{2.18}$ & $80.67_{1.99}$ & $22.90_{4.49}$ & $23.01_{2.66}$ & $50.63_{17.00}$ & $\mathbf{74.07_{4.23}}$ & $19.63_{3.45}$ & $20.76_{3.28}$ & $42.74_{22.05}$ & $83.17_{1.57}$ \\
  COG-MS & $20.63_{8.74}$ & $26.11_{4.82}$ & $43.57_{16.80}$ & $\mathbf{73.42_{6.72}}$ & $23.96_{3.41}$ & $24.99_{3.38}$ & $52.36_{16.96}$ & $74.15_{4.45}$ & $21.13_{4.24}$ & $22.67_{5.13}$ & $43.17_{22.18}$ & $83.00_{1.05}$ \\
  COG-TST & $11.24_{4.32}$ & $15.63_{3.58}$ & $61.39_{23.73}$ & $77.33_{2.09}$ & $24.21_{8.73}$ & $33.00_{24.36}$ & $62.43_{17.53}$ & $75.76_{4.75}$ & $10.79_{3.30}$ & $\mathbf{9.10_{1.74}}$ & $32.23_{11.60}$ & $72.45_{24.42}$ \\
  \midrule
 Label Smoothing & $27.25_{8.32}$ & $24.46_{2.61}$ & $64.13_{23.29}$ & $85.41_{13.43}$ & $22.40_{4.72}$ & $28.71_{8.38}$ & $63.19_{12.02}$ & $76.08_{3.09}$ & $12.23_{5.77}$ & $11.66_{6.50}$ & $27.22_{11.83}$ & $74.04_{28.90}$ \\
 \midrule
 Temp. Scaling & $20.24_{5.18}$ & $33.59_{25.20}$ & $66.79_{16.02}$ & $78.23_{3.06}$ & $24.28_{3.67}$ & $25.01_{1.70}$ & $59.11_{13.25}$ & $78.52_{4.12}$ & $36.93_{1.15}$ & $37.08_{1.10}$ & $53.48_{17.14}$ & $82.83_{1.14}$ \\
  Matrix Scaling & $30.13_{29.41}$ & $30.16_{29.38}$ & $68.95_{21.59}$ & $78.93_{2.55}$ & $13.89_{8.78}$ & $23.63_{30.26}$ & $70.17_{14.95}$ & $77.00_{1.27}$ & $59.68_{2.69}$ & $67.51_{3.68}$ & $77.04_{2.03}$ & $92.62_{2.22}$ \\
  TST & $\mathbf{10.61_{4.73}}$ & $\mathbf{12.00_{4.10}}$ & $58.87_{27.22}$ & $77.56_{2.47}$ & $23.07_{4.78}$ & $38.30_{22.05}$ & $66.90_{12.15}$ & $77.93_{2.33}$ & $28.67_{6.88}$ & $26.84_{2.65}$ & $38.05_{2.66}$ & $82.39_{14.30}$ \\
  \midrule
 V-TST (MC=1) & $25.93_{31.22}$ & $25.88_{31.24}$ & $49.77_{29.86}$ & $78.80_{5.05}$ & $30.78_{28.80}$ & $30.78_{28.80}$ & $\mathbf{43.73_{22.23}}$ & $79.42_{2.50}$ & $12.48_{1.81}$ & $16.95_{6.28}$ & $40.86_{31.27}$ & $74.44_{25.93}$ \\
  V-TST (MC=10) & $16.71_{3.28}$ & $17.23_{2.46}$ & $\mathbf{38.25_{8.40}}$ & $79.39_{1.69}$ & $23.52_{7.01}$ & $23.52_{7.01}$ & $48.56_{17.78}$ & $78.79_{2.92}$ & $\mathbf{5.44_{1.98}}$ & $10.73_{9.18}$ & $\mathbf{20.49_{6.35}}$ & $\mathbf{61.71_{29.70}}$ \\
\bottomrule
\end{tabular}
}
\end{table}

\clearpage
\subsection{Performance under Covariate Shifts and OOD}\label{app:shift_ood_extended}

\paragraph{Metrics used in the main text.} Tab.~\ref{tab:shift_ood_results_full_ood} reports the performance of all methods in out-of-distribution detection and Tab.~\ref{tab:shift_results_full} reports their performance under covariate shifts.

\begin{table}[h]
\caption{Performance in out-of-distribution detection or all models and all variants seen in this work.}
\label{tab:shift_ood_results_full_ood}
\centering
\setlength{\tabcolsep}{3pt}
\resizebox{\textwidth}{!}{
\begin{tabular}{l|cc|c|cc|c|cc|c}
\toprule
   & \multicolumn{3}{c|}{\textbf{SVHN}} & \multicolumn{3}{c|}{\textbf{CIFAR-10}} & \multicolumn{3}{c}{\textbf{CIFAR-100}}\\
  \textbf{Method} & AUROC $(\uparrow)$ & FPR95 $(\downarrow)$ & Rank $(\downarrow)$ & AUROC $(\uparrow)$ & FPR95 $(\downarrow)$  & Rank $(\downarrow)$ & AUROC $(\uparrow)$ & FPR95 $(\downarrow)$  & Rank $(\downarrow)$ \\
\midrule
 Baseline & $94.78_{0.24}$ & $34.43_{1.27}$ & 8.00 & $83.97_{2.70}$ & $76.39_{3.20}$ & 7.00 & $71.92_{1.33}$ & $87.67_{1.34}$ & 7.00 \\
 \midrule
  COG-FC & $96.60_{0.41}$ & $21.60_{2.59}$ & 2.00 & $87.00_{1.32}$ & $73.98_{4.03}$ & 3.00 & $68.50_{2.22}$ & $85.83_{1.43}$ & 8.00 \\
  COG+LS & $93.80_{0.61}$ & $22.58_{2.45}$ & 6.50 & $81.93_{3.10}$ & $73.43_{3.18}$ & 6.00 & $65.48_{7.85}$ & $83.71_{3.42}$ & 6.50 \\
  COG-TS & $96.21_{0.35}$ & $24.71_{2.66}$ & 4.50 & $86.04_{0.89}$ & $77.01_{2.21}$ & 6.00 & $67.31_{1.22}$ & $88.67_{0.81}$ & 10.50 \\
  COG-MS & $\mathbf{96.70_{0.28}}$ & $\mathbf{20.36_{2.18}}$ & $\color{PineGreen}{\mathbf{1.00}}$ & $86.20_{0.56}$ & $77.27_{1.74}$ & 6.00 & $69.40_{1.55}$ & $86.22_{1.06}$ & 8.00 \\
  COG-TST & $96.60_{0.09}$ & $22.66_{0.51}$ & 3.00 & $\mathbf{87.41_{0.70}}$ & $71.37_{2.34}$ & $\color{PineGreen}{\mathbf{1.50}}$ & $73.92_{2.72}$ & $\mathbf{79.96_{2.31}}$ & $\color{PineGreen}{\mathbf{1.50}}$ \\
  \midrule
 Label Smoothing & $80.58_{1.66}$ & $42.84_{2.40}$ & 11.50 & $72.17_{6.88}$ & $78.88_{3.09}$ & 11.50 & $\mathbf{74.35_{2.98}}$ & $84.06_{1.86}$ & 2.00 \\
 \midrule
 Temp. Scaling & $95.33_{0.25}$ & $31.14_{1.85}$ & 6.50 & $85.09_{2.43}$ & $75.04_{4.22}$ & 5.50 & $73.23_{1.53}$ & $85.46_{2.00}$ & 3.50 \\
  Matrix Scaling & $96.04_{0.25}$ & $25.31_{1.88}$ & 5.50 & $86.86_{2.06}$ & $\mathbf{70.92_{7.20}}$ & 2.00 & $58.23_{1.55}$ & $93.01_{0.33}$ & 12.00 \\
  TST & $94.85_{0.03}$ & $35.96_{0.32}$ & 8.50 & $84.51_{0.23}$ & $77.53_{0.31}$ & 8.00 & $69.91_{0.41}$ & $85.77_{0.84}$ & 5.50 \\
  \midrule
 V-TST (MC=1) & $92.28_{0.91}$ & $46.49_{3.14}$ & 11.50 & $77.36_{2.70}$ & $80.33_{1.93}$ & 11.50 & $69.59_{0.70}$ & $86.92_{0.41}$ & 8.00 \\
  V-TST (MC=10) & $94.66_{0.47}$ & $35.35_{2.40}$ & 9.00 & $79.23_{2.27}$ & $78.61_{1.96}$ & 10.00 & $70.75_{0.85}$ & $85.78_{0.64}$ & 5.50 \\
\bottomrule
\end{tabular}
}
\end{table}

\begin{table}[h]
\caption{Performance under covariate shifts. Shift metrics are averaged across all considered covariate shifts.}
\label{tab:shift_results_full}
\centering
\setlength{\tabcolsep}{3pt}
\resizebox{\textwidth}{!}{
\begin{tabular}{l|ccc|c|ccc|c|ccc|c}
\toprule
   & \multicolumn{4}{c|}{\textbf{SVHN}} & \multicolumn{4}{c|}{\textbf{CIFAR-10}} & \multicolumn{4}{c}{\textbf{CIFAR-100}}\\
  \textbf{Method} & ACC $(\uparrow)$ & NLL $(\downarrow)$ & Brier $(\downarrow)$ & Rank $(\downarrow)$  & ACC $(\uparrow)$ & NLL $(\downarrow)$ & Brier $(\downarrow)$ & Rank $(\downarrow)$ & ACC $(\uparrow)$ & NLL $(\downarrow)$ & Brier $(\downarrow)$ & Rank $(\downarrow)$\\
\midrule
 Baseline & $47.01_{1.06}$ & $5.29_{0.23}$ & $0.94_{0.02}$ & 8.67 & $74.28_{0.59}$ & $2.03_{0.11}$ & $0.45_{0.01}$ & 6.67 & $45.85_{1.36}$ & $5.70_{0.21}$ & $0.92_{0.02}$ & 7.67 \\
 \midrule
  COG-FC & $45.82_{1.32}$ & $2.43_{0.12}$ & $0.80_{0.02}$ & 6.00 & $68.80_{0.68}$ & $1.57_{0.09}$ & $0.50_{0.01}$ & 9.00 & $44.09_{0.42}$ & $2.88_{0.04}$ & $0.77_{0.01}$ & 4.00 \\
  COG+LS & $46.05_{0.57}$ & $\mathbf{1.91_{0.02}}$ & $0.77_{0.00}$ & 3.00 & $69.31_{0.42}$ & $1.11_{0.02}$ & $0.46_{0.01}$ & 7.33 & $44.92_{0.69}$ & $2.90_{0.11}$ & $0.81_{0.03}$ & 5.67 \\
  COG-TS & $45.59_{0.50}$ & $3.35_{0.12}$ & $0.88_{0.01}$ & 10.33 & $67.98_{0.22}$ & $2.36_{0.17}$ & $0.55_{0.01}$ & 11.33 & $40.99_{0.22}$ & $5.23_{0.63}$ & $0.93_{0.02}$ & 9.33 \\
  COG-MS & $46.38_{0.51}$ & $2.53_{0.10}$ & $0.81_{0.01}$ & 6.00 & $67.67_{0.63}$ & $1.68_{0.11}$ & $0.52_{0.01}$ & 10.67 & $43.63_{0.80}$ & $4.10_{1.23}$ & $0.80_{0.02}$ & 6.67 \\
  COG-TST & $45.68_{0.67}$ & $2.28_{0.04}$ & $0.80_{0.01}$ & 5.67 & $67.42_{0.42}$ & $1.39_{0.03}$ & $0.51_{0.01}$ & 10.00 & $40.81_{0.50}$ & $2.71_{0.03}$ & $0.77_{0.00}$ & 4.33 \\
  \midrule
 Label Smoothing & $46.19_{0.59}$ & $2.19_{0.04}$ & $0.86_{0.01}$ & 6.33 & $74.22_{0.42}$ & $1.03_{0.02}$ & $0.42_{0.01}$ & 4.00 & $\mathbf{47.14_{0.91}}$ & $\mathbf{2.54_{0.10}}$ & $\mathbf{0.71_{0.03}}$ & $\color{PineGreen}{\mathbf{1.00}}$ \\
 \midrule
 Temp. Scaling & $47.01_{1.06}$ & $3.05_{0.12}$ & $0.85_{0.02}$ & 7.00 & $74.28_{0.59}$ & $1.19_{0.05}$ & $0.41_{0.01}$ & 4.00 & $45.85_{1.36}$ & $3.32_{0.10}$ & $0.81_{0.02}$ & 5.67 \\
  Matrix Scaling & $\mathbf{47.27_{0.72}}$ & $2.14_{0.02}$ & $\mathbf{0.75_{0.01}}$ & $\color{PineGreen}{\mathbf{1.67}}$ & $\mathbf{74.35_{0.66}}$ & $\mathbf{0.84_{0.03}}$ & $\mathbf{0.36_{0.01}}$ & $\color{PineGreen}{\mathbf{1.00}}$ & $30.45_{1.07}$ & $29.62_{3.40}$ & $1.33_{0.03}$ & 12.00 \\
  TST & $41.74_{0.11}$ & $2.66_{0.02}$ & $0.84_{0.00}$ & 9.00 & $72.85_{0.21}$ & $1.10_{0.02}$ & $0.42_{0.00}$ & 5.67 & $38.23_{0.33}$ & $3.24_{0.08}$ & $0.84_{0.01}$ & 8.67 \\
  \midrule
 V-TST (MC=1) & $40.31_{0.26}$ & $2.38_{0.16}$ & $0.83_{0.02}$ & 8.33 & $72.90_{0.24}$ & $1.03_{0.06}$ & $0.41_{0.01}$ & 4.00 & $36.77_{0.26}$ & $2.83_{0.04}$ & $0.80_{0.01}$ & 6.67 \\
  V-TST (MC=10) & $41.66_{0.26}$ & $2.06_{0.07}$ & $0.77_{0.01}$ & 5.00 & $73.91_{0.22}$ & $0.91_{0.03}$ & $0.38_{0.01}$ & 3.00 & $38.44_{0.24}$ & $2.66_{0.02}$ & $0.77_{0.00}$ & 4.33 \\
\bottomrule
\end{tabular}
}
\end{table}

\paragraph{Calibration Errors.}  Here, we additionally report the results for the different calibration error metrics. In particular, V-TST performs particularly strongly according to the considered calibration errors. \method-FC achieves competitive performance across the different metrics and binning configurations.

\begin{table}[h]
\caption{ECE averaged across all considered covariate shifts for all models and all variants seen in this work for
multiple choices of the number of bins.}
\label{tab:app_shift_ood_ece}
\centering
\setlength{\tabcolsep}{3pt}
\resizebox{\textwidth}{!}{
\begin{tabular}{l|cccc|cccc|cccc}
\toprule
   & \multicolumn{4}{c|}{\textbf{SVHN}} & \multicolumn{4}{c|}{\textbf{CIFAR-10}} & \multicolumn{4}{c}{\textbf{CIFAR-100}}\\
  \textbf{Method} & ECE-10 & ECE-15 & ECE-100 & ECE-500 & ECE-10 & ECE-15 & ECE-100 & ECE-500 & ECE-10 & ECE-15 & ECE-100 & ECE-500 \\
\midrule
 Baseline & $43.81_{1.01}$ & $43.81_{1.01}$ & $43.81_{1.01}$ & $43.82_{1.01}$ & $20.68_{0.56}$ & $20.68_{0.56}$ & $20.68_{0.56}$ & $20.66_{0.56}$ & $41.90_{0.89}$ & $41.89_{0.89}$ & $41.89_{0.89}$ & $41.86_{0.89}$ \\
 \midrule
  COG-FC & $29.55_{2.32}$ & $29.55_{2.32}$ & $29.55_{2.32}$ & $29.56_{2.31}$ & $20.46_{0.86}$ & $20.45_{0.86}$ & $20.45_{0.86}$ & $20.43_{0.86}$ & $22.93_{0.19}$ & $22.93_{0.19}$ & $22.93_{0.19}$ & $22.93_{0.19}$ \\
  COG+LS & $24.90_{0.61}$ & $24.90_{0.61}$ & $24.91_{0.61}$ & $24.91_{0.61}$ & $11.96_{1.14}$ & $11.96_{1.14}$ & $11.99_{1.10}$ & $12.00_{1.09}$ & $25.93_{3.92}$ & $25.93_{3.92}$ & $25.93_{3.92}$ & $25.93_{3.92}$ \\
  COG-TS & $37.76_{0.75}$ & $37.76_{0.75}$ & $37.76_{0.75}$ & $37.77_{0.76}$ & $24.80_{0.63}$ & $24.80_{0.63}$ & $24.80_{0.63}$ & $24.78_{0.63}$ & $38.37_{2.30}$ & $38.37_{2.30}$ & $38.36_{2.29}$ & $38.35_{2.29}$ \\
  COG-MS & $30.83_{1.39}$ & $30.83_{1.39}$ & $30.83_{1.39}$ & $30.83_{1.39}$ & $21.30_{0.92}$ & $21.29_{0.92}$ & $21.29_{0.92}$ & $21.27_{0.92}$ & $26.76_{2.95}$ & $26.76_{2.95}$ & $26.76_{2.95}$ & $26.75_{2.95}$ \\
  COG-TST & $29.10_{0.97}$ & $29.10_{0.97}$ & $29.10_{0.97}$ & $29.11_{0.97}$ & $19.22_{0.40}$ & $19.22_{0.40}$ & $19.21_{0.40}$ & $19.20_{0.40}$ & $16.43_{0.47}$ & $16.43_{0.47}$ & $16.43_{0.46}$ & $16.43_{0.46}$ \\
  \midrule
 Label Smoothing & $34.60_{0.89}$ & $34.60_{0.89}$ & $34.60_{0.89}$ & $34.61_{0.89}$ & $11.39_{0.40}$ & $11.39_{0.40}$ & $11.71_{0.29}$ & $11.75_{0.35}$ & $11.14_{5.10}$ & $11.16_{5.09}$ & $11.17_{5.08}$ & $11.19_{5.07}$ \\
 \midrule
 Temp. Scaling & $35.90_{1.13}$ & $35.89_{1.13}$ & $35.90_{1.13}$ & $35.90_{1.13}$ & $16.24_{0.65}$ & $16.23_{0.65}$ & $16.23_{0.65}$ & $16.21_{0.65}$ & $30.94_{0.89}$ & $30.93_{0.89}$ & $30.92_{0.89}$ & $30.92_{0.89}$ \\
  Matrix Scaling & $23.93_{0.87}$ & $23.93_{0.87}$ & $23.94_{0.87}$ & $23.95_{0.87}$ & $\mathbf{7.61_{0.98}}$ & $\mathbf{7.60_{0.98}}$ & $\mathbf{7.59_{0.98}}$ & $\mathbf{7.59_{0.99}}$ & $65.41_{1.37}$ & $65.40_{1.37}$ & $65.39_{1.37}$ & $65.37_{1.37}$ \\
  TST & $31.22_{0.17}$ & $31.22_{0.17}$ & $31.22_{0.17}$ & $31.23_{0.17}$ & $15.62_{0.22}$ & $15.62_{0.22}$ & $15.61_{0.22}$ & $15.60_{0.22}$ & $25.33_{1.60}$ & $25.33_{1.60}$ & $25.33_{1.60}$ & $25.33_{1.60}$ \\
  \midrule
 V-TST (MC=1) & $27.01_{3.15}$ & $27.01_{3.15}$ & $27.01_{3.15}$ & $27.01_{3.15}$ & $11.11_{2.15}$ & $11.10_{2.15}$ & $11.08_{2.14}$ & $11.08_{2.13}$ & $14.11_{1.78}$ & $14.11_{1.78}$ & $14.10_{1.78}$ & $14.11_{1.78}$ \\
  V-TST (MC=10) & $\mathbf{20.52_{2.34}}$ & $\mathbf{20.52_{2.34}}$ & $\mathbf{20.52_{2.34}}$ & $\mathbf{20.54_{2.34}}$ & $8.26_{2.01}$ & $8.25_{2.01}$ & $8.24_{2.00}$ & $8.26_{1.98}$ & $\mathbf{9.57_{1.59}}$ & $\mathbf{9.57_{1.59}}$ & $\mathbf{9.57_{1.59}}$ & $\mathbf{9.57_{1.59}}$ \\
\bottomrule
\end{tabular}
}
\end{table}

\begin{table}[h]
\caption{RMSCE averaged across all considered covariate shifts for all models and all variants seen in this work for multiple choices of the number of bins.}
\label{tab:app_shift_ood_rmsce}
\centering
\setlength{\tabcolsep}{3pt}
\resizebox{\textwidth}{!}{
\begin{tabular}{l|cccc|cccc|cccc}
\toprule
   & \multicolumn{4}{c|}{\textbf{SVHN}} & \multicolumn{4}{c|}{\textbf{CIFAR-10}} & \multicolumn{4}{c}{\textbf{CIFAR-100}}\\
  \textbf{Method} & RMSCE-10 & RMSCE-15 & RMSCE-100 & RMSCE-500 & RMSCE-10 & RMSCE-15 & RMSCE-100 & RMSCE-500 & RMSCE-10 & RMSCE-15 & RMSCE-100 & RMSCE-500 \\
\midrule
 Baseline & $44.25_{1.00}$ & $44.43_{0.99}$ & $45.42_{0.97}$ & $46.12_{0.97}$ & $21.71_{0.49}$ & $22.21_{0.47}$ & $24.29_{0.40}$ & $25.44_{0.39}$ & $42.51_{0.86}$ & $42.80_{0.84}$ & $44.18_{0.77}$ & $44.98_{0.75}$ \\
 \midrule
  COG-FC & $30.73_{2.20}$ & $31.00_{2.21}$ & $31.68_{2.37}$ & $31.82_{2.39}$ & $21.66_{0.79}$ & $22.17_{0.78}$ & $23.72_{0.86}$ & $24.20_{0.92}$ & $24.08_{0.18}$ & $24.29_{0.18}$ & $24.63_{0.17}$ & $24.68_{0.17}$ \\
  COG+LS & $26.50_{0.73}$ & $26.71_{0.64}$ & $27.18_{0.66}$ & $27.31_{0.66}$ & $14.31_{0.88}$ & $14.26_{0.71}$ & $14.75_{0.80}$ & $14.81_{0.81}$ & $27.11_{3.86}$ & $27.88_{4.18}$ & $28.26_{4.31}$ & $28.29_{4.32}$ \\
  COG-TS & $38.64_{0.79}$ & $38.92_{0.80}$ & $40.05_{0.89}$ & $40.58_{0.92}$ & $25.64_{0.56}$ & $26.09_{0.53}$ & $27.94_{0.53}$ & $28.86_{0.58}$ & $39.19_{2.20}$ & $39.48_{2.19}$ & $40.50_{2.26}$ & $40.87_{2.38}$ \\
  COG-MS & $31.92_{1.35}$ & $32.18_{1.37}$ & $32.90_{1.42}$ & $33.04_{1.43}$ & $22.41_{0.86}$ & $22.90_{0.84}$ & $24.41_{0.88}$ & $24.89_{0.92}$ & $27.80_{2.91}$ & $28.04_{2.97}$ & $28.52_{3.15}$ & $28.60_{3.18}$ \\
  COG-TST & $30.26_{1.02}$ & $30.51_{1.05}$ & $31.13_{1.12}$ & $31.24_{1.13}$ & $20.40_{0.37}$ & $20.85_{0.37}$ & $22.05_{0.35}$ & $22.34_{0.37}$ & $17.32_{0.53}$ & $17.41_{0.53}$ & $17.51_{0.54}$ & $17.53_{0.54}$ \\
  \midrule
 Label Smoothing & $36.06_{1.09}$ & $35.99_{0.77}$ & $36.87_{0.93}$ & $37.05_{0.96}$ & $15.57_{1.06}$ & $14.59_{0.19}$ & $16.33_{0.68}$ & $16.59_{0.65}$ & $12.91_{5.21}$ & $13.15_{5.35}$ & $13.25_{5.40}$ & $13.29_{5.39}$ \\
 \midrule
 Temp. Scaling & $36.74_{1.09}$ & $37.00_{1.09}$ & $37.90_{1.14}$ & $38.22_{1.16}$ & $17.72_{0.53}$ & $18.28_{0.53}$ & $19.90_{0.59}$ & $20.38_{0.64}$ & $32.07_{0.86}$ & $32.43_{0.86}$ & $33.49_{0.90}$ & $33.77_{0.93}$ \\
  Matrix Scaling & $24.91_{0.88}$ & $25.03_{0.90}$ & $25.19_{0.92}$ & $25.31_{0.92}$ & $\mathbf{8.67_{1.11}}$ & $\mathbf{8.84_{1.15}}$ & $\mathbf{9.07_{1.21}}$ & $\mathbf{9.13_{1.20}}$ & $65.64_{1.36}$ & $65.65_{1.36}$ & $65.79_{1.33}$ & $65.93_{1.30}$ \\
  TST & $32.63_{0.17}$ & $32.92_{0.17}$ & $33.60_{0.18}$ & $33.76_{0.18}$ & $17.19_{0.21}$ & $17.64_{0.20}$ & $18.96_{0.22}$ & $19.32_{0.22}$ & $26.45_{1.63}$ & $26.63_{1.66}$ & $26.92_{1.73}$ & $26.96_{1.74}$ \\
  \midrule
 V-TST (MC=1) & $28.31_{3.17}$ & $28.47_{3.23}$ & $28.66_{3.35}$ & $28.76_{3.35}$ & $12.05_{2.18}$ & $12.26_{2.21}$ & $12.58_{2.29}$ & $12.65_{2.29}$ & $15.30_{1.87}$ & $15.36_{1.88}$ & $15.43_{1.90}$ & $15.46_{1.90}$ \\
  V-TST (MC=10) & $\mathbf{21.70_{2.47}}$ & $\mathbf{21.80_{2.53}}$ & $\mathbf{21.91_{2.55}}$ & $\mathbf{22.05_{2.54}}$ & $9.25_{2.14}$ & $9.42_{2.19}$ & $9.61_{2.23}$ & $9.68_{2.22}$ & $\mathbf{10.39_{1.68}}$ & $\mathbf{10.41_{1.69}}$ & $\mathbf{10.45_{1.70}}$ & $\mathbf{10.49_{1.69}}$ \\
\bottomrule
\end{tabular}
}
\end{table}

\begin{table}[h]
\caption{MCE averaged across all considered covariate shifts for all models and all variants seen in this work for multiple choices of the number of bins.}
\label{tab:app_shift_ood_mce}
\centering
\setlength{\tabcolsep}{3pt}
\resizebox{\textwidth}{!}{
\begin{tabular}{l|cccc|cccc|cccc}
\toprule
   & \multicolumn{4}{c|}{\textbf{SVHN}} & \multicolumn{4}{c|}{\textbf{CIFAR-10}} & \multicolumn{4}{c}{\textbf{CIFAR-100}}\\
  \textbf{Method} & MCE-10 & MCE-15 & MCE-100 & MCE-500 & MCE-10 & MCE-15 & MCE-100 & MCE-500 & MCE-10 & MCE-15 & MCE-100 & MCE-500 \\
\midrule
 Baseline & $58.64_{1.18}$ & $62.43_{0.81}$ & $72.23_{7.88}$ & $76.60_{4.84}$ & $44.30_{0.51}$ & $46.83_{0.37}$ & $54.91_{13.78}$ & $65.07_{13.82}$ & $59.43_{0.36}$ & $62.38_{0.52}$ & $64.12_{0.60}$ & $72.86_{10.41}$ \\
 \midrule
  COG-FC & $45.33_{2.35}$ & $46.36_{2.83}$ & $47.61_{2.88}$ & $58.79_{14.22}$ & $37.89_{1.23}$ & $38.70_{1.44}$ & $47.84_{18.37}$ & $71.61_{17.00}$ & $33.81_{0.21}$ & $34.23_{0.21}$ & $40.17_{10.67}$ & $49.31_{24.20}$ \\
  COG+LS & $39.56_{1.40}$ & $58.27_{19.86}$ & $97.99_{1.03}$ & $98.13_{1.02}$ & $25.85_{1.53}$ & $26.18_{1.67}$ & $44.33_{24.12}$ & $60.95_{22.11}$ & $41.05_{5.79}$ & $46.22_{9.70}$ & $77.67_{15.92}$ & $92.00_{16.61}$ \\
  COG-TS & $54.80_{1.62}$ & $57.11_{1.40}$ & $59.03_{1.37}$ & $74.21_{11.88}$ & $52.15_{16.14}$ & $54.01_{15.10}$ & $61.46_{16.53}$ & $77.84_{2.28}$ & $53.65_{2.26}$ & $55.01_{2.77}$ & $55.75_{3.00}$ & $57.57_{2.84}$ \\
  COG-MS & $46.55_{1.60}$ & $47.67_{1.92}$ & $49.19_{1.52}$ & $53.37_{1.47}$ & $38.17_{1.23}$ & $39.19_{1.23}$ & $57.12_{21.10}$ & $72.99_{16.32}$ & $39.09_{4.42}$ & $39.43_{4.28}$ & $40.26_{4.15}$ & $50.85_{21.98}$ \\
  COG-TST & $44.42_{1.99}$ & $45.04_{2.12}$ & $62.32_{22.12}$ & $67.51_{17.97}$ & $34.27_{0.97}$ & $34.33_{1.02}$ & $44.96_{22.03}$ & $64.10_{24.58}$ & $24.25_{0.73}$ & $24.30_{0.70}$ & $\mathbf{25.80_{1.56}}$ & $\mathbf{27.57_{0.88}}$ \\
  \midrule
 Label Smoothing & $50.06_{1.22}$ & $62.35_{9.76}$ & $96.58_{2.72}$ & $97.83_{0.90}$ & $32.35_{0.51}$ & $36.56_{2.03}$ & $45.74_{19.46}$ & $76.54_{18.34}$ & $\mathbf{22.37_{8.25}}$ & $\mathbf{22.65_{8.22}}$ & $27.67_{9.58}$ & $39.98_{16.23}$ \\
 \midrule
 Temp. Scaling & $51.29_{1.05}$ & $53.13_{0.83}$ & $67.01_{15.78}$ & $73.50_{13.01}$ & $34.41_{0.84}$ & $35.20_{1.16}$ & $36.15_{1.07}$ & $55.45_{22.68}$ & $47.34_{1.16}$ & $47.59_{1.47}$ & $48.55_{1.63}$ & $66.84_{21.99}$ \\
  Matrix Scaling & $34.46_{1.40}$ & $35.03_{1.52}$ & $37.08_{0.85}$ & $67.03_{23.03}$ & $\mathbf{14.27_{2.12}}$ & $\mathbf{14.53_{2.17}}$ & $\mathbf{16.15_{1.67}}$ & $60.54_{31.64}$ & $69.92_{0.37}$ & $74.36_{0.41}$ & $81.05_{0.81}$ & $81.73_{0.81}$ \\
  TST & $48.38_{0.38}$ & $48.50_{0.37}$ & $50.88_{0.85}$ & $67.24_{15.34}$ & $31.46_{0.21}$ & $31.93_{0.21}$ & $48.22_{23.13}$ & $74.67_{19.58}$ & $50.45_{27.73}$ & $50.77_{27.54}$ & $51.76_{27.02}$ & $63.19_{30.27}$ \\
  \midrule
 V-TST (MC=1) & $40.53_{4.82}$ & $41.00_{4.68}$ & $43.94_{4.53}$ & $54.19_{16.27}$ & $20.08_{3.81}$ & $20.61_{3.87}$ & $25.79_{5.80}$ & $\mathbf{52.55_{21.56}}$ & $52.92_{42.98}$ & $52.94_{42.97}$ & $53.62_{42.34}$ & $55.82_{40.36}$ \\
  V-TST (MC=10) & $\mathbf{31.87_{3.70}}$ & $\mathbf{32.27_{3.89}}$ & $\mathbf{34.24_{3.97}}$ & $\mathbf{42.04_{4.61}}$ & $16.02_{3.71}$ & $16.28_{3.82}$ & $19.85_{3.89}$ & $71.49_{18.33}$ & $31.53_{38.30}$ & $31.73_{38.19}$ & $32.29_{37.87}$ & $40.41_{35.32}$ \\
\bottomrule
\end{tabular}
}
\end{table}

\clearpage

\subsection{ViT ablation}\label{sec:app_vit}
In this subsection, we report calibration metrics for ViT on CIFAR-10. We train the ViT model from scratch using the architecture described in~\citep{pablos}. The results are reported in Tab.~\ref{tab:app_VIT_calibration_metrics}.

\begin{table}[h]
\caption{Comparison of calibration errors for ViT in CIFAR-10 for multiple choices of the number of bins.}
\label{tab:app_VIT_calibration_metrics}
\centering
\setlength{\tabcolsep}{3pt}
\resizebox{\textwidth}{!}{
\begin{tabular}{l|cccc|cccc|cccc}
\toprule
  \textbf{Method} & ECE-10 & ECE-15 & ECE-100 & ECE-500 & RMSCE-10 & RMSCE-15 & RMSCE-100 & RMSCE-500 & MCE-10 & MCE-15 & MCE-100 & MCE-500 \\
\midrule
 Baseline & $7.73_{0.66}$ [4] & $7.73_{0.66}$ [4] & $7.90_{0.66}$ [4] & $9.18_{0.77}$ [4] & $8.49_{0.79}$ [3] & $8.77_{0.86}$ [4] & $9.67_{0.89}$ [4] & $12.46_{0.87}$ [3] & $15.84_{3.92}$ [2] & $15.94_{3.67}$ [2] & $41.11_{21.89}$ [2] & $76.03_{1.74}$ [3] \\
  COG-FC & $7.70_{0.70}$ [3] & $7.70_{0.67}$ [3] & $7.87_{0.61}$ [3] & $8.99_{0.61}$ [3] & $8.61_{0.73}$ [5] & $8.88_{0.75}$ [5] & $9.75_{0.72}$ [5] & $12.52_{0.68}$ [4] & $28.76_{28.91}$ [6] & $29.45_{28.51}$ [6] & $59.26_{23.83}$ [5] & $77.84_{1.67}$ [5] \\
  \midrule
 Temp Scaling & $1.65_{0.34}$ [2] & $1.72_{0.32}$ [2] & $3.11_{0.24}$ [2] & $6.19_{0.13}$ [2] & $1.96_{0.30}$ [2] & $2.16_{0.33}$ [2] & $4.28_{0.36}$ [2] & $8.85_{0.11}$ [2] & $21.00_{5.63}$ [4] & $21.00_{5.63}$ [4] & $\mathbf{40.90_{28.62}}$ [1] & $78.78_{2.61}$ [6] \\
  COG-TS & $8.09_{1.46}$ [6] & $8.11_{1.45}$ [6] & $8.28_{1.40}$ [6] & $9.50_{1.06}$ [6] & $8.89_{1.50}$ [6] & $9.08_{1.57}$ [6] & $10.00_{1.56}$ [6] & $12.85_{1.24}$ [6] & $\mathbf{14.22_{2.15}}$ [1] & $\mathbf{15.73_{2.69}}$ [1] & $58.58_{25.29}$ [4] & $75.97_{2.26}$ [2] \\
  \midrule
 Matrix Scaling & $\mathbf{1.00_{0.13}}$ [1] & $\mathbf{1.22_{0.21}}$ [1] & $\mathbf{2.72_{0.38}}$ [1] & $\mathbf{5.87_{0.22}}$ [1] & $\mathbf{1.44_{0.19}}$ [1] & $\mathbf{1.82_{0.32}}$ [1] & $\mathbf{4.02_{0.46}}$ [1] & $\mathbf{8.68_{0.08}}$ [1] & $20.99_{10.35}$ [3] & $20.99_{10.35}$ [3] & $61.38_{28.21}$ [6] & $\mathbf{75.84_{12.01}}$ [1] \\
  COG-MS & $7.74_{0.54}$ [5] & $7.76_{0.53}$ [5] & $7.91_{0.54}$ [5] & $9.25_{0.40}$ [5] & $8.53_{0.54}$ [4] & $8.76_{0.62}$ [3] & $9.63_{0.59}$ [3] & $12.53_{0.49}$ [5] & $27.80_{29.72}$ [5] & $28.35_{29.50}$ [5] & $56.64_{22.11}$ [3] & $77.62_{2.36}$ [4] \\
\bottomrule
\end{tabular}
}
\end{table}

\subsection{Multi-Task Learning}\label{app:MTL_extended}
Since to the best of our knowledge, there are no MTL-specific calibration error metrics, we follow~\citep{vandenhirtz2024stochastic,van2026neurosymbolic} and measure the overall calibration error as the unweighted average of the calibration errors computed independently for each task, across all norms. As shown in Tab.~\ref{tab:appendix_mtl_calib_errors}, \method-FC and \method+LS consistently improve upon their counterparts, while \method-TS performs worse than Temperature Scaling.

\begin{figure}[t]
    \centering
    \includegraphics[width=1\linewidth]{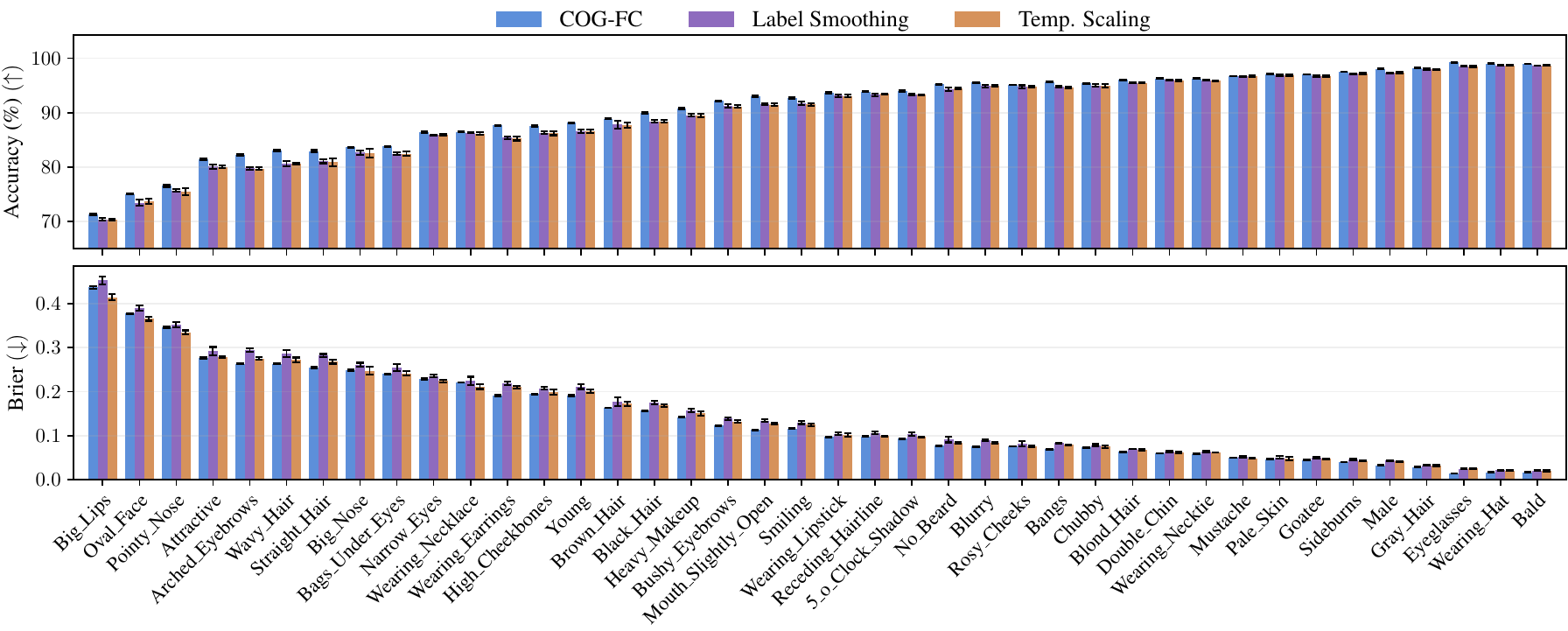}
    \caption{Per-task metrics on CelebA. Tasks are ordered by their mean Brier. \method-FC achieves higher accuracy than Temperature Scaling and Label Smoothing on all 40 tasks. In terms of Brier, \method-FC outperforms Temperature Scaling on 32 of 40 tasks and Label Smoothing on all 40 tasks.}
    \label{fig:per_task_metrics_main}
\end{figure}

\begin{table}[h]
\centering
\caption{Calibration errors for ResNet101 on CelebA, averaged across tasks. Bold indicates the best value within each group.}
\label{tab:appendix_mtl_calib_errors}
\setlength{\tabcolsep}{3pt}
\resizebox{\textwidth}{!}{
\begin{tabular}{l|cccc|cccc|cccc}
\toprule
 &  \multicolumn{12}{c}{\textbf{CelebA}} \\
 \textbf{Method} & ECE-10 & ECE-15 & ECE-100 & ECE-500 & RMSCE-10 & RMSCE-15 & RMSCE-100 & RMSCE-500 & MCE-10 & MCE-15 & MCE-100 & MCE-500 \\
\midrule
 Baseline & $5.84_{0.08}$ & $5.85_{0.08}$ & $6.01_{0.08}$ & $6.56_{0.08}$ & $7.52_{0.16}$ & $7.83_{0.14}$ & $9.29_{0.11}$ & $11.47_{0.10}$ & $26.89_{0.74}$ & $29.35_{0.69}$ & $49.93_{0.46}$ & $82.51_{1.17}$  \\
 COG-FC & $\mathbf{5.01}_{0.04}$ & $\mathbf{5.03}_{0.03}$ & $\mathbf{5.19}_{0.04}$ & $\mathbf{5.77}_{0.03}$ & $\mathbf{6.50}_{0.07}$ & $\mathbf{6.82}_{0.06}$ & $\mathbf{8.27}_{0.06}$ & $\mathbf{10.56}_{0.05}$ & $\mathbf{23.38}_{0.86}$ & $\mathbf{25.50}_{0.54}$ & $\mathbf{48.25}_{0.45}$ & $\mathbf{81.91}_{0.65}$  \\
\cmidrule(lr){1-13}
 Label Smoothing & $4.88_{0.05}$ & $4.90_{0.06}$ & $5.09_{0.05}$ & $5.76_{0.05}$ & $6.59_{0.11}$ & $6.92_{0.11}$ & $8.26_{0.10}$ & $10.45_{0.07}$ & $23.80_{0.65}$ & $25.46_{0.77}$ & $46.98_{1.62}$ & $80.59_{0.41}$  \\
 COG+LS & $\mathbf{4.09}_{0.04}$ & $\mathbf{4.11}_{0.05}$ & $\mathbf{4.33}_{0.04}$ & $\mathbf{5.03}_{0.04}$ & $\mathbf{5.61}_{0.04}$ & $\mathbf{5.94}_{0.05}$ & $\mathbf{7.27}_{0.03}$ & $\mathbf{9.64}_{0.05}$ & $\mathbf{19.57}_{0.32}$ & $\mathbf{21.82}_{0.42}$ & $\mathbf{45.07}_{0.51}$ & $\mathbf{79.47}_{0.56}$  \\
\cmidrule(lr){1-13}
 Temp. Scaling & $\mathbf{3.27}_{0.09}$ & $\mathbf{3.29}_{0.09}$ & $\mathbf{3.53}_{0.08}$ & $\mathbf{4.40}_{0.06}$ & $\mathbf{4.78}_{0.17}$ & $\mathbf{4.98}_{0.16}$ & $\mathbf{6.06}_{0.13}$ & $\mathbf{8.63}_{0.10}$ & $\mathbf{17.11}_{0.81}$ & $\mathbf{19.02}_{0.93}$ & $\mathbf{39.36}_{0.92}$ & $\mathbf{76.39}_{1.60}$  \\
 COG-TS & $5.03_{0.02}$ & $5.05_{0.01}$ & $5.21_{0.01}$ & $5.83_{0.02}$ & $6.35_{0.04}$ & $6.68_{0.05}$ & $8.07_{0.07}$ & $10.41_{0.06}$ & $22.01_{1.07}$ & $23.87_{0.90}$ & $45.77_{0.94}$ & $80.16_{0.36}$  \\
\bottomrule
\end{tabular}
}
\end{table}

\begin{figure}[h]
    \centering
    \includegraphics[width=1\linewidth]{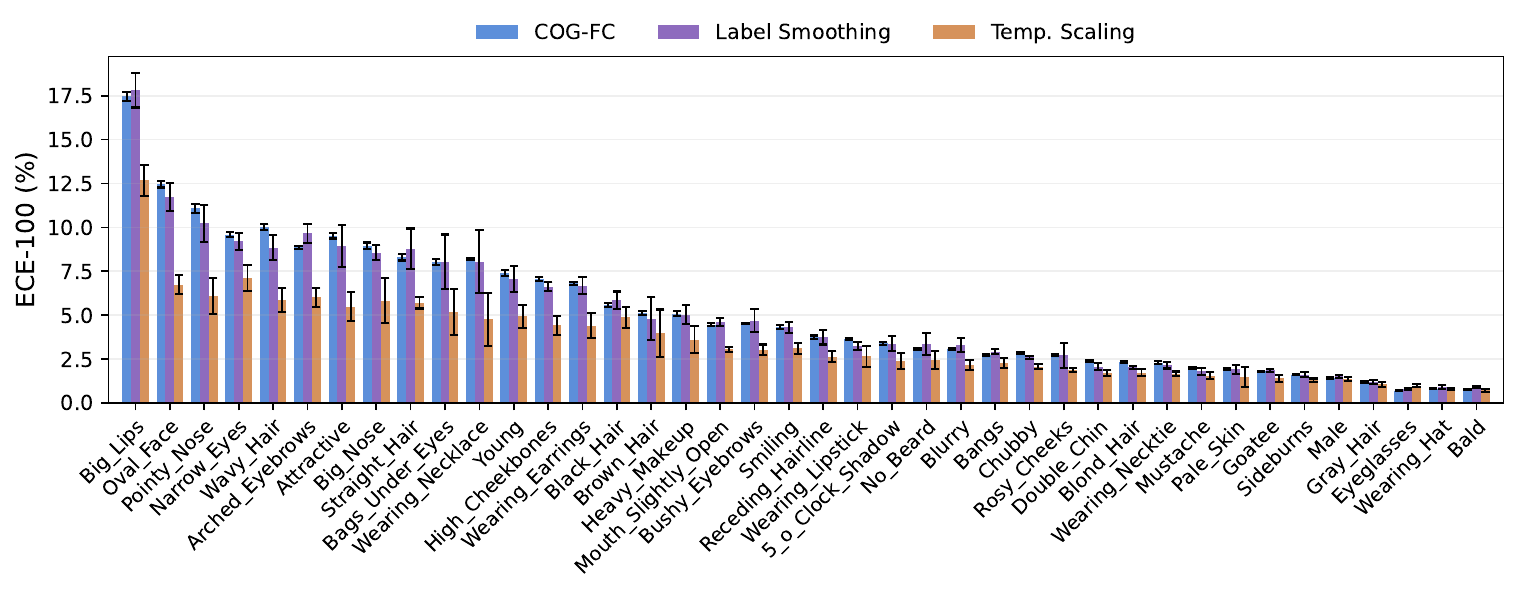}
    \caption{Per-task ECE-100 for the best-performing models evaluated in the main text.}
    \label{fig:app_per_task_ece}
\end{figure}

\paragraph{Per-task ECE-100.} In Fig.~\ref{fig:app_per_task_ece}, we plot the ECE-100 for the best-performing models on CelebA. We observe the same overall ordering of task difficulty across methods, with all methods identifying the same tasks as the most challenging, including \textit{Big Lips}. We also observe substantial variance for some tasks across all models, particularly for Label Smoothing, consistent with the variability observed in the corresponding Brier score results in Fig.~\ref{fig:per_task_metrics_main}.

\section{Results with a WRN backbone}\label{app:wrn}
In this section, we provide a brief summary of the results for in-distribution calibration, out-of-distribution detection, and performance under covariate shift using a WRN-28-10 backbone~\citep{zagoruyko2016wide}. The experimental settings are otherwise identical to those used with the ResNet backbone.

\paragraph{All Experimental Results.} In Tab.~\ref{tab:WRN_all_VS_all_ID}, we compare all model families in terms of calibration performance. On SVHN, we observe that almost all methods achieve similar Brier scores, with the main differences arising in NLL and accuracy, where the \method\ variants consistently dominate. On CIFAR-10, the (V)-TST models perform particularly well in both Brier score and accuracy. Finally, on CIFAR-100, we find that the \method\ variants generally achieve strong performance across the considered metrics. We again observe the same optimization issue with Matrix Scaling on CIFAR-100. In Tables~\ref{tab:WRN_app_all_VS_all_ECE}-~\ref{tab:WRN_app_all_VS_all_MCE}, we report ECE, RMSCE and MCE values for the same models. 

\begin{table}[h]
\caption{Comparison of all models and all variants seen in this work. We report $\text{mean}_{\text{std}}$. The best value in each column is shown in bold. \textbf{Rank} denotes the mean rank across ACC, NLL, and Brier for a given dataset (1 = best; ties share a rank).}
\label{tab:WRN_all_VS_all_ID}
\centering
\setlength{\tabcolsep}{3pt}
\resizebox{\textwidth}{!}{
\begin{tabular}{l|cccc|cccc|cccc}
\toprule
   & \multicolumn{4}{c|}{\textbf{SVHN}} & \multicolumn{4}{c|}{\textbf{CIFAR-10}} & \multicolumn{4}{c}{\textbf{CIFAR-100}} \\
  \textbf{Method} & ACC $(\uparrow)$ & NLL $(\downarrow)$ & Brier $(\downarrow)$ & \textbf{Rank} $(\downarrow)$ & ACC $(\uparrow)$ & NLL $(\downarrow)$ & Brier $(\downarrow)$ & \textbf{Rank} $(\downarrow)$ & ACC $(\uparrow)$ & NLL $(\downarrow)$ & Brier $(\downarrow)$ & \textbf{Rank} $(\downarrow)$ \\
\midrule
 Baseline & $95.35_{0.20}$ & $0.35_{0.01}$ & $0.08_{0.00}$ & 9.67 & $91.33_{0.60}$ & $0.53_{0.02}$ & $0.15_{0.01}$ & 7.67 & $69.49_{1.15}$ & $2.33_{0.11}$ & $0.52_{0.02}$ & 7.67 \\
 \midrule
  COG-FC & $95.60_{0.15}$ & $0.17_{0.00}$ & $\mathbf{0.07_{0.00}}$ & 3.00 & $88.88_{0.34}$ & $0.36_{0.01}$ & $0.17_{0.00}$ & 8.00 & $67.89_{0.55}$ & $\mathbf{1.19_{0.04}}$ & $0.44_{0.01}$ & 3.00 \\
  COG+LS & $\mathbf{95.92_{0.19}}$ & $0.22_{0.00}$ & $\mathbf{0.07_{0.00}}$ & 3.67 & $89.41_{0.30}$ & $0.37_{0.01}$ & $0.16_{0.00}$ & 7.67 & $71.98_{1.04}$ & $1.32_{0.04}$ & $0.40_{0.01}$ & 2.33 \\
  COG-TS & $95.64_{0.15}$ & $\mathbf{0.16_{0.00}}$ & $\mathbf{0.07_{0.00}}$ & 2.00 & $88.55_{0.11}$ & $0.38_{0.01}$ & $0.17_{0.00}$ & 9.00 & $65.14_{0.57}$ & $1.34_{0.06}$ & $0.48_{0.01}$ & 5.67 \\
  COG-MS & $95.76_{0.19}$ & $\mathbf{0.16_{0.01}}$ & $\mathbf{0.07_{0.00}}$ & $\color{PineGreen}{\mathbf{1.33}}$ & $87.74_{1.17}$ & $0.40_{0.02}$ & $0.18_{0.01}$ & 11.00 & $66.31_{0.86}$ & $1.35_{0.01}$ & $0.47_{0.01}$ & 5.33 \\
  COG-TST & $95.54_{0.10}$ & $0.17_{0.00}$ & $\mathbf{0.07_{0.00}}$ & 3.33 & $88.07_{0.49}$ & $0.40_{0.02}$ & $0.18_{0.01}$ & 10.67 & $61.45_{0.90}$ & $1.38_{0.03}$ & $0.50_{0.01}$ & 8.00 \\
  \midrule
 Label Smoothing & $95.74_{0.21}$ & $0.24_{0.01}$ & $0.08_{0.00}$ & 7.67 & $91.18_{0.36}$ & $0.39_{0.01}$ & $0.15_{0.01}$ & 7.33 & $\mathbf{73.58_{0.43}}$ & $1.26_{0.02}$ & $\mathbf{0.38_{0.01}}$ & $\color{PineGreen}{\mathbf{1.33}}$ \\
 \midrule
 Temp. Scaling & $95.35_{0.20}$ & $0.21_{0.01}$ & $\mathbf{0.07_{0.00}}$ & 5.67 & $91.33_{0.60}$ & $0.33_{0.02}$ & $0.14_{0.01}$ & 5.00 & $69.49_{1.15}$ & $1.44_{0.06}$ & $0.46_{0.01}$ & 5.00 \\
  Matrix Scaling & $95.48_{0.18}$ & $0.17_{0.01}$ & $\mathbf{0.07_{0.00}}$ & 3.67 & $91.46_{0.50}$ & $\mathbf{0.25_{0.02}}$ & $\mathbf{0.12_{0.01}}$ & 2.00 & $54.84_{1.33}$ & $21.89_{2.93}$ & $0.88_{0.03}$ & 12.00 \\
  TST & $95.27_{0.02}$ & $0.18_{0.00}$ & $\mathbf{0.07_{0.00}}$ & 5.67 & $92.27_{0.06}$ & $0.26_{0.00}$ & $\mathbf{0.12_{0.00}}$ & $\color{PineGreen}{\mathbf{1.67}}$ & $62.70_{0.36}$ & $1.57_{0.03}$ & $0.53_{0.00}$ & 10.00 \\
  \midrule
 V-TST (MC=1) & $94.70_{0.11}$ & $0.23_{0.00}$ & $0.09_{0.00}$ & 11.33 & $91.76_{0.09}$ & $0.31_{0.01}$ & $0.13_{0.00}$ & 3.67 & $61.69_{0.46}$ & $1.54_{0.03}$ & $0.52_{0.01}$ & 9.33 \\
  V-TST (MC=10) & $95.24_{0.04}$ & $0.20_{0.00}$ & $0.08_{0.00}$ & 9.00 & $\mathbf{92.33_{0.05}}$ & $0.27_{0.00}$ & $\mathbf{0.12_{0.00}}$ & $\color{PineGreen}{\mathbf{1.67}}$ & $63.21_{0.42}$ & $1.41_{0.01}$ & $0.50_{0.00}$ & 7.33 \\
\bottomrule
\end{tabular}
}
\end{table}

\begin{table}[h]
\caption{Comparison of ECE ($\downarrow$ lower is better) for all models and all variants seen in this work for multiple choices of the number of bins. We report $\text{mean}_{\text{std}}$. The best value in each column is shown in bold.}
\label{tab:WRN_app_all_VS_all_ECE}
\centering
\setlength{\tabcolsep}{3pt}
\resizebox{\textwidth}{!}{
\begin{tabular}{l|cccc|cccc|cccc}
\toprule
   & \multicolumn{4}{c|}{\textbf{SVHN}} & \multicolumn{4}{c|}{\textbf{CIFAR-10}} & \multicolumn{4}{c}{\textbf{CIFAR-100}} \\
  \textbf{Method} & ECE-10 & ECE-15 & ECE-100 & ECE-500 & ECE-10 & ECE-15 & ECE-100 & ECE-500 & ECE-10 & ECE-15 & ECE-100 & ECE-500 \\
\midrule
 Baseline & $3.63_{0.15}$ & $3.63_{0.15}$ & $3.67_{0.16}$ & $3.83_{0.15}$ & $6.70_{0.41}$ & $6.70_{0.40}$ & $6.80_{0.37}$ & $7.21_{0.34}$ & $22.90_{0.88}$ & $22.90_{0.88}$ & $22.91_{0.88}$ & $23.12_{0.89}$ \\
 \midrule
  COG-FC & $1.07_{0.10}$ & $1.08_{0.09}$ & $1.27_{0.08}$ & $1.89_{0.08}$ & $4.25_{0.33}$ & $4.22_{0.34}$ & $4.55_{0.28}$ & $5.62_{0.26}$ & $7.82_{1.49}$ & $7.82_{1.49}$ & $8.01_{1.46}$ & $9.45_{1.19}$ \\
  COG+LS & $7.46_{0.14}$ & $7.48_{0.14}$ & $7.70_{0.14}$ & $8.29_{0.14}$ & $4.16_{0.12}$ & $4.37_{0.12}$ & $5.29_{0.12}$ & $6.93_{0.09}$ & $3.69_{0.40}$ & $4.67_{0.41}$ & $5.62_{0.33}$ & $7.90_{0.38}$ \\
  COG-TS & $1.00_{0.07}$ & $1.01_{0.08}$ & $1.23_{0.11}$ & $1.88_{0.08}$ & $4.60_{0.17}$ & $4.61_{0.18}$ & $4.89_{0.13}$ & $5.94_{0.14}$ & $9.84_{1.33}$ & $9.84_{1.33}$ & $9.87_{1.33}$ & $10.87_{1.17}$ \\
  COG-MS & $1.24_{0.12}$ & $1.26_{0.12}$ & $1.42_{0.14}$ & $1.93_{0.11}$ & $5.17_{0.27}$ & $5.19_{0.27}$ & $5.39_{0.32}$ & $6.33_{0.32}$ & $11.82_{1.04}$ & $11.82_{1.04}$ & $11.90_{1.01}$ & $12.53_{0.87}$ \\
  COG-TST & $0.97_{0.16}$ & $1.00_{0.16}$ & $1.22_{0.16}$ & $\mathbf{1.83_{0.14}}$ & $4.69_{0.47}$ & $4.69_{0.46}$ & $4.97_{0.38}$ & $5.95_{0.31}$ & $\mathbf{1.13_{0.40}}$ & $\mathbf{1.23_{0.59}}$ & $\mathbf{2.95_{0.22}}$ & $\mathbf{6.68_{0.26}}$ \\
  \midrule
 Label Smoothing & $7.01_{0.43}$ & $7.00_{0.28}$ & $7.39_{0.25}$ & $7.77_{0.30}$ & $5.40_{0.87}$ & $5.60_{0.27}$ & $6.43_{0.15}$ & $7.30_{0.09}$ & $3.91_{0.59}$ & $4.94_{0.29}$ & $5.73_{0.38}$ & $7.77_{0.27}$ \\
 \midrule
 Temp. Scaling & $2.53_{0.12}$ & $2.53_{0.12}$ & $2.59_{0.12}$ & $2.89_{0.11}$ & $4.76_{0.29}$ & $4.82_{0.23}$ & $4.97_{0.21}$ & $5.64_{0.27}$ & $15.64_{0.68}$ & $15.64_{0.68}$ & $15.71_{0.67}$ & $16.28_{0.61}$ \\
  Matrix Scaling & $\mathbf{0.70_{0.10}}$ & $\mathbf{0.75_{0.11}}$ & $\mathbf{1.05_{0.13}}$ & $\mathbf{1.83_{0.10}}$ & $\mathbf{0.98_{0.26}}$ & $\mathbf{1.02_{0.27}}$ & $\mathbf{1.74_{0.21}}$ & $\mathbf{3.53_{0.19}}$ & $43.27_{1.61}$ & $43.27_{1.61}$ & $43.29_{1.62}$ & $43.40_{1.63}$ \\
  TST & $1.48_{0.02}$ & $1.50_{0.02}$ & $1.67_{0.03}$ & $2.19_{0.03}$ & $3.13_{0.05}$ & $3.14_{0.06}$ & $3.40_{0.07}$ & $4.39_{0.08}$ & $14.86_{1.07}$ & $14.86_{1.07}$ & $14.93_{1.09}$ & $15.39_{1.03}$ \\
  \midrule
 V-TST (MC=1) & $1.37_{0.78}$ & $1.38_{0.77}$ & $1.77_{0.57}$ & $2.45_{0.45}$ & $3.27_{0.43}$ & $3.28_{0.41}$ & $3.55_{0.35}$ & $4.54_{0.27}$ & $9.79_{2.61}$ & $9.82_{2.61}$ & $10.07_{2.48}$ & $11.50_{2.10}$ \\
  V-TST (MC=10) & $2.37_{0.71}$ & $2.38_{0.70}$ & $2.55_{0.63}$ & $3.17_{0.52}$ & $2.10_{0.33}$ & $2.11_{0.34}$ & $2.57_{0.28}$ & $3.79_{0.21}$ & $6.17_{2.37}$ & $6.21_{2.32}$ & $6.66_{1.95}$ & $8.93_{1.38}$ \\

\bottomrule
\end{tabular}
}
\end{table}

\begin{table}[t]
\caption{Comparison of RMSCE ($\downarrow$ lower is better) for all models and all variants seen in this work for multiple choices of the number of bins. We report $\text{mean}_{\text{std}}$. The best value in each column is shown in bold.}
\label{tab:WRN_app_all_VS_all_RMSCE}
\centering
\setlength{\tabcolsep}{3pt}
\resizebox{\textwidth}{!}{
\begin{tabular}{l|cccc|cccc|cccc}
\toprule
   & \multicolumn{4}{c|}{\textbf{SVHN}} & \multicolumn{4}{c|}{\textbf{CIFAR-10}} & \multicolumn{4}{c}{\textbf{CIFAR-100}} \\
  \textbf{Method} & RMSCE-10 & RMSCE-15 & RMSCE-100 & RMSCE-500 & RMSCE-10 & RMSCE-15 & RMSCE-100 & RMSCE-500 & RMSCE-10 & RMSCE-15 & RMSCE-100 & RMSCE-500 \\
\midrule
 Baseline & $5.25_{0.27}$ & $5.89_{0.27}$ & $7.68_{0.18}$ & $9.34_{0.12}$ & $8.35_{0.61}$ & $8.91_{0.55}$ & $11.50_{0.42}$ & $13.94_{0.43}$ & $24.34_{0.91}$ & $25.02_{0.98}$ & $27.40_{0.90}$ & $29.24_{0.84}$ \\
   \midrule
  COG-FC & $1.58_{0.18}$ & $1.76_{0.20}$ & $2.87_{0.22}$ & $5.48_{0.34}$ & $5.47_{0.50}$ & $5.93_{0.56}$ & $7.52_{0.44}$ & $10.64_{0.48}$ & $8.78_{1.67}$ & $8.99_{1.78}$ & $9.97_{1.71}$ & $13.08_{1.42}$ \\
  COG+LS & $7.56_{0.14}$ & $7.62_{0.14}$ & $7.97_{0.17}$ & $9.13_{0.18}$ & $4.48_{0.08}$ & $4.76_{0.11}$ & $6.09_{0.21}$ & $9.40_{0.25}$ & $5.01_{0.30}$ & $5.73_{0.45}$ & $7.18_{0.53}$ & $11.35_{0.37}$ \\
  COG-TS & $1.64_{0.10}$ & $1.82_{0.13}$ & $2.92_{0.28}$ & $5.48_{0.29}$ & $5.79_{0.28}$ & $6.25_{0.31}$ & $7.86_{0.17}$ & $11.14_{0.34}$ & $10.93_{1.47}$ & $11.12_{1.48}$ & $11.93_{1.50}$ & $14.54_{1.38}$ \\
  COG-MS & $1.99_{0.14}$ & $2.27_{0.19}$ & $3.29_{0.24}$ & $5.69_{0.27}$ & $6.48_{0.50}$ & $6.91_{0.34}$ & $8.46_{0.58}$ & $11.42_{0.58}$ & $13.12_{1.15}$ & $13.31_{1.22}$ & $14.21_{1.23}$ & $16.43_{0.99}$ \\
  COG-TST & $1.58_{0.25}$ & $1.77_{0.35}$ & $2.92_{0.36}$ & $5.42_{0.34}$ & $5.83_{0.60}$ & $6.32_{0.63}$ & $7.73_{0.69}$ & $10.65_{0.53}$ & $\mathbf{1.48_{0.64}}$ & $\mathbf{1.68_{0.78}}$ & $\mathbf{4.03_{0.29}}$ & $\mathbf{9.25_{0.25}}$ \\
    \midrule
 Label Smoothing & $7.19_{0.34}$ & $7.18_{0.29}$ & $7.84_{0.25}$ & $9.09_{0.30}$ & $6.34_{0.58}$ & $6.89_{0.37}$ & $8.23_{0.26}$ & $10.86_{0.23}$ & $5.25_{0.56}$ & $6.09_{0.69}$ & $7.39_{0.58}$ & $11.14_{0.42}$ \\
   \midrule
 Temp. Scaling & $4.10_{0.16}$ & $4.49_{0.17}$ & $5.50_{0.24}$ & $7.25_{0.23}$ & $6.60_{0.33}$ & $7.10_{0.24}$ & $8.86_{0.32}$ & $11.65_{0.28}$ & $17.23_{0.75}$ & $17.62_{0.75}$ & $19.23_{0.55}$ & $21.18_{0.54}$ \\
  Matrix Scaling & $\mathbf{1.09_{0.09}}$ & $\mathbf{1.32_{0.15}}$ & $\mathbf{2.47_{0.09}}$ & $\mathbf{5.06_{0.12}}$ & $\mathbf{1.83_{0.36}}$ & $\mathbf{1.93_{0.38}}$ & $\mathbf{4.05_{0.50}}$ & $\mathbf{8.53_{0.47}}$ & $43.40_{1.59}$ & $43.50_{1.57}$ & $44.03_{1.50}$ & $44.80_{1.39}$ \\
  TST & $2.31_{0.06}$ & $2.57_{0.05}$ & $3.56_{0.13}$ & $5.99_{0.16}$ & $4.61_{0.14}$ & $4.84_{0.18}$ & $6.48_{0.17}$ & $10.01_{0.19}$ & $16.04_{1.13}$ & $16.38_{1.15}$ & $17.45_{1.21}$ & $19.36_{1.11}$ \\
  \midrule
 V-TST (MC=1) & $2.22_{1.08}$ & $2.37_{1.01}$ & $3.41_{0.73}$ & $5.65_{0.53}$ & $4.55_{0.68}$ & $4.77_{0.65}$ & $6.26_{0.56}$ & $9.91_{0.34}$ & $10.80_{2.84}$ & $11.07_{2.93}$ & $12.10_{2.88}$ & $14.88_{2.41}$ \\
  V-TST (MC=10) & $2.86_{0.97}$ & $2.99_{0.97}$ & $3.88_{0.77}$ & $6.06_{0.51}$ & $3.32_{0.49}$ & $3.46_{0.58}$ & $5.14_{0.43}$ & $9.24_{0.37}$ & $6.90_{2.31}$ & $7.09_{2.34}$ & $8.29_{2.12}$ & $12.08_{1.49}$ \\
\bottomrule
\end{tabular}
}
\end{table}

\begin{table}[t]
\caption{Comparison of MCE ($\downarrow$ lower is better) for all models and all variants seen in this work for multiple choices of the number of bins. We report $\text{mean}_{\text{std}}$. The best value in each column is shown in bold.}
\label{tab:WRN_app_all_VS_all_MCE}
\centering
\setlength{\tabcolsep}{3pt}
\resizebox{\textwidth}{!}{
\begin{tabular}{l|cccc|cccc|cccc}
\toprule
   & \multicolumn{4}{c|}{\textbf{SVHN}} & \multicolumn{4}{c|}{\textbf{CIFAR-10}} & \multicolumn{4}{c}{\textbf{CIFAR-100}} \\
  \textbf{Method} & MCE-10 & MCE-15 & MCE-100 & MCE-500 & MCE-10 & MCE-15 & MCE-100 & MCE-500 & MCE-10 & MCE-15 & MCE-100 & MCE-500 \\
\midrule
 Baseline & $40.91_{18.18}$ & $35.91_{3.53}$ & $64.66_{5.09}$ & $86.39_{5.07}$ & $34.74_{1.87}$ & $37.27_{4.82}$ & $62.85_{4.98}$ & $88.29_{1.94}$ & $46.89_{1.18}$ & $50.50_{2.59}$ & $63.17_{12.17}$ & $81.55_{5.53}$ \\
 \midrule
  COG-FC & $29.45_{28.99}$ & $31.63_{27.43}$ & $50.32_{24.93}$ & $76.26_{2.50}$ & $15.82_{1.90}$ & $19.94_{3.79}$ & $64.66_{15.10}$ & $73.02_{2.25}$ & $13.84_{3.09}$ & $14.60_{3.64}$ & $23.54_{2.02}$ & $85.11_{2.45}$ \\
  COG+LS & $28.77_{31.43}$ & $30.17_{30.53}$ & $54.89_{25.22}$ & $81.89_{9.10}$ & $13.48_{6.73}$ & $17.64_{4.10}$ & $69.21_{20.08}$ & $77.80_{1.23}$ & $11.97_{2.43}$ & $12.48_{2.21}$ & $40.61_{32.25}$ & $74.45_{20.82}$ \\
  COG-TS & $46.93_{31.63}$ & $56.62_{34.70}$ & $74.02_{15.11}$ & $80.11_{2.95}$ & $20.19_{6.53}$ & $32.24_{13.47}$ & $61.07_{22.01}$ & $74.28_{4.21}$ & $17.48_{2.92}$ & $18.71_{3.04}$ & $26.67_{5.20}$ & $71.28_{19.95}$ \\
  COG-MS & $19.20_{8.44}$ & $26.31_{13.57}$ & $60.68_{26.69}$ & $78.79_{4.65}$ & $16.12_{0.84}$ & $36.00_{22.68}$ & $67.60_{17.03}$ & $74.53_{2.90}$ & $20.77_{2.37}$ & $21.35_{2.66}$ & $43.62_{24.09}$ & $83.56_{3.11}$ \\
  COG-TST & $29.41_{29.53}$ & $29.35_{29.59}$ & $62.56_{23.51}$ & $77.91_{2.75}$ & $18.00_{5.62}$ & $26.74_{4.31}$ & $62.90_{14.30}$ & $\mathbf{72.38_{1.93}}$ & $\mathbf{4.50_{2.22}}$ & $\mathbf{5.46_{1.55}}$ & $\mathbf{13.96_{0.93}}$ & $\mathbf{62.05_{27.63}}$ \\
  \midrule
 Label Smoothing & $16.08_{1.36}$ & $31.15_{24.57}$ & $74.36_{4.73}$ & $84.32_{9.63}$ & $26.12_{2.19}$ & $25.58_{3.26}$ & $61.86_{23.73}$ & $87.29_{10.27}$ & $12.72_{2.50}$ & $14.58_{1.45}$ & $29.08_{10.96}$ & $94.47_{1.24}$ \\
 \midrule
 Temp. Scaling & $22.74_{4.89}$ & $24.41_{1.46}$ & $60.12_{17.94}$ & $\mathbf{74.47_{7.26}}$ & $24.67_{6.86}$ & $24.68_{1.03}$ & $64.35_{9.49}$ & $81.21_{1.80}$ & $43.30_{27.09}$ & $31.84_{0.59}$ & $52.08_{22.23}$ & $85.07_{3.89}$ \\
  Matrix Scaling & $\mathbf{13.04_{7.59}}$ & $\mathbf{12.60_{8.17}}$ & $48.11_{24.22}$ & $75.64_{2.90}$ & $\mathbf{11.09_{8.83}}$ & $19.19_{7.38}$ & $57.00_{17.43}$ & $72.69_{3.14}$ & $61.14_{2.41}$ & $65.99_{2.27}$ & $80.29_{2.35}$ & $95.26_{2.02}$ \\
  TST & $44.39_{34.12}$ & $45.29_{33.01}$ & $79.53_{1.86}$ & $79.74_{1.72}$ & $18.69_{4.36}$ & $20.84_{4.05}$ & $\mathbf{44.86_{16.75}}$ & $76.60_{3.04}$ & $25.41_{2.27}$ & $38.53_{27.14}$ & $36.97_{4.46}$ & $78.64_{12.89}$ \\
  \midrule
 V-TST (MC=1) & $17.31_{3.82}$ & $17.48_{3.44}$ & $\mathbf{39.69_{8.96}}$ & $77.45_{2.56}$ & $16.51_{4.04}$ & $19.32_{5.87}$ & $53.67_{21.74}$ & $77.96_{2.75}$ & $18.18_{4.85}$ & $18.81_{5.30}$ & $32.59_{6.33}$ & $62.45_{18.01}$ \\
  V-TST (MC=10) & $15.73_{7.78}$ & $17.62_{5.65}$ & $62.68_{26.18}$ & $80.77_{1.84}$ & $13.76_{4.44}$ & $\mathbf{14.18_{4.31}}$ & $45.09_{19.32}$ & $77.35_{1.95}$ & $11.73_{2.79}$ & $12.23_{3.18}$ & $24.49_{11.50}$ & $64.94_{25.51}$ \\
\bottomrule
\end{tabular}
}
\end{table}

\begin{table}[t]
\caption{Performance in out-of-distribution detection or all models and all variants seen in this work.}
\label{tab:WRN_results_full_ood}
\centering
\setlength{\tabcolsep}{3pt}
\resizebox{\textwidth}{!}{
\begin{tabular}{l|cc|c|cc|c|cc|c}
\toprule
   & \multicolumn{3}{c|}{\textbf{SVHN}} & \multicolumn{3}{c|}{\textbf{CIFAR-10}} & \multicolumn{3}{c}{\textbf{CIFAR-100}}\\
  \textbf{Method} & AUROC $(\uparrow)$ & FPR95 $(\downarrow)$ & Rank $(\downarrow)$ & AUROC $(\uparrow)$ & FPR95 $(\downarrow)$ & Rank $(\downarrow)$ & AUROC $(\uparrow)$ & FPR95 $(\downarrow)$ &Rank $(\downarrow)$ \\
\midrule
 Baseline & $93.26_{0.75}$ & $42.67_{3.11}$ & 9.50 & $87.00_{1.36}$ & $71.21_{4.41}$ & 4.50 & $70.66_{2.85}$ & $86.80_{1.38}$ & 8.50 \\
 \midrule
  COG-FC & $96.11_{0.34}$ & $26.30_{2.79}$ & 2.50 & $85.11_{1.13}$ & $77.61_{2.59}$ & 9.50 & $70.20_{3.78}$ & $84.34_{2.25}$ & 8.50 \\
  COG+LS & $93.98_{0.27}$ & $25.37_{1.76}$ & 4.00 & $82.40_{2.77}$ & $74.75_{4.02}$ & 9.00 & $73.32_{2.30}$ & $81.67_{1.68}$ & 4.00 \\
  COG-TS & $95.43_{0.48}$ & $31.51_{3.42}$ & 4.50 & $85.53_{0.57}$ & $76.66_{1.60}$ & 8.50 & $67.50_{2.59}$ & $87.76_{1.51}$ & 10.50 \\
  COG-MS & $95.69_{0.31}$ & $29.02_{2.13}$ & 3.50 & $84.58_{1.10}$ & $78.71_{3.14}$ & 11.00 & $68.82_{4.94}$ & $88.05_{1.70}$ & 10.50 \\
  COG-TST & $\mathbf{96.41_{0.38}}$ & $\mathbf{24.16_{2.84}}$ & $\color{PineGreen}{\mathbf{1.00}}$ & $86.42_{1.73}$ & $75.03_{5.52}$ & 7.00 & $73.16_{2.23}$ & $81.65_{2.84}$ & 4.50 \\
  \midrule
 Label Smoothing & $80.74_{1.00}$ & $38.56_{1.67}$ & 9.50 & $67.88_{1.51}$ & $75.99_{1.67}$ & 10.50 & $\mathbf{75.69_{1.50}}$ & $80.67_{2.57}$ & $\color{PineGreen}{\mathbf{1.50}}$\\
 Temp. Scaling & $93.81_{0.77}$ & $39.81_{3.38}$ & 8.00 & $87.63_{1.38}$ & $69.27_{5.46}$ & 2.50 & $72.34_{2.92}$ & $83.72_{1.93}$ & 6.50 \\
  Matrix Scaling & $94.66_{0.66}$ & $34.41_{3.05}$ & 5.50 & $\mathbf{89.11_{1.20}}$ & $\mathbf{64.47_{6.49}}$ & $\color{PineGreen}{\mathbf{1.00}}$ & $58.62_{1.37}$ & $92.03_{0.61}$ & 12.00 \\
  TST & $93.84_{0.04}$ & $40.50_{0.37}$ & 8.00 & $88.34_{0.04}$ & $69.75_{0.57}$ & 2.50 & $70.77_{0.42}$ & $83.70_{0.32}$ & 6.50 \\
  \midrule
 V-TST (MC=1) & $88.52_{0.89}$ & $51.82_{0.69}$ & 11.50 & $84.66_{0.63}$ & $72.49_{1.24}$ & 7.50 & $73.21_{0.35}$ & $80.86_{0.51}$ & 3.50 \\
  V-TST (MC=10) & $91.70_{0.67}$ & $43.13_{0.36}$ & 10.50 & $86.66_{0.48}$ & $69.94_{0.87}$ & 4.50 & $74.20_{0.42}$ & $\mathbf{79.59_{0.18}}$ & $\color{PineGreen}{\mathbf{1.50}}$ \\

\bottomrule
\end{tabular}
}
\end{table}

\begin{table}[t]
\caption{Performance under covariate shifts. Shift metrics are averaged across all considered covariate shifts.}
\label{tab:WRN_shift_results_full}
\centering
\setlength{\tabcolsep}{3pt}
\resizebox{\textwidth}{!}{
\begin{tabular}{l|ccc|c|ccc|c|ccc|c|}
\toprule
   & \multicolumn{4}{c|}{\textbf{SVHN}} & \multicolumn{4}{c|}{\textbf{CIFAR-10}} & \multicolumn{4}{c|}{\textbf{CIFAR-100}}\\
  \textbf{Method} & ACC $(\uparrow)$ & NLL $(\downarrow)$ & Brier $(\downarrow)$ &Rank $(\downarrow)$ & ACC $(\uparrow)$ & NLL $(\downarrow)$ & Brier  $(\downarrow)$ &Rank $(\downarrow)$ & ACC $(\uparrow)$ & NLL $(\downarrow)$ & Brier $(\downarrow)$ &Rank $(\downarrow)$ \\
\midrule
 Baseline & $44.53_{0.94}$ & $5.95_{0.19}$ & $1.00_{0.02}$ & 10.67 & $74.24_{0.54}$ & $1.99_{0.13}$ & $0.45_{0.01}$ & 8.33 & $46.50_{1.29}$ & $5.16_{0.34}$ & $0.91_{0.02}$ & 8.33 \\
 \midrule
  COG-FC & $45.34_{0.93}$ & $2.34_{0.06}$ & $0.81_{0.01}$ & 4.00 & $67.42_{0.31}$ & $1.28_{0.01}$ & $0.50_{0.00}$ & 9.33 & $44.03_{1.53}$ & $2.71_{0.11}$ & $0.76_{0.02}$ & 4.33 \\
  COG+LS & $\mathbf{45.48_{1.00}}$ & $\mathbf{1.87_{0.03}}$ & $\mathbf{0.76_{0.01}}$ & $\color{PineGreen}{\mathbf{1.00}}$ & $67.67_{0.35}$ & $1.10_{0.01}$ & $0.47_{0.00}$ & 7.67 & $47.04_{1.07}$ & $2.59_{0.07}$ & $0.71_{0.02}$ & 2.00 \\
  COG-TS & $44.78_{0.68}$ & $3.27_{0.09}$ & $0.89_{0.02}$ & 9.00 & $66.55_{0.39}$ & $1.92_{0.04}$ & $0.55_{0.01}$ & 11.67 & $45.39_{2.00}$ & $3.73_{0.21}$ & $0.83_{0.03}$ & 8.00 \\
  COG-MS & $44.79_{0.35}$ & $2.53_{0.09}$ & $0.83_{0.01}$ & 6.00 & $67.22_{1.29}$ & $1.47_{0.07}$ & $0.51_{0.02}$ & 10.67 & $40.61_{1.31}$ & $2.87_{0.14}$ & $0.79_{0.02}$ & 7.33 \\
  COG-TST & $44.79_{0.99}$ & $2.28_{0.05}$ & $0.80_{0.02}$ & 3.67 & $68.02_{0.73}$ & $1.37_{0.03}$ & $0.49_{0.01}$ & 8.67 & $38.08_{0.60}$ & $2.77_{0.02}$ & $0.78_{0.00}$ & 7.00 \\
  \midrule
 Label Smoothing & $44.81_{0.82}$ & $2.18_{0.03}$ & $0.87_{0.01}$ & 4.33 & $74.65_{1.09}$ & $1.02_{0.03}$ & $0.41_{0.01}$ & 4.67 & $\mathbf{50.78_{1.05}}$ & $\mathbf{2.40_{0.05}}$ & $\mathbf{0.67_{0.01}}$ & $\color{PineGreen}{\mathbf{1.00}}$ \\
 \midrule
 Temp. Scaling & $44.53_{0.94}$ & $3.38_{0.09}$ & $0.91_{0.01}$ & 10.00 & $74.24_{0.54}$ & $1.18_{0.06}$ & $0.41_{0.01}$ & 6.00 & $46.50_{1.29}$ & $3.10_{0.17}$ & $0.80_{0.02}$ & 6.33 \\
  Matrix Scaling & $44.82_{0.52}$ & $2.30_{0.07}$ & $0.80_{0.01}$ & 3.33 & $74.66_{0.96}$ & $\mathbf{0.88_{0.05}}$ & $0.37_{0.01}$ & 2.33 & $33.22_{1.45}$ & $40.21_{3.78}$ & $1.30_{0.03}$ & 12.00 \\
  TST & $41.12_{0.06}$ & $2.60_{0.02}$ & $0.88_{0.00}$ & 9.33 & $75.56_{0.05}$ & $0.94_{0.01}$ & $0.37_{0.00}$ & 2.33 & $41.06_{0.22}$ & $3.22_{0.09}$ & $0.83_{0.01}$ & 8.67 \\
  \midrule
 V-TST (MC=1) & $40.36_{0.13}$ & $2.54_{0.12}$ & $0.88_{0.02}$ & 9.33 & $74.90_{0.05}$ & $1.04_{0.04}$ & $0.39_{0.01}$ & 4.00 & $40.36_{0.23}$ & $2.86_{0.12}$ & $0.79_{0.02}$ & 7.33 \\
  V-TST (MC=10) & $41.22_{0.15}$ & $2.20_{0.07}$ & $0.82_{0.01}$ & 6.00 & $\mathbf{75.71_{0.04}}$ & $0.89_{0.03}$ & $\mathbf{0.36_{0.01}}$ & $\color{PineGreen}{\mathbf{1.33}}$ & $41.61_{0.25}$ & $2.65_{0.07}$ & $0.76_{0.01}$ & 4.33 \\
\bottomrule
\end{tabular}
}
\end{table}

\begin{table}[h]
\caption{ECE averaged across all considered covariate shifts for all models and all variants seen in this work for multiple choices of the number of bins.}
\label{tab:WRN_shift_ood_ece}
\centering
\setlength{\tabcolsep}{3pt}
\resizebox{\textwidth}{!}{
\begin{tabular}{l|cccc|cccc|cccc|}
\toprule
   & \multicolumn{4}{c|}{\textbf{SVHN}} & \multicolumn{4}{c|}{\textbf{CIFAR-10}} & \multicolumn{4}{c|}{\textbf{CIFAR-100}}\\
  \textbf{Method} & ECE-10 & ECE-15 & ECE-100 & ECE-500 & ECE-10 & ECE-15 & ECE-100 & ECE-500 & ECE-10 & ECE-15 & ECE-100 & ECE-500 \\
\midrule
 Baseline & $47.06_{0.83}$ & $47.06_{0.83}$ & $47.06_{0.83}$ & $47.06_{0.83}$ & $20.89_{0.53}$ & $20.89_{0.53}$ & $20.89_{0.53}$ & $20.86_{0.53}$ & $40.89_{1.24}$ & $40.88_{1.24}$ & $40.88_{1.24}$ & $40.86_{1.24}$ \\
 \midrule
  COG-FC & $30.31_{0.98}$ & $30.31_{0.98}$ & $30.30_{0.98}$ & $30.31_{0.98}$ & $17.97_{0.28}$ & $17.97_{0.28}$ & $17.96_{0.28}$ & $17.94_{0.28}$ & $20.91_{1.30}$ & $20.91_{1.30}$ & $20.90_{1.30}$ & $20.90_{1.30}$ \\
  COG+LS & $\mathbf{22.99_{0.56}}$ & $\mathbf{22.99_{0.56}}$ & $\mathbf{22.99_{0.55}}$ & $\mathbf{23.00_{0.55}}$ & $10.02_{0.33}$ & $10.02_{0.33}$ & $10.02_{0.33}$ & $10.02_{0.33}$ & $12.78_{2.20}$ & $12.79_{2.17}$ & $12.81_{2.18}$ & $12.82_{2.17}$ \\
  COG-TS & $38.11_{0.94}$ & $38.11_{0.94}$ & $38.11_{0.94}$ & $38.12_{0.94}$ & $23.93_{0.32}$ & $23.93_{0.32}$ & $23.93_{0.32}$ & $23.91_{0.32}$ & $32.18_{1.10}$ & $32.17_{1.10}$ & $32.16_{1.10}$ & $32.16_{1.10}$ \\
  COG-MS & $32.44_{1.20}$ & $32.44_{1.20}$ & $32.44_{1.20}$ & $32.45_{1.19}$ & $19.02_{0.87}$ & $19.02_{0.87}$ & $19.01_{0.87}$ & $19.00_{0.87}$ & $20.49_{1.95}$ & $20.49_{1.95}$ & $20.49_{1.95}$ & $20.49_{1.95}$ \\
  COG-TST & $28.25_{1.33}$ & $28.25_{1.33}$ & $28.25_{1.33}$ & $28.26_{1.34}$ & $18.00_{0.21}$ & $17.99_{0.21}$ & $17.98_{0.21}$ & $17.97_{0.21}$ & $12.52_{0.62}$ & $12.52_{0.62}$ & $12.52_{0.62}$ & $12.52_{0.62}$ \\
  \midrule
 Label Smoothing & $34.25_{0.52}$ & $34.25_{0.52}$ & $34.25_{0.52}$ & $34.26_{0.52}$ & $11.10_{0.60}$ & $11.08_{0.60}$ & $11.19_{0.61}$ & $11.19_{0.61}$ & $\mathbf{11.57_{1.01}}$ & $\mathbf{11.59_{0.99}}$ & $\mathbf{11.60_{1.00}}$ & $\mathbf{11.61_{0.99}}$ \\
 \midrule
 Temp. Scaling & $39.69_{0.83}$ & $39.69_{0.83}$ & $39.69_{0.83}$ & $39.70_{0.83}$ & $16.51_{0.58}$ & $16.51_{0.58}$ & $16.50_{0.58}$ & $16.48_{0.58}$ & $29.35_{1.25}$ & $29.35_{1.25}$ & $29.34_{1.25}$ & $29.33_{1.25}$ \\
  Matrix Scaling & $28.09_{0.82}$ & $28.09_{0.82}$ & $28.09_{0.82}$ & $28.10_{0.81}$ & $\mathbf{9.81_{0.81}}$ & $\mathbf{9.81_{0.81}}$ & $\mathbf{9.79_{0.82}}$ & $\mathbf{9.78_{0.81}}$ & $64.47_{1.72}$ & $64.46_{1.72}$ & $64.45_{1.72}$ & $64.43_{1.72}$ \\
  TST & $34.83_{0.08}$ & $34.83_{0.08}$ & $34.83_{0.08}$ & $34.83_{0.08}$ & $12.75_{0.12}$ & $12.75_{0.12}$ & $12.74_{0.12}$ & $12.72_{0.12}$ & $27.52_{1.57}$ & $27.52_{1.57}$ & $27.51_{1.57}$ & $27.51_{1.57}$ \\
  \midrule
 V-TST (MC=1) & $32.29_{2.16}$ & $32.29_{2.16}$ & $32.29_{2.16}$ & $32.29_{2.16}$ & $13.28_{0.95}$ & $13.27_{0.95}$ & $13.25_{0.95}$ & $13.22_{0.94}$ & $19.75_{3.50}$ & $19.75_{3.50}$ & $19.75_{3.50}$ & $19.75_{3.50}$ \\
  V-TST (MC=10) & $27.39_{1.82}$ & $27.39_{1.82}$ & $27.39_{1.82}$ & $27.39_{1.82}$ & $10.56_{1.01}$ & $10.56_{1.01}$ & $10.52_{1.01}$ & $10.50_{1.00}$ & $15.57_{2.96}$ & $15.57_{2.96}$ & $15.57_{2.96}$ & $15.57_{2.96}$ \\
\bottomrule
\end{tabular}
}
\end{table}

\begin{table}[h]
\caption{RMSCE averaged across all considered covariate shifts for all models and all variants seen in this work for multiple choices of the number of bins.}
\label{tab:WRN_shift_ood_rmsce}
\centering
\setlength{\tabcolsep}{3pt}
\resizebox{\textwidth}{!}{
\begin{tabular}{l|cccc|cccc|cccc|}
\toprule
   & \multicolumn{4}{c|}{\textbf{SVHN}} & \multicolumn{4}{c|}{\textbf{CIFAR-10}} & \multicolumn{4}{c|}{\textbf{CIFAR-100}}\\
  \textbf{Method} & RMSCE-10 & RMSCE-15 & RMSCE-100 & RMSCE-500 & RMSCE-10 & RMSCE-15 & RMSCE-100 & RMSCE-500 & RMSCE-10 & RMSCE-15 & RMSCE-100 & RMSCE-500 \\
\midrule
 Baseline & $47.45_{0.83}$ & $47.59_{0.83}$ & $48.37_{0.83}$ & $49.01_{0.86}$ & $21.87_{0.62}$ & $22.35_{0.65}$ & $24.44_{0.81}$ & $25.63_{0.90}$ & $41.56_{1.20}$ & $41.87_{1.18}$ & $43.27_{1.11}$ & $44.06_{1.09}$ \\
 \midrule
  COG-FC & $31.40_{1.02}$ & $31.66_{1.01}$ & $32.16_{1.07}$ & $32.25_{1.06}$ & $19.19_{0.36}$ & $19.65_{0.37}$ & $20.75_{0.43}$ & $20.93_{0.43}$ & $22.07_{1.34}$ & $22.26_{1.36}$ & $22.59_{1.45}$ & $22.63_{1.46}$ \\
  COG+LS & $\mathbf{24.11_{0.56}}$ & $\mathbf{24.76_{0.65}}$ & $\mathbf{25.07_{0.63}}$ & $\mathbf{25.20_{0.64}}$ & $11.89_{0.27}$ & $11.79_{0.26}$ & $12.09_{0.27}$ & $12.15_{0.27}$ & $14.94_{2.26}$ & $15.37_{2.34}$ & $15.54_{2.35}$ & $15.57_{2.35}$ \\
  COG-TS & $38.94_{0.92}$ & $39.21_{0.94}$ & $40.27_{0.98}$ & $40.71_{1.02}$ & $24.84_{0.35}$ & $25.29_{0.37}$ & $27.01_{0.44}$ & $27.74_{0.44}$ & $33.18_{1.06}$ & $33.52_{1.05}$ & $34.54_{1.02}$ & $34.81_{1.03}$ \\
  COG-MS & $33.51_{1.18}$ & $33.79_{1.18}$ & $34.39_{1.22}$ & $34.49_{1.22}$ & $20.18_{0.83}$ & $20.63_{0.83}$ & $21.74_{0.92}$ & $22.03_{0.97}$ & $21.67_{2.00}$ & $21.85_{2.01}$ & $22.16_{2.05}$ & $22.19_{2.06}$ \\
  COG-TST & $29.51_{1.35}$ & $29.75_{1.35}$ & $30.20_{1.44}$ & $30.30_{1.45}$ & $19.19_{0.13}$ & $19.62_{0.10}$ & $20.66_{0.11}$ & $20.91_{0.14}$ & $\mathbf{13.44_{0.77}}$ & $\mathbf{13.51_{0.78}}$ & $\mathbf{13.57_{0.79}}$ & $\mathbf{13.60_{0.79}}$ \\\midrule
 Label Smoothing & $35.91_{0.50}$ & $35.66_{0.55}$ & $36.75_{0.57}$ & $36.98_{0.64}$ & $14.16_{1.26}$ & $14.09_{0.71}$ & $15.35_{0.94}$ & $15.57_{0.94}$ & $13.92_{0.96}$ & $14.43_{0.91}$ & $14.62_{0.91}$ & $14.66_{0.91}$ \\
 \midrule
 Temp. Scaling & $40.46_{0.84}$ & $40.68_{0.85}$ & $41.57_{0.94}$ & $41.94_{0.99}$ & $17.99_{0.75}$ & $18.56_{0.80}$ & $20.24_{0.94}$ & $20.70_{0.97}$ & $30.59_{1.21}$ & $30.94_{1.20}$ & $31.92_{1.21}$ & $32.19_{1.23}$ \\
  Matrix Scaling & $29.23_{0.83}$ & $29.39_{0.84}$ & $29.56_{0.83}$ & $29.65_{0.83}$ & $\mathbf{11.20_{0.99}}$ & $\mathbf{11.47_{1.03}}$ & $\mathbf{11.86_{1.07}}$ & $\mathbf{11.93_{1.06}}$ & $64.57_{1.71}$ & $64.58_{1.71}$ & $64.70_{1.68}$ & $64.82_{1.65}$ \\
  TST & $36.16_{0.08}$ & $36.50_{0.08}$ & $37.39_{0.10}$ & $37.58_{0.11}$ & $14.34_{0.13}$ & $14.79_{0.14}$ & $15.94_{0.16}$ & $16.16_{0.16}$ & $28.66_{1.57}$ & $28.90_{1.60}$ & $29.44_{1.69}$ & $29.52_{1.72}$ \\
  \midrule
 V-TST (MC=1) & $33.57_{2.09}$ & $33.76_{2.11}$ & $34.04_{2.17}$ & $34.12_{2.17}$ & $14.64_{0.95}$ & $15.05_{0.97}$ & $15.89_{1.06}$ & $16.01_{1.07}$ & $21.12_{3.56}$ & $21.28_{3.60}$ & $21.57_{3.73}$ & $21.62_{3.74}$ \\
  V-TST (MC=10) & $28.84_{1.83}$ & $29.01_{1.85}$ & $29.18_{1.89}$ & $29.26_{1.89}$ & $11.97_{1.03}$ & $12.33_{1.07}$ & $12.93_{1.18}$ & $13.01_{1.19}$ & $16.66_{3.06}$ & $16.78_{3.09}$ & $16.96_{3.17}$ & $17.00_{3.17}$ \\
\bottomrule
\end{tabular}
}
\end{table}

\begin{table}[h]
\caption{MCE averaged across all considered covariate shifts for all models and all variants seen in this work for multiple choices of the number of bins.}
\label{tab:WRN_shift_ood_mce}
\centering
\setlength{\tabcolsep}{3pt}
\resizebox{\textwidth}{!}{
\begin{tabular}{l|cccc|cccc|cccc|}
\toprule
   & \multicolumn{4}{c|}{\textbf{SVHN}} & \multicolumn{4}{c|}{\textbf{CIFAR-10}} & \multicolumn{4}{c|}{\textbf{CIFAR-100}}\\
  \textbf{Method} & MCE-10 & MCE-15 & MCE-100 & MCE-500 & MCE-10 & MCE-15 & MCE-100 & MCE-500 & MCE-10 & MCE-15 & MCE-100 & MCE-500 \\
\midrule
 Baseline & $60.58_{0.79}$ & $64.16_{0.72}$ & $68.20_{1.46}$ & $74.20_{0.61}$ & $44.61_{2.00}$ & $47.01_{1.82}$ & $60.00_{15.71}$ & $76.09_{2.62}$ & $59.26_{0.73}$ & $61.89_{0.86}$ & $68.29_{10.60}$ & $69.41_{9.97}$ \\
 \midrule
  COG-FC & $45.30_{1.81}$ & $45.53_{1.47}$ & $47.61_{1.88}$ & $\mathbf{51.98_{1.85}}$ & $31.98_{0.97}$ & $32.05_{0.80}$ & $32.98_{0.97}$ & $82.21_{1.30}$ & $31.84_{2.22}$ & $32.16_{2.21}$ & $33.02_{2.18}$ & $46.77_{26.24}$ \\
  COG+LS & $\mathbf{36.84_{1.36}}$ & $41.11_{2.38}$ & $81.65_{18.07}$ & $95.96_{6.08}$ & $19.74_{0.50}$ & $19.88_{0.61}$ & $23.76_{6.11}$ & $59.58_{32.80}$ & $27.12_{3.57}$ & $27.44_{3.97}$ & $49.32_{29.19}$ & $67.27_{30.02}$ \\
  COG-TS & $54.03_{1.25}$ & $56.26_{1.50}$ & $58.00_{1.73}$ & $69.84_{12.18}$ & $49.59_{17.78}$ & $50.80_{17.10}$ & $66.09_{19.91}$ & $79.75_{1.81}$ & $47.66_{1.03}$ & $47.98_{1.28}$ & $48.84_{1.04}$ & $50.20_{1.24}$ \\
  COG-MS & $48.16_{1.48}$ & $48.68_{1.56}$ & $57.82_{15.40}$ & $72.49_{15.32}$ & $33.43_{1.39}$ & $33.61_{1.66}$ & $34.62_{1.69}$ & $54.18_{23.89}$ & $31.79_{2.98}$ & $43.52_{28.00}$ & $34.01_{5.52}$ & $48.23_{25.65}$ \\
  COG-TST & $43.54_{1.81}$ & $43.89_{2.15}$ & $46.10_{2.06}$ & $63.69_{19.31}$ & $32.06_{0.62}$ & $32.02_{0.56}$ & $33.26_{0.47}$ & $\mathbf{53.98_{24.49}}$ & $\mathbf{20.16_{1.85}}$ & $\mathbf{20.33_{1.92}}$ & $\mathbf{21.17_{2.05}}$ & $33.41_{11.28}$ \\
  \midrule
 Label Smoothing & $51.63_{1.04}$ & $61.88_{3.98}$ & $84.78_{11.94}$ & $96.78_{0.45}$ & $31.25_{0.75}$ & $36.42_{5.48}$ & $51.99_{19.66}$ & $84.35_{6.56}$ & $26.35_{1.22}$ & $26.49_{1.10}$ & $53.97_{28.16}$ & $79.22_{26.95}$ \\
 \midrule
 Temp. Scaling & $54.51_{1.48}$ & $56.90_{1.50}$ & $58.98_{2.13}$ & $76.12_{8.13}$ & $35.06_{2.02}$ & $35.44_{1.88}$ & $38.76_{5.63}$ & $54.64_{23.93}$ & $46.10_{1.24}$ & $46.35_{1.52}$ & $46.99_{1.36}$ & $57.42_{19.52}$ \\
  Matrix Scaling & $40.97_{1.37}$ & $\mathbf{41.07_{1.22}}$ & $59.94_{22.55}$ & $69.96_{20.07}$ & $\mathbf{19.13_{1.79}}$ & $\mathbf{19.40_{1.78}}$ & $\mathbf{23.68_{5.54}}$ & $67.91_{19.62}$ & $70.04_{0.86}$ & $74.74_{0.81}$ & $82.00_{1.14}$ & $83.05_{1.16}$ \\
  TST & $53.19_{0.15}$ & $54.67_{0.24}$ & $55.88_{0.40}$ & $59.70_{0.77}$ & $27.57_{0.30}$ & $27.51_{0.33}$ & $40.38_{23.93}$ & $55.90_{25.83}$ & $40.96_{2.60}$ & $41.12_{2.51}$ & $41.89_{2.57}$ & $43.45_{2.58}$ \\
  \midrule
 V-TST (MC=1) & $47.33_{2.88}$ & $47.85_{2.58}$ & $57.20_{16.63}$ & $66.53_{17.01}$ & $27.68_{2.21}$ & $27.77_{2.18}$ & $33.13_{9.68}$ & $62.66_{28.70}$ & $31.65_{5.07}$ & $31.85_{5.12}$ & $33.03_{4.88}$ & $34.87_{4.98}$ \\
  V-TST (MC=10) & $42.39_{2.70}$ & $42.98_{2.71}$ & $\mathbf{44.85_{2.51}}$ & $64.20_{19.17}$ & $22.25_{2.36}$ & $22.46_{2.09}$ & $38.10_{26.33}$ & $61.53_{30.68}$ & $24.66_{4.31}$ & $24.92_{4.50}$ & $25.91_{4.53}$ & $\mathbf{28.96_{3.63}}$ \\
\bottomrule
\end{tabular}
}
\end{table}
\end{document}